\documentclass[10pt,twocolumn,letterpaper]{article}

\usepackage[pagenumbers]{wacv} 

\usepackage{graphicx}
\usepackage{amsmath}
\usepackage{amssymb}
\usepackage{booktabs}

\usepackage{comment}
\usepackage{bbding}
\usepackage{multirow}
\usepackage{dsfont}
\usepackage{subcaption}
\usepackage{times}
\newenvironment{packed_itemize}{
\begin{list}{\labelitemi}{\leftmargin=2em}
\vspace{-6pt}
 \setlength{\itemsep}{0pt}
 \setlength{\parskip}{0pt}
 \setlength{\parsep}{0pt}
}{\end{list}}
\newcommand{\methodname}{COT\xspace}

\usepackage[pagebackref,breaklinks,colorlinks]{hyperref}

\usepackage[capitalize]{cleveref}
\crefname{section}{Sec.}{Secs.}
\Crefname{section}{Section}{Sections}
\Crefname{table}{Table}{Tables}
\crefname{table}{Tab.}{Tabs.}

\def\wacvPaperID{673} 
\def\confName{WACV}
\def\confYear{2024}

\begin{document}

\title{Synergizing Contrastive Learning and Optimal Transport for 3D Point Cloud Domain Adaptation}

\author{%
Siddharth Katageri$^{1*}$\hspace{1cm}
Arkadipta De$^{2*}$\hspace{1cm}
Chaitanya Devaguptapu$^{2*}$\hspace{1cm}
VSSV Prasad$^{2}$\\
Charu Sharma$^{1}$\hspace{1cm}
Manohar Kaul$^{2}$\\
\textsuperscript{1}IIIT Hyderabad, India\hspace{0.5cm}
\textsuperscript{2}Fujitsu Research India\\
{\tt\small siddharth.katageri@research.iiit.ac.in},
{\tt\small charu.sharma@iiit.ac.in},
{\tt\small email@chaitanya.one},\\
{\tt\small \{Arkadipta.De, venkatasivasaivaraprasad.kasu, manohar.kaul\}@fujitsu.com}
}
\maketitle
\def\thefootnote{*}\footnotetext{These authors contributed equally to this work.}

\begin{abstract}
Recently, the fundamental problem of unsupervised domain adaptation (UDA) on 3D point clouds has been motivated by a wide variety of applications in robotics, virtual reality, and scene understanding, to name a few. 
The point cloud data acquisition procedures manifest themselves as significant domain discrepancies and geometric variations among both similar and dissimilar classes. 
The standard domain adaptation methods developed for images do not directly translate to point cloud data because of their complex geometric nature. 
To address this challenge, 
we leverage the idea of multimodality and alignment between distributions. We propose a new UDA architecture for point cloud classification that benefits from multimodal contrastive learning to get better class separation in both domains individually. Further, the use of optimal transport (OT) aims at learning source and target data distributions jointly to reduce the cross-domain shift and provide a better alignment.
We conduct a comprehensive empirical study on PointDA-10 and GraspNetPC-10 and show that our method achieves state-of-the-art performance on GraspNetPC-10 (with $\approx$ $4$-$12$\% margin) and best average performance on PointDA-10. 
Our ablation studies and decision boundary analysis also validate the significance of our contrastive learning module and OT alignment. \href{https://siddharthkatageri.github.io/COT/}{https://siddharthkatageri.github.io/COT}.
\end{abstract}

\section{Introduction}
\label{sec:intro}
Representation learning on 3D point clouds is rife with challenges, due to point clouds being irregular, unstructured, and unordered. Despite these hindrances posed by the nature of this complex dataset, learning representations on point clouds have achieved success in a gamut of computer vision areas, such as robotics \cite{voxnet}, self-driving vehicles \cite{mahjourian2018unsupervised}, and scene understanding \cite{zhu2017target}, to name a few.

While a majority of the point cloud representation learning works have focused on improving performance in supervised and unsupervised tasks~\cite{su15mvcnn,pointnet,wu20143d}, very few have focused on the task of \emph{domain adaptation} (DA) between disparate point cloud datasets.  
This is in part due to the significant differences in underlying structures (i.e., different backgrounds, orientations, illuminations etc. obtained from a variety of data acquisition methods and devices), which in turn manifest themselves as geometric variations and discrepancies between the source and target point cloud domains.
\begin{figure}
    \centering
    \includegraphics[width=\columnwidth,trim={0 1.8cm 0 6.25cm},clip]{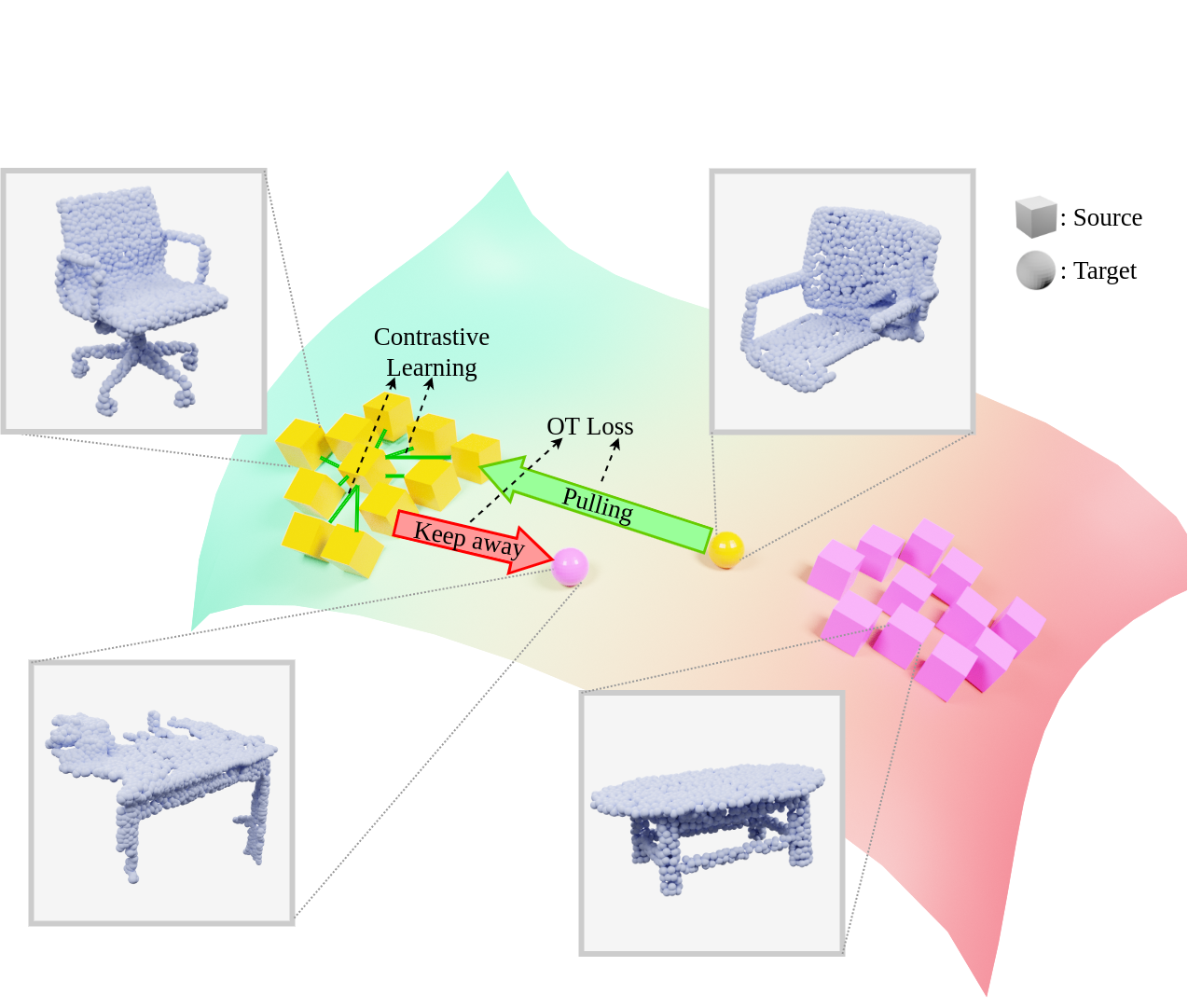}
    \caption{Overview of our method for UDA. Contrastive learning (CL) and optimal transport (OT) are designed to complement each other synergistically. CL establishes class clusters, while OT aligns objects across domains. The colors of data points denote different classes.}
    \label{fig:teaser}
\end{figure}
An important aspect of achieving cross-domain generalization is to leverage the trained model on simulated data (easy-to-get annotations) and generalize it to real-world data for which obtaining labels is a cumbersome task. The problem persists even in controlled simulated environments. For example, in VR environments, a chair's visual representation can vary significantly between a game and architectural design software. In the more demanding setting of \emph{unsupervised domain adaptation} (UDA) for classification, the source domain consists of labeled point clouds, while the target domain is completely unlabeled. 


Recent works focus on incorporating self-supervised learning (SSL) approaches to learn similar features for both domains, along with a regular source domain supervision~\cite{achituve2021self, zou2021geometry, Shen_2022_CVPR}. 
The point clouds belonging to the same class must not only be closer in each individual domain, but also achieve cross-domain alignment.
However, our analysis reveals that explicit cross-domain alignment is underexplored, given the significant margins between classification accuracies on source and target domains. 

Based on our aforementioned observations, we draw inspiration from recent SSL contrastive learning research \cite{simclr, khosla2020supervised, afham2022crosspoint}, which has enjoyed major success in other domains such as image and text. We propose a \textbf{C}ontrastive SSL method on point clouds 
to improve class separation individually in both source and target domains that share a common label space. In addition, optimal transport (OT) based methods \cite{damodaran2018deepjdot} have also shown promising results as they jointly learn the embeddings between both domains by comparing their underlying probability distributions and exploiting the geometry of the feature space. Thus, we employ \textbf{OT} to achieve better cross-domain alignment for domain adaptation. Figure~\ref{fig:teaser} provides a visual overview of our method (COT).



\begin{figure*}[h!]
  \begin{center}
  \includegraphics[scale=0.8,trim={0 0 1.7cm 0},clip]{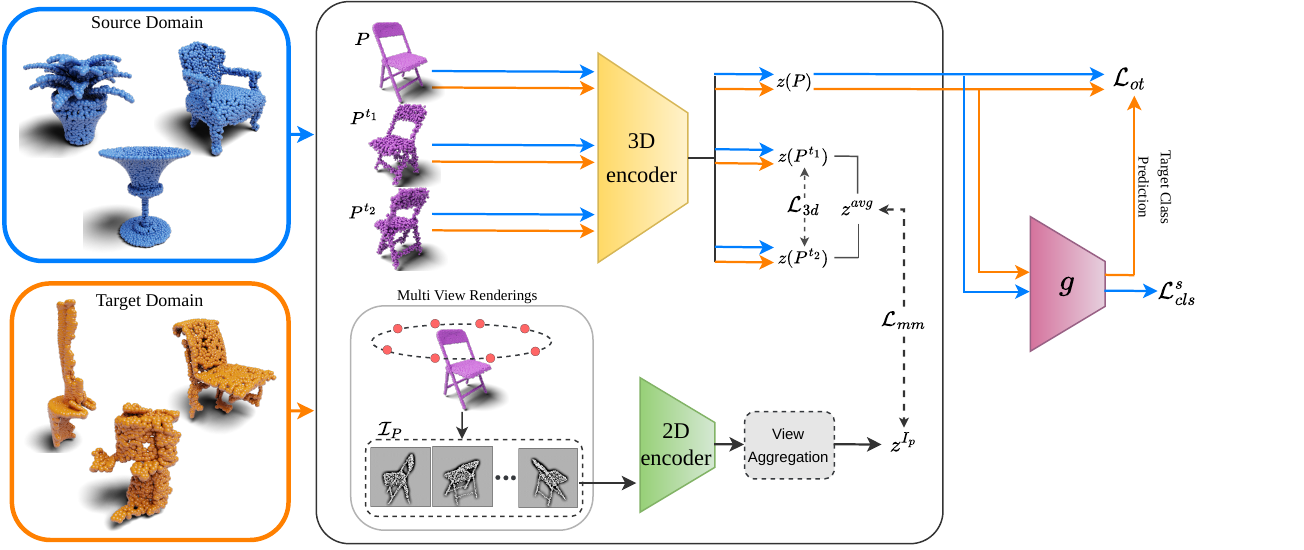}
  \end{center}
  \caption{Overview of our framework. Three main components: self-supervised contrastive training ($\mathcal{L}_{3d}$, $\mathcal{L}_{mm}$), self-supervised OT training between both domains ($\mathcal{L}_{ot}$) and a supervised training on source domain ($\mathcal{L}_{cls}$). Contrastive loss uses features from shared Point Cloud and Image encoders with point cloud augmentations and 2D image projections. OT and classifier losses takes features of original point cloud samples from shared Point Cloud encoder.}
  \label{fig:model-arch}
\end{figure*}

To reduce the domain shift and learn high quality transferable point cloud embeddings, we leverage the idea of multi-modality 
within the source and target domains and alignment 
between both their underlying data distributions. We design an end-to-end framework which consists of a multimodal self-supervised contrastive learning setup (shown in Fig. \ref{fig:model-arch}) for both source and target domains individually and OT loss for domain alignment. We also incorporate a regular supervised branch that considers labels from the source domain for training. 
The aim of our setup is to exploit the multimodality of the input data to learn quality embeddings in their respective domains, while reducing the cross-domain shift with the OT alignment.

\textbf{Main Contributions:}
\begin{packed_itemize}
    \item To the best of our knowledge, we are the first to propose the use of multimodal contrastive learning within individual domains along with OT for domain alignment for 3D point cloud domain adaptation. 
    \item We build an end-to-end framework with 
    two contrastive losses between 3D point cloud augmentations and between a point cloud and its 2D image projections. We also include OT loss for domain alignment.
    \item We perform an exhaustive empirical study on two popular benchmarks called PointDA-10 and GraspNetPC-10. Our method achieves state-of-the-art performance on GraspNetPC-10 (with $\approx$ $4$-$12$\% margin) and the best average performance on PointDA-10. Our method outperforms existing methods in the majority of cases with significant margins on challenging real-world datasets. We also conduct an ablation study and explore decision boundaries for our self-supervised contrastive and OT losses to elucidate the individual contributions of each component in our method.

\end{packed_itemize}

\section{Related Work}

\noindent\textbf{Domain Adaptation on Point Clouds}
Very few works~\cite{pointdan2019,achituve2021self, zou2021geometry, Shen_2022_CVPR} focus on the problem of domain adaptation on point clouds. \cite{pointdan2019} introduces a benchmark, \emph{PointDA-10} and an approach based on local and global alignment. \cite{achituve2021self} introduces a self-supervised approach based on deformation reconstruction and leverages \emph{PointMixup}~\cite{pointmixup2020}. \cite{zou2021geometry} learns a domain-shared representation of semantic categories by leveraging two self supervised geometric learning tasks as feature regularizers. \cite{Shen_2022_CVPR} proposes a self-supervised task of learning geometry-aware implicits for domain-specific variations and additionally propose a new dataset called \emph{GraspNetPC-10} that is developed from \emph{GraspNet}~\cite{graspnet2022}. These works mainly rely on the self-supervision task to improve adaptation, whereas we additionally propose to explicitly align classes across domains.


\noindent\textbf{Optimal Transport for Domain Adaptation}
Optimal transport based approaches~\cite{shen2017wasserstein,damodaran2018deepjdot,gautheron2018feature,Xu_2020_CVPR,fatras2021unbalanced} are commonly used in image domain adaptation by aligning the source and target representations. \cite{shen2017wasserstein} uses Wasserstein distance as a core loss in promoting similarities between embedded representations and proposes \emph{Wasserstein Distance Guided Representation Learning} (WDGRL). \cite{damodaran2018deepjdot} proposed \emph{DeepJDOT}, which computes a coupling matrix to transport the source samples to the target domain. \cite{gautheron2018feature} presents a new feature selection method that leverages the shift between the domains. \cite{Xu_2020_CVPR} proposed \emph{reliable weighted optimal transport} (RWOT) that exploits the spatial prototypical information and the intra-domain structure to dynamically measure the sample-level domain discrepancy across domains to obtain a precise-pair-wise optimal transport plan. \cite{fatras2021unbalanced} proposes an unbalanced optimal transport coupled with a mini-batch strategy to deal with large-scale datasets.
\section{Methodology}
\label{sec:method}
This section describes our method for UDA of point clouds for classification task. Our method is endowed by \emph{multimodal self-supervised contrastive learning and OT for domain alignment}. The self-supervised multi-modal contrastive learning module leverages both, the 3D information and their corresponding 2D image projections of point clouds. It produces initial class clusters in the source and target domains individually. Subsequently, our OT module better aligns the same class clusters across domains. We additionally also train a classifier on the source domain to improve the class separation, which in turn lessens the burden on our adaptation module.

Our setup aims at learning high quality embeddings, jointly for source and target domains, by exploiting both \emph{contrastive learning with augmentations} and the \emph{multimodal information of the input point clouds}, while simultaneously \emph{reducing the domain shift} across the domains. Our architecture is illustrated in Figure \ref{fig:model-arch}. 
To this end, we begin by describing self-supervised contrastive learning in Section \ref{subsec:ssl}. Next, Section \ref{subsec:preliminaryOT} briefly presents background concepts pertaining to OT and the Wasserstein distance, followed by an explanation of the domain alignment between source and target domains using OT in Section \ref{DAOT}. Finally, the overall training objective is presented in Section \ref{overall}.


Let a point cloud $P=\{x_1, \dots, x_n \}$, where $x_i \in \mathds{R}^3$, be a set of 3D points of cardinality $n$. 
Let $\mathcal{D}^s = \{ P_i^s, y_i \}_{i=1}^{n_s}$ denote the \emph{labeled source domain dataset}, where $P_i^s$ denotes the $i$-th source point cloud and $y_i$ its associated class label that takes values in $\mathcal{Y}=\{ 1, \dots, K \}$. Note that $\mathcal{Y}$ is a set of shared class labels that is \emph{common} to both the source and target domains.
The \emph{target domain dataset} $\mathcal{D}^t = \{ P_i^t \}_{i=1}^{n_t}$ contains unlabeled point clouds. The cardinality of $\mathcal{D}^s$ and $\mathcal{D}^t$ are $n_s$ and $n_t$ respectively.
Then, the task of UDA for point cloud classification boils down to learning a \emph{domain invariant function} $f: \mathcal{P} \rightarrow \mathcal{Y}$, where $\mathcal{P}$ is a union of unlabeled point clouds from both $\mathcal{D}^s$ and $\mathcal{D}^t$.

\subsection{Self-Supervised Contrastive Learning}
\label{subsec:ssl}
Motivated by the advancement of contrastive learning \cite{simclr, khosla2020supervised}, where the goal is to pull samples from common classes closer in the embedding space, we build a method to extract 3D and 2D features of point clouds and fuse this information to form initial domain class clusters. 

We employ a contrastive loss between augmented versions of a point cloud, which we term as a \emph{3D-modal association loss}, to learn similar features for samples from the same class. This loss forces the point cloud learning to be invariant to geometric transformations.
Additionally, we introduce a contrastive loss between the 3D point cloud features and their corresponding projected 2D image features, termed as \emph{multi-modal association loss}. The intuition behind this multi-modal loss is to take advantage of the rich multi-view latent 2D information inherent in the 3D point clouds.
Next, we explain these components in detail.

\noindent\textbf{3D-modal association loss} Let $P_b$ be a point cloud from a randomly drawn batch $B$ of size $k$ from either $\mathcal{D}^s$ or $\mathcal{D}^t$. Given a set of affine transformations $T$, we generate two augmented point clouds $P_b^{t_1}$ and $P_b^{t_2}$, where $t_1$ and $t_2$ are compositions of transformations picked randomly from $T$. Additionally, we use random point dropout and add random noise to each point in a point cloud individually to introduce object surface distortions. These transformations introduce geometric variations, which are then used to curate samples that serve as positive pairs.
The augmented point clouds $P_b^{t_1}$ and $P_b^{t_2}$ are then mapped to a $d$-dimensional feature space using a 3D encoder function 
producing embeddings $z(P_b^{t_1})$ and $z(P_b^{t_2})$, respectively.
These embeddings serve as \emph{positive pairs} and therefore our objective is to place them closer to one another in the feature space.

We define the similarity between the $i$-th embedding transformed by $t_{x}$ and the $j$-th embedding transformed by $t_{x}$, with $x \in \{1,2\}$, as 
\begin{equation}
\langle (i,t_{x}), (j,t_{x}) \rangle_{ST} = 
\exp \left( s( z(P_i^{t_{x}}), z(P_j^{t_{x}})) / \tau \right) 
\end{equation}
where $s$ denotes the cosine-similarity function and $\tau$ is the temperature hyperparameter. 

Our 3D-modal association loss is then given by
\begin{equation}
    \footnotesize
        \mathcal{L}_{3d} = - \log 
        \left\{   
        \frac{\langle (i,t_{1}), (i,t_{2}) \rangle_{ST}}
        {\sum\limits_{j=1}^k { \langle (i,t_{1}), (j,t_{1}) \rangle_{ST}} 
        +
        \sum\limits_{j=1}^k { \langle (i,t_{1}), (j,t_{2}) \rangle_{ST}}
        } \right\} 
        \label{loss_3d}
\end{equation}


For both source and target, we randomly draw respective batches and perform 3D-modal association separately. This method of self-supervised contrastive learning generates class clusters in both domains individually and has been shown to be useful especially for the target domain, as its supervision signal is missing. We further guide the feature learning by introducing image modality in the optimization. We explain our multi-modal association loss next.

\noindent\textbf{Multi-modal association loss} We consider using point cloud projections in our method, as the image features can provide another level of discriminative information.
2D projections from various viewpoints allow capturing \emph{silhouette} and \emph{surface boundary} information for shape understanding that is harder to derive from just point-wise distances.
Breaking away from the common way of fusing multimodal information \cite{vqamain, KAFLE20173} where the embeddings of two modalities are fused by simply concatenating or averaging them, we instead compute associative losses between 3D features and image features to establish 2D-3D correspondence understanding helping to provide informative global representation.

As contrastive learning is known to be good for alignment tasks, we advocate using a contrastive objective to fuse multimodal (3D and 2D) information. Let $\mathcal{I}_P=\{I_n\}_{n=1}^m$ be the set of $m$ 2D image projections of point cloud $P$.
To generate these images, we set virtual cameras around the object in a circular fashion to obtain views of the object from all directions. For a point cloud $P$, each of its corresponding 2D images is passed to a 2D encoder, generating a $d$-dimensional embedding.
Following \cite{su15mvcnn, Hamdi_2021_ICCV}, we use a simple max-pooling operation to aggregate feature information from all views and get a $d$-dimensional vector $z^{I_{P}}$.
In order to fuse the 3D augmented point cloud embeddings (i.e., $z(P^{t_1})$ and $z(P^{t_2})$) with the 2D point cloud embedding $z^{I_P}$, we compute the average of the 3D augmented point cloud embeddings to get $z^{avg}$.
We then use the $z^{avg}$ and $z^{I_{P}}$ that contain summarized information from 3D and 2D modalities respectively in a self-supervised contrastive loss to maximize their similarity in the embedding space. 
We define the similarity between the $i$-th embedding $z_i$ and the $j$-th embedding $z'_{j}$ as 
$\langle z_i, z'_j \rangle_S = 
\exp \left( s( z_i,z'_j  ) / \tau \right) 
$. Then, our multi-modal association loss is given by
\begin{equation}
    \mathcal{L}_{mm} = - \log \left\{   
    \frac{
\langle z_i^{avg}, z_i^{I_{P}}  \rangle_{S}    
    }
    {
\sum\limits_{j=1}^k { \langle z_i^{avg},z_j^{avg} \rangle_{S}   }
+
\sum\limits_{j=1}^k { \langle z_i^{avg},z_j^{I_P} \rangle_{S}   }
    }                    
                        \right\}
    \label{loss_mm}
\end{equation}
The total self-supervised contrastive loss is given by adding the 3D-modal association loss ($\mathcal{L}_{3d}$) that maximizes the similarity between augmentations of a point cloud and the multi-modal association loss ($\mathcal{L}_{mm}$) that maximizes the similarity between 3D and 2D features of a point cloud.

\subsection{Optimal Transport and Wasserstein Distance}
\label{subsec:preliminaryOT}
Optimal transport offers a way to compare two probability distributions irrespective of whether the measures have common support. It aims to find the most efficient way of transferring mass between two probability distributions, considering the underlying geometry of the probability space. Formally, given two probability distributions $\mu$ and $\nu$ on a metric space $\mathcal{X}$, for $p \ge 1$,  the $p$-Wasserstein distance \cite{villani} is given by $W_{p}(\mu, \nu) = \left(\inf_{\pi \in \Pi(\mu, \nu)} \int_{\mathcal{X} \times \mathcal{X}} c(x, y)^p d\pi(x, y) \right)^{1/p}$
where $\pi$ is a transport plan that defines a flow between mass from $\mu$ to locations in $\nu$, $\Pi(\mu, \nu)$ is the joint probability distribution with the marginals $\mu$ and $\nu$ and $c(x, y)$ is the ground metric which assigns a cost of moving a unit of mass $x \in \mathcal{X}$ from $\mu$ to some location $y \in \mathcal{X}$ in $\nu$.

For the discrete case, given two discrete distributions $\hat{\mu}=\sum_{i=1}^{m}a_{i}\delta(x_i)$ and $\hat{\nu}=\sum_{j=1}^{n}b_{j}\delta(y_j)$, where $\{a_i\}_{i=1}^{m}$ and $\{b_j\}_{j=1}^{n}$ are the probability masses that should sum to 1, $\{x_i\}_{i=1}^{m}$ and $\{y_j\}_{j=1}^{n}$ are the support points in $\mathbb{R}^{d}$ with $m$ and $n$ being the number of points in each measure. The discrete form of the above equation can be given as $W_p(\hat{\mu}, \hat{\nu}) = \left(\min_{\psi \in U(a,b)} \langle C^{p}, \psi \rangle_F  \right)^{1/p}$, where $\langle \cdot, \cdot \rangle_F$ denotes the Frobenius dot-product, $C^p \in \mathbb{R}^{m \times n}_+$ is the pairwise ground metric distance, $\psi$ is the coupling matrix and $U$ is the set of all possible valid coupling matrices, i.e. $U(a,b)=\{\psi \in \mathbb{R}^{m \times n}: \psi \mathds{1}_n=a,  \psi^{\top}\mathds{1}_m=b\}$.

\subsection{Domain Alignment via Optimal Transport}
\label{DAOT}
As explained in Section \ref{subsec:ssl}, contrastive learning generates class clusters in source and target domains individually. The underlying idea is to further achieve alignment of point clouds belonging to the same class across two domains. We leverage an OT based loss that uses point cloud features and source labels for domain alignment. The classifier $g:\mathbb{R}^d \rightarrow \mathcal{Y}$ that maps the point cloud embedding from feature space to label space also needs to work well for the target domain. The OT flow is greatly dependent on the choice of the cost function as shown by \cite{ground_metric}. Here, as we want to jointly optimize the feature and the classifier decision boundary learning, we define our cost function as 
\begin{equation}
    \sum_{i=1}^{k} \sum_{j=1}^{k} c(z^{s}_{i}, z^{t}_{j})= \alpha||z^{s}_{i} - z^{t}_{j}||^2_2 + \beta ||y_i^s - g(z^{t}_{j})||^2_2
\end{equation}
where superscripts $s$ and $t$ denote the source and target domains, respectively.
$\alpha$, $\beta$ are the weight coefficients. Here, the first term computes the squared-$L2$ distance between the embeddings of source and target samples. The second term computes squared-$L2$ distance between the classifier's target class prediction and the source ground truth label. Jointly, these two terms play an important role in pulling or keeping apart source and target samples for achieving domain alignment. For example, if a target sample lies far from a source sample having the same class, the first term would give a high cost. However, for a decently trained classifier, the distance between its target class prediction and source ground truth label would be less, thus making the second term low. This indicates that these source and target samples must be pulled closer. Conversely, if a target sample lies close to a source sample having a different class, the first term would be low, and the second term would be high, indicating this sample should be kept apart. As evident from the example, the second term is a guiding entity for inter-domain class alignment. It penalizes source-target samples based on their classes and triggers a pulling mechanism. The problem of finding optimal matching can be formulated as $\psi^{\ast} = \min_{\psi \in U(a_s, b_t)}\langle C^p, \psi \rangle_F$, where $\psi^{\ast}$ is the ideal coupling matrix, $a_s$ and $b_t$ are the uniform marginal distributions of source and target samples from a batch. The optimal coupling matrix $\psi^{\ast}$ is computed by freezing the weights of the 3D encoder function and the classifier function $g$. The OT loss for domain alignment is given by
\begin{equation}
     \mathcal{L}_{ot} = \sum_{i=1}^{k} \sum_{j=1}^{k} \psi^{\ast}_{ij} ( \alpha||z^{s}_{i} - z^{t}_{j}||_2^2 + \beta \mathcal{L}_{ce}(y_i^s, g(z^{t}_{j}))
\end{equation}
where $\mathcal{L}_{ce}$ is the cross-entropy loss.

\subsection{Overall Training Loss}
\label{overall}
The overall pipeline of our unsupervised DA method is trained with the combination of the following objective functions $ 
    \mathcal{L}_{total} = \mathcal{L}_{3d} + \mathcal{L}_{mm} + \mathcal{L}_{ot} + \mathcal{L}_{cls}^{s}$.
The loss consists of three self-supervised losses (i.e., $\mathcal{L}_{3d}$, $\mathcal{L}_{mm}$ and $\mathcal{L}_{ot}$) and a supervised loss $\mathcal{L}_{cls}^{s}$. Besides three SSL tasks, supervised learning is performed based on source samples and labels. For this purpose, a regular cross-entropy loss or a \emph{mixup} variant can be applied \cite{zhang2018mixup}. We use a supervised loss $(\mathcal{L}_{cls}^{s})$ inspired by the PointMixup method (PCM) \cite{pointmixup2020}. PCM is a data augmentation method for point clouds by computing interpolation between samples. 
Augmentation strategies have proven to be effective and enhance the representation capabilities of the model. Similarly, PCM has shown its potential to generalize across domains and robustness to noise and geometric transformations.

We also employ the \emph{self-paced self-training} (SPST) strategy introduced by \cite{zou2021geometry} to improve the alignment between domains. In SPST, pseudo-labels for the target samples are generated using the classifier's prediction and confidence threshold. 
The first step computes the pseudo labels for the target samples depending on the confidence of their class predictions, while the next step updates the point cloud encoder and classifier with the computed pseudo labels for target and ground truth labels of source.
In our method, we use SPST strategy as a fine-tuning step for our models.

\begin{table*}[ht!]
\centering
\scriptsize
\resizebox{\textwidth}{!}{%
\begin{tabular}{lc||ccccccc} 
\hline
\textbf{Methods} & SPST & M $\rightarrow$ S & M $\rightarrow$ S* & S $\rightarrow$ M & S $\rightarrow$ S* & S* $\rightarrow$ M & S* $\rightarrow$ S & Avg. \\ 
\hline
Supervised & & 93.9 $\pm$ 0.2  & 78.4 $\pm$ 0.6 & 96.2 $\pm$ 0.1 & 78.4 $\pm$ 0.6 & 96.2 $\pm$ 0.1 & 93.9 $\pm$ 0.2 & 89.5 \\
\multicolumn{1}{c}{Baseline(w/o adap.)} & & 83.3 $\pm$ 0.7 & 43.8 $\pm$ 2.3 & 75.5 $\pm$ 1.8 & 42.5 $\pm$ 1.4 & 63.8 $\pm$ 3.9 & 64.2 $\pm$ 0.8 & 62.2 \\ 
\hline
DANN\cite{DANN} & & 74.8 $\pm$ 2.8 & 42.1 $\pm$ 0.6 & 57.5 $\pm$ 0.4  & 50.9 $\pm$ 1.0 & 43.7 $\pm$ 2.9 & 71.6 $\pm$ 1.0 & 56.8 \\
PointDAN\cite{pointdan2019} & & 83.9 $\pm$ 0.3 & 44.8 $\pm$ 1.4 & 63.3 $\pm$ 1.1 & 45.7 $\pm$ 0.7 & 43.6 $\pm$ 2.0 & 56.4 $\pm$ 1.5 & 56.3 \\
RS\cite{RS} & & 79.9 $\pm$ 0.8   & 46.7 $\pm$ 4.8  & 75.2 $\pm$ 2.0 & 51.4 $\pm$ 3.9 & 71.8 $\pm$ 2.3 & 71.2 $\pm$ 2.8  & 66.0 \\
Defrec+PCM\cite{achituve2021self} & & 81.7 $\pm$ 0.6  & 51.8 $\pm$ 0.3 & \textbf{78.6} $\pm$ 0.7 & 54.5 $\pm$ 0.3 & 73.7 $\pm$ 1.6 & 71.1 $\pm$ 1.4 & 68.6 \\ \hline
\multirow{2}{*}{GAST\cite{zou2021geometry}} & & \underline{83.9} $\pm$ 0.2 & \textbf{56.7} $\pm$ 0.3 & 76.4 $\pm$ 0.2& \underline{55.0} $\pm$ 0.2 & 73.4 $\pm$ 0.3 & 72.2 $\pm$ 0.2 & 69.5 \\
                      & \Checkmark & \textcolor{blue}{\underline{84.8} $\pm$ 0.1} & \textcolor{blue}{\textbf{59.8} $\pm$ 0.2} & \textcolor{blue}{80.8 $\pm$ 0.6} & \textcolor{blue}{\underline{56.7} $\pm$ 0.2} & \textcolor{blue}{81.1 $\pm$ 0.8} & \textcolor{blue}{\underline{74.9} $\pm$ 0.5} & \textcolor{blue}{73.0} \\ \hline
\multirow{2}{*}{ImplicitPCDA\cite{Shen_2022_CVPR}} & & \textbf{85.8} $\pm$ 0.3 & \underline{55.3} $\pm$ 0.3 & 77.2 $\pm$ 0.4 & \textbf{55.4} $\pm$ 0.5 & \underline{73.8}$\pm$ 0.6 & \underline{72.4} $\pm$ 1.0& \underline{70.0} \\
                              & \Checkmark & \textcolor{blue}{\textbf{86.2} $\pm$ 0.2} & \textcolor{blue}{\underline{58.6} $\pm$ 0.1} & \textcolor{blue}{\underline{81.4} $\pm$ 0.4} & \textcolor{blue}{\textbf{56.9} $\pm$ 0.2} & \textcolor{blue}{\underline{81.5} $\pm$ 0.5} & \textcolor{blue}{74.4 $\pm$ 0.6} & \textcolor{blue}{\underline{73.2}} \\ 
\hline
\multirow{3}{*}{\methodname} 
 & & 83.2 $\pm$ 0.3 & 54.6 $\pm$ 0.1  & \underline{78.5} $\pm$ 0.4 & 53.3 $\pm$ 1.1  & \textbf{79.4} $\pm$ 0.4 & \textbf{77.4} $\pm$ 0.5 & \textbf{71.0} \\
                       & \Checkmark & \textcolor{blue}{84.7 $\pm$ 0.2} & \textcolor{blue}{57.6 $\pm$ 0.2} & \textcolor{blue}{\textbf{89.6} $\pm$ 1.4}  & \textcolor{blue}{51.6 $\pm$ 0.8} & \textcolor{blue}{\textbf{85.5} $\pm$ 2.2} & \textcolor{blue}{\textbf{77.6} $\pm$ 0.5} & \textcolor{blue}{\textbf{74.4}}  \\ 
                      \hline
\end{tabular}}
\caption{Classification accuracy (\%) on the PointDA-10. M: ModelNet, S: ShapNet, S*: ScanNet; $\rightarrow$ indicates the adaptation direction. OT: Optimal transport, SPST: self-paced self-training. Results in black and blue represent accuracy without and with SPST strategy, respectively. Bold represents the best result and underlined represents the second best for both the colors.
}
\label{tab:pointda-10}
\end{table*}

\begin{table*}[htb!]
\centering
\tiny
\resizebox{0.9\textwidth}{!}{%
\begin{tabular}{lc||ccccccc} 
\hline
\textbf{Methods} & SPST & Syn. $\rightarrow$ Kin. & Syn $\rightarrow$ RS. & Kin. $\rightarrow$ RS. & RS. $\rightarrow$ Kin. &  Avg. \\ 
\hline
Supervised & & 97.2 $\pm$ 0.8  & 95.6 $\pm$ 0.4 & 95.6 $\pm$ 0.3 & 97.2 $\pm$ 0.4 & 96.4 \\
\multicolumn{1}{c}{Baseline(w/o adap.)} & & 61.3 $\pm$ 1.0 & 54.4 $\pm$ 0.9 & 53.4 $\pm$ 1.3 & 68.5 $\pm$ 0.5 &  59.4  \\ 
\hline
DANN\cite{DANN} & & 78.6 $\pm$ 0.3 & 70.3 $\pm$ 0.5 & 46.1 $\pm$ 2.2  & 67.9 $\pm$ 0.3 & 65.7 \\
PointDAN\cite{pointdan2019} & & 77.0 $\pm$ 0.2 & 72.5 $\pm$ 0.3 & 65.9 $\pm$ 1.2 & 82.3 $\pm$ 0.5 & 74.4\\
RS\cite{RS} & & 67.3 $\pm$ 0.4   & 58.6 $\pm$ 0.8  & 55.7 $\pm$ 1.5 & 69.6 $\pm$ 0.4 &  62.8\\
Defrec+PCM\cite{achituve2021self} & & 80.7 $\pm$ 0.1  & 70.5 $\pm$ 0.4 & 65.1 $\pm$ 0.3 & 77.7 $\pm$ 1.2 &  73.5 \\ \hline
\multirow{2}{*}{GAST\cite{zou2021geometry}} & & 69.8 $\pm$ 0.4 & 61.3 $\pm$ 0.3 & 58.7 $\pm$ 1.0 & 70.6 $\pm$ 0.3& 65.1 \\
                      & \Checkmark & \textcolor{blue}{81.3$\pm$ 1.8} & \textcolor{blue}{72.3 $\pm$ 0.8} & \textcolor{blue}{61.3 $\pm$ 0.9} & \textcolor{blue}{80.1 $\pm$ 0.5} & \textcolor{blue}{73.8} \\ \hline
\multirow{2}{*}{ImplicitPCDA\cite{Shen_2022_CVPR}} & & \underline{81.2} $\pm$ 0.3 & \underline{73.1} $\pm$ 0.2 & \underline{66.4} $\pm$ 0.5 & \underline{82.6} $\pm$ 0.4 & \underline{75.8} \\
                              & \Checkmark & \textcolor{blue}{\underline{94.6} $\pm$ 0.4} & \textcolor{blue}{\underline{80.5} $\pm$ 0.2} & \textcolor{blue}{\underline{76.8} $\pm$ 0.4} & \textcolor{blue}{\underline{85.9} $\pm$ 0.3} &  \textcolor{blue}{\underline{84.4}} \\ 
\hline
\multirow{2}{*}{\methodname} & & \textbf{87.7} $\pm$ 0.7 & \textbf{80.2} $\pm$ 2.1  & \textbf{69.3} $\pm$ 5.2 & \textbf{85.8} $\pm$ 4.3  & \textbf{80.0} \\
                      & \Checkmark & \textcolor{blue}{\textbf{98.2} $\pm$ 0.5} & \textcolor{blue}{\textbf{83.7} $\pm$ 0.2} & \textcolor{blue}{\textbf{81.9} $\pm$ 2.1} & \textcolor{blue}{\textbf{98.0} $\pm$ 0.1} & \textcolor{blue}{\textbf{91.0}} \\
                      \hline
\end{tabular}}
\caption{Classification accuracy (\%) on the GraspNet-10 dataset. Sys.: Synthetic domain, Kin.: Kinect domain, RS.: Real domain; $\rightarrow$ indicates the adaptation direction. OT: Optimal transport, and SPST: self-paced self-training. Results in black and blue represent accuracy without and with SPST strategy, respectively. Bold represents the best result and underlined represents the second best for both the colors.
}
\label{tab:Graspnet-10}
\end{table*}

\section{Experiments}
We conduct an exhaustive experimental study to show the effectiveness of the learned representations and the significance of our COT. Our model is evaluated on two benchmark datasets with and without the SPST strategy for the classification task. We consider recent state-of-the-art self-supervised methods such as DANN \cite{DANN}, PointDAN \cite{pointdan2019}, RS \cite{RS}, DefRec+PCM \cite{achituve2021self}, GAST \cite{zou2021geometry} and ImplicitPCDA \cite{Shen_2022_CVPR} for comparison. Additionally, we report results for the baseline without adaptation (unsupervised) which trains the model using labels from the source domain and tests on the target domain. The supervised method is the upper bound which takes labels from the target domain into consideration during training. We will release our code upon acceptance.

\subsection{Datasets}
\paragraph{PointDA-10} introduced by \cite{pointdan2019} is a combination of ten common classes from 
ModelNet~\cite{wu20143d}, ShapeNet~\cite{chang2015shapenet} and ScanNet~\cite{dai2017scannet}. ModelNet and ShapeNet are synthetic datasets sampled
from 3D CAD models, containing $4,183$ training, $856$ test samples and $17,378$ training, $2,492$ test samples, respectively. On the other hand, ScanNet consists of point clouds from scanned and reconstructed real-world scenes and consists of  $6,110$ training and $1,769$ test samples. Point clouds in ScanNet are usually incomplete because of occlusion by surrounding objects in the scene or self-occlusion in addition to realistic sensor noises. We follow the standard data preparation procedure used in \cite{pointdan2019, achituve2021self, zou2021geometry, Shen_2022_CVPR}. 

\paragraph{GraspNetPC-10} \cite{Shen_2022_CVPR} consists of synthetic and real-world point clouds for ten object classes. It is developed from  GraspNet \cite{graspnet2022} by re-projecting raw depth scans to 3D space and applying object segmentation masks to crop out the corresponding point clouds. Raw depth scans are captured by two different depth cameras, Kinect2 and Intel Realsense to generate real-world point clouds. In the Synthetic, Kinect, and RealSense domains, there are $12,000$ training, $10,973$ training, $2,560$ testing, and $10,698$ training, $2,560$ testing point clouds, respectively. There exist different levels of geometric distortions and missing parts. Unlike PointDA-10, point clouds in GraspNetPC-10 are not aligned and all domains have almost uniform class distribution.
\vspace{-2mm}
\paragraph{Implementation Details} We use DGCNN\cite{dgcnn2019} as the point cloud feature extractor and pre-trained ResNet-50\cite{he2015deep} as the feature extractor for images to get $1024$-dimensional embedding vectors. For the contrastive losses ($\mathcal{L}_{3d}$, $\mathcal{L}_{mm}$) we convert these $1024$-dimensional embeddings to $256$ dimensions using projection layers. The classifier network consists of three fully connected layers with dropout and batch normalization.
We use rendered point cloud images of size $224 \times 224$ and set the number of views to $12$.
In total, we train our models for $150$ epochs for PointDA-10 and $120$ epochs for GraspNetPC-10 with a batchsize of $32$ on NVIDIA RTX-2080Ti GPUs and perform three runs with different seeds. We report results from the model with the best classification accuracy on source validation set, as target labels are unavailable. We provide more details about the implementation setup in our supplementary material.

\begin{figure*}[h!]
     \centering
     \begin{subfigure}[b]{0.20\textwidth}
         \centering
         \includegraphics[width=0.8\textwidth]{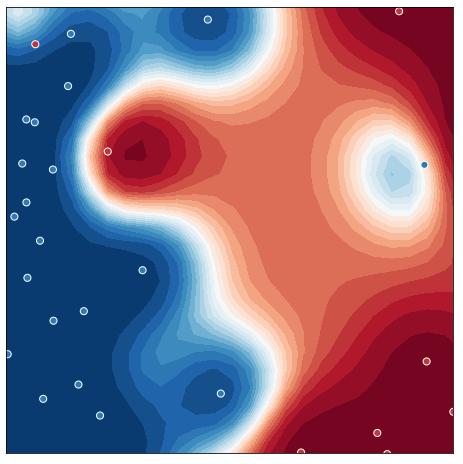}
         \caption{}
         \label{fig:init_boundary_pcm_only_1}
     \end{subfigure}
     \begin{subfigure}[b]{0.20\textwidth}
         \centering
         \includegraphics[width=0.8\textwidth]{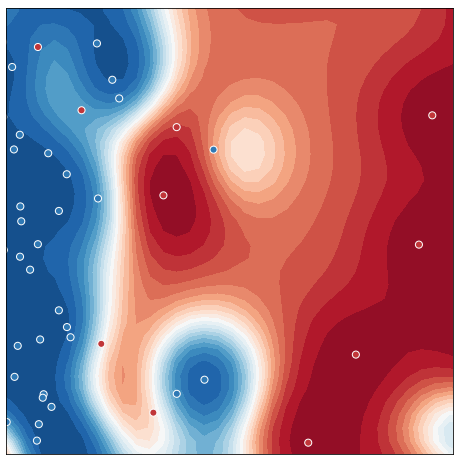}
         \caption{}
         \label{fig:init_boundary_pcm_cl_1}
     \end{subfigure}
     \begin{subfigure}[b]{0.20\textwidth}
         \centering
         \includegraphics[width=0.8\textwidth]{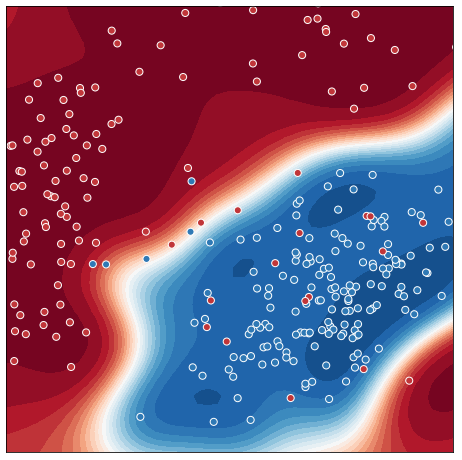}
         \caption{}
         \label{fig:init_boundary_pcm_cl_ot_1}
     \end{subfigure}
     \begin{subfigure}[b]{0.20\textwidth}
         \centering
         \includegraphics[width=0.8\textwidth]{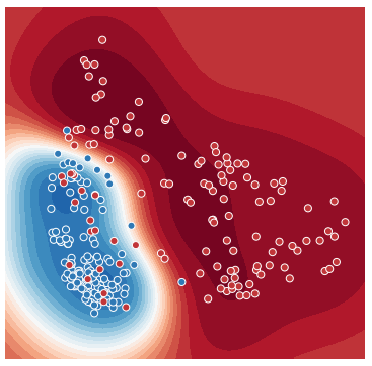}
         \caption{}
         \label{fig:init_boundary_pcm_cl_ot_spst_1}
     \end{subfigure}
     \\
     \begin{subfigure}[b]{0.20\textwidth}
         \centering
         \includegraphics[width=0.8\textwidth]{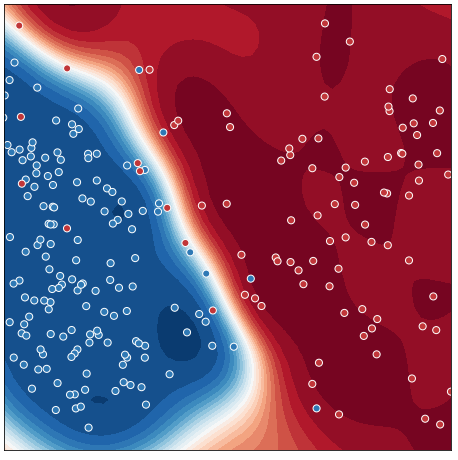}
         \caption{}
         \label{fig:final_boundary_pcm_only_1}
     \end{subfigure}
     \begin{subfigure}[b]{0.20\textwidth}
         \centering
         \includegraphics[width=0.8\textwidth]{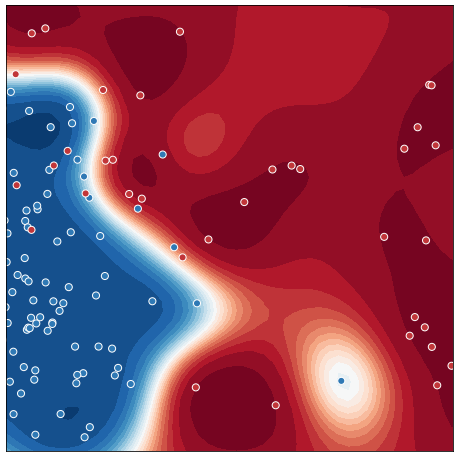}
         \caption{}
         \label{fig:final_boundary_pcm_cl_1}
     \end{subfigure}
     \begin{subfigure}[b]{0.20\textwidth}
         \centering
         \includegraphics[width=0.8\textwidth]{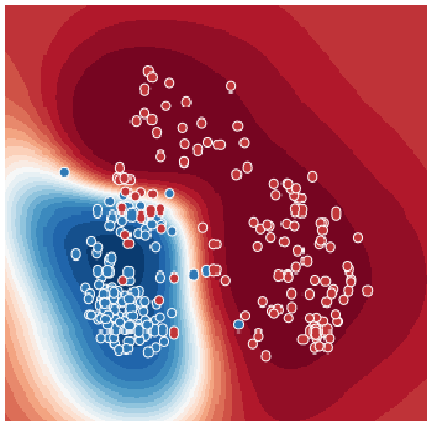}
         \caption{}
         \label{fig:final_boundary_pcm_cl_ot_1}
     \end{subfigure}
     \begin{subfigure}[b]{0.20\textwidth}
     \centering
     \includegraphics[width=0.8\textwidth]{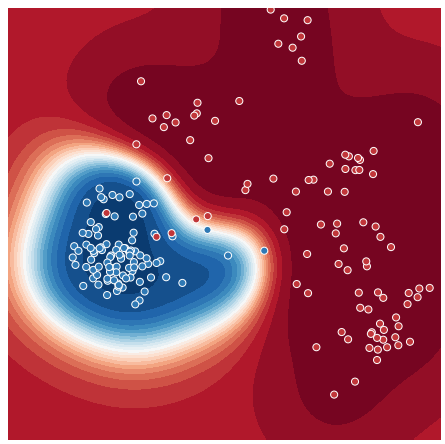}
     \caption{}
     \label{fig:final_boundary_pcm_cl_ot_spst_1}
     \end{subfigure}
    \caption{Early (top-row) and final (bottom-row) epochs decision boundaries on target samples for One-vs-Rest (Monitor class) for \textit{S} $\rightarrow$ \textit{M}. (a), (e) Only PCM (without adaptation), (b), (f) Contrastive learning with PCM, (c), (g) Optimal transport and contrastive learning with PCM (Our COT) and (d), (h) Our COT fine-tuned with SPST.}
    \label{fig:all_decision boundary_1}
\end{figure*}

\subsection{Unsupervised DA: Classification}
In Tables (\ref{tab:pointda-10}, \ref{tab:Graspnet-10}), we compare the results of our \methodname with the existing point cloud domain adaptation methods \cite{pointdan2019,achituve2021self, zou2021geometry, Shen_2022_CVPR} on PointDA-10 and GraspNetPC-10 datasets respectively. Similar to \cite{Shen_2022_CVPR} and \cite{zou2021geometry}, we also test our methodology with SPST strategy. As shown in Table \ref{tab:pointda-10}, \methodname achieves SoTA performance in terms of the overall average performance on PointDA-10 dataset. 
We observe that COT beats existing methods by a huge margin when the target dataset is synthetic. This is because target point clouds have well-defined geometry, and the classifier can make accurate predictions with high confidence, thus majorly helping alignment. As existing methods only propose to use self-learning tasks, their performance is very low compared to our self-learning task with explicit domain alignment endowed by OT. For the settings where the target dataset is real, it becomes harder for the classifier to provide good predictions, making the alignment process noisy. In these settings, we achieve on-par results compared to the existing methods. In $S \rightarrow M$, our method with SPST strategy outperforms existing methods $\approx 8\%$, and in $M \rightarrow S$, we achieve on-par results compared to the existing methods. We also use t-SNE to visualize the learned features of both domains (shown in supplementary). For PointDA-10, we observe that when the target domain is synthetic, the learned features are distinctive; however, when the target domain is real, the features lack distinctive power. This portrays the challenging setting of synthetic to real adaptation. Overall we achieve the highest average accuracy on PointDA-10 dataset showing effectiveness of COT.
\begin{figure}[h!]
     \centering
     \begin{subfigure}[b]{0.23\textwidth}
         \centering
         \includegraphics[trim={0 0 2cm 0},clip,width=\textwidth]{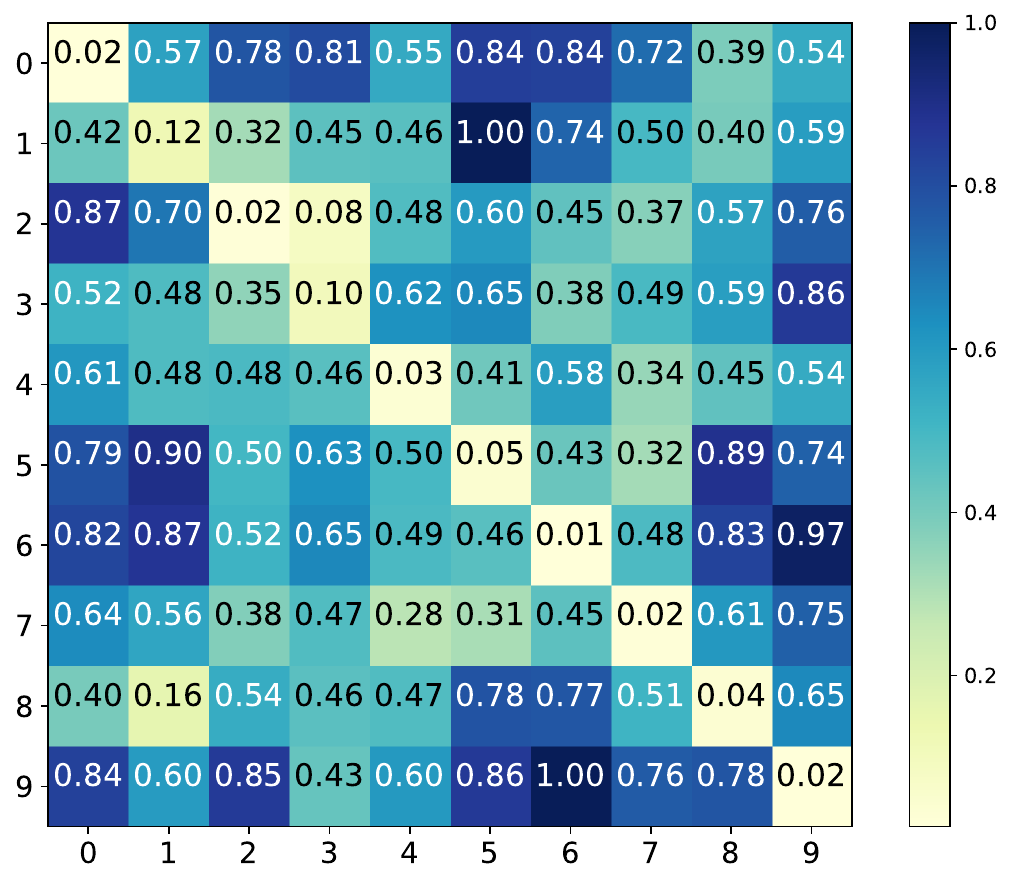}
         \caption{}
         \label{fig:dist_pcm_shape_model}
     \end{subfigure}
     \begin{subfigure}[b]{0.24\textwidth}
         \centering
         \includegraphics[trim={0 0 0 0},clip,width=42.7mm, height= 39.5mm]{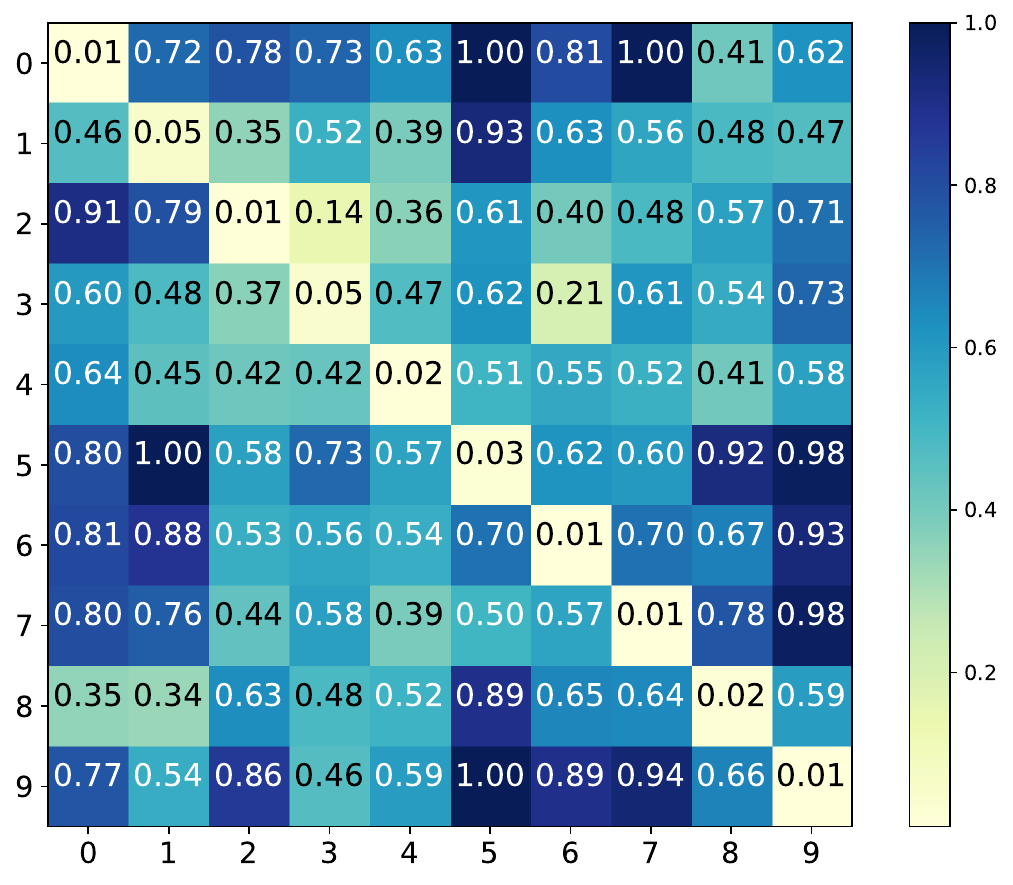}
         \caption{}
         \label{fig:dist_spst_shape_model}
     \end{subfigure}
        \caption{Class-wise MMD for $S\rightarrow M$ for (a) baseline (only PCM w/o adaptation), and (b) our COT with SPST.}
        \label{fig:dist_matrix}
\end{figure}

Our method outperforms all the existing methods with a significant margin on all the combinations of GraspNetPC-10 dataset, as shown in Table \ref{tab:Graspnet-10}. COT beats existing methods in both with and without SPST strategy; also, in some cases, it beats the supervised method (upper bound). It is interesting to note the difference in behaviour of COT and other methods on real-world data in PointDA-10 and GraspNetPC-10. PointDA-10, in general, has a very skewed class-wise sample distribution and has a small set of real-world samples. Whereas, GraspNetPC-10 has almost uniform class-wise sample distribution with approximately double the size of ScanNet. COT performs significantly better with larger datasets and almost equal class-wise sample distribution. Existing methods that propose classification-based \cite{zou2021geometry} or geometry-aware implicit learning-based \cite{Shen_2022_CVPR} tasks fall short in terms of performance boost compared to COT when real-world datasets are large and have uniform class distribution. This shows the effectiveness of COT for unsupervised domain adaptation achieving SoTA performance on real-world data from GraspNet-10 dataset.

\begin{table*}[ht!]
\centering
\tiny
\resizebox{\textwidth}{!}{%
\begin{tabular}{cccc||ccccccc} 
\hline
$\mathcal{L}_{3d}$ & $\mathcal{L}_{ot}$ & $\mathcal{L}_{mm}$ & SPST & M $\rightarrow$ S & M $\rightarrow$ S* & S $\rightarrow$ M & S $\rightarrow$ S* & S* $\rightarrow$ M & S* $\rightarrow$ S & Avg. \\ 
\hline
\Checkmark & \Checkmark &  &  & 82.50 & 53.82 & 74.65 & 47.26 & 75.35 & 71.39 & 67.5\\
\Checkmark &  & \Checkmark &  & 82.66 & 46.64 & \textbf{78.50} & \textbf{53.82} & \textbf{82.24} & 75.40 & 69.9 \\
\Checkmark & \Checkmark & \Checkmark &  & \textbf{83.20} & \textbf{54.61} & \textbf{78.50}  & 53.30 & 79.44 & \textbf{77.41} & \textbf{71.0} \\
\hline
\Checkmark & \Checkmark &  & \Checkmark & \textbf{84.91} & 56.76 & 84.93 & 47.26 & 77.22 & 73.07 & 70.7 \\
\Checkmark &  & \Checkmark & \Checkmark & \textbf{84.91} & 54.32 & 85.51 & \textbf{53.31} & \textbf{86.0} & 75.92 & 73.3 \\
\Checkmark & \Checkmark & \Checkmark & \Checkmark & 84.71 & \textbf{57.66} & \textbf{89.60} & 51.61 & 85.50 & \textbf{77.69} & \textbf{74.4} \\
\hline
\end{tabular}}
\caption{Ablation Study: Target classification accuracy for UDA task on PointDA-10 dataset. Bold represents best results.
}
\label{tab:ablation_study}
\end{table*}

\subsection{Domain Alignment}
In this section, we discuss our used sampling strategy for creating a batch and explain its working in our $\mathcal{L}_{ot}$ loss for domain alignment. For every iteration, we use random sampling to draw source and target batches independently. Note that it does not ensure the coherence of source and target classes in a batch. Using these batches, the OT flow finds the best one-to-one matching amongst both domains using the defined cost function and updates both network's (encoder and classifier) weights to minimize the $\mathcal{L}_{ot}$ loss. Even though we use random sampling we find that repeating this process for multiple iterations eventually converges the overall alignment loss ($\mathcal{L}_{ot}$) giving discriminative features for classes with aligned source and target distributions. For examining the distance between class clusters from the source and target, we compute the maximum mean discrepancy (MMD) between learned point cloud features. In Figure \ref{fig:dist_matrix}, we show class-wise MMD, where Figures \ref{fig:dist_pcm_shape_model}, \ref{fig:dist_spst_shape_model} are for baseline (without adaptation) and our COT respectively on ShapeNet to ModelNet. The diagonal of the matrix represents MMD between the same classes from source and target, and the upper and lower triangular matrices represent MMD between different classes for source and target. It is clearly evident that the MMD matrix for our COT has higher distances in the upper and lower triangular regions than the baseline. This shows that classes within the source and target individually are well separated. Further, the diagonal values for our COT are lower than the baseline without adaptation, indicating that the same classes in source and target are closer for features obtained from our method. Overall, we can see that point cloud embeddings generated by COT have better inter-class distances and source and target class alignment.

\subsection{Discussion: Decision Boundary}
\label{subsec:discussion}
We also examine the decision boundaries of our learned models. Figure \ref{fig:all_decision boundary_1} illustrates the decision boundaries from early (top-row) and final (bottom-row) epochs for four variants of our model. 
For this experiment, we select target samples from the hidden space of our trained models. We consider four variants of our model, i.e., $i)$ only PCM (no adaptation), $ii)$ contrastive learning with PCM, $iii)$ contrastive learning and OT with PCM (our COT method), and $iv)$ our COT fine-tuned with SPST strategy. 
All the representations are retrieved with the labels predicted by our trained model. Next, we fit the SVM and consider a ``one-vs-rest strategy" to visualize the decision boundaries.

From Figures \ref{fig:init_boundary_pcm_only_1} to \ref{fig:init_boundary_pcm_cl_ot_spst_1} and \ref{fig:final_boundary_pcm_only_1} to \ref{fig:final_boundary_pcm_cl_ot_spst_1}, we can clearly interpret that the baseline model with only PCM and no adaptation leads to irregular boundaries in Figures~\ref{fig:init_boundary_pcm_only_1} and \ref{fig:final_boundary_pcm_only_1}. The representations are enhanced, and the boundary becomes smoother by applying contrastive learning to both the domains in Figures~\ref{fig:init_boundary_pcm_cl_1} and \ref{fig:final_boundary_pcm_cl_1}. In contrast, training the model with our COT, which includes the previous two strategies (PCM and contrastive learning) along with OT loss further improves the decision boundaries in Figures~\ref{fig:init_boundary_pcm_cl_ot_1} and \ref{fig:final_boundary_pcm_cl_ot_1}. Finally, with the SPST strategy, which finetunes the COT with pseudo labels of target samples, the region gets even more compact and smoother in Figures~\ref{fig:init_boundary_pcm_cl_ot_spst_1} and \ref{fig:final_boundary_pcm_cl_ot_spst_1}. This shows that contrastive learning separates the two classes which are improved by OT alignment. Also, SPST further makes the classes more compact and achieves the best results.

\subsection{Ablation Studies}
We perform ablation studies to understand the significance of proposed losses in our method. In Table \ref{tab:ablation_study}, we compare the results of our COT trained with various components on PointDA-10. $\mathcal{L}_{3d}$ is always used as it is our base self-learning task for 3D point clouds. The significance of $\mathcal{L}_{ot}$ can be seen by comparing row 3 and row 2. When $\mathcal{L}_{ot}$ is removed from COT, the performance drops on almost all settings. Comparing row 3 and row 1, we can see the effect of $\mathcal{L}_{mm}$ as the performance decreases for all settings when it is turned off. In both cases, the average accuracy also drops. This indicates positive contribution of both $\mathcal{L}_{ot}$ and $\mathcal{L}_{mm}$ in the formulation of our COT. A similar trend is also observed with the SPST strategy as well. Comparing row 6 with rows 4 and 5, we see the best performance when both losses are used. Also, note that SPST increases the performance for all three settings shown. Overall, these results suggest that both image modality and OT-based domain alignment are crucial for achieving the best results.

\section{Conclusion}
In this work, we tackled the domain adaptation problem on 3D point clouds for classification. We introduced a novel methodology to synergize contrastive learning and optimal transport for effective UDA. Our method focuses on reducing the domain shift and learning high-quality transferable point cloud embeddings. Our empirical study reveals the effectiveness of COT as it outperforms existing methods in overall average accuracy on one dataset, and achieves SoTA performance on another. The conducted ablation studies demonstrate the significance of our proposed method. One limitation of our method is that it currently assumes a fixed set of classes in both domains that limit generalizability. 
A key factor for better domain matching would be to improve the cost function used for computing optimal coupling.
Also, an interesting future direction would be to extend our OT-based approach for UDA of point clouds for segmentation or object detection in indoor scenes.

{\small
\bibliographystyle{ieee_fullname}
\bibliography{egbib}
}

\clearpage
\title{\Large\bfseries Appendix}
\maketitle
\appendix

\section{Implementation Details}
In this section, we mention our implementation details for reproducibility purposes. For data pre-processing, we follow \cite{achituve2021self} and align the positive $Z$ axis of all point clouds from the whole PointDA-10 dataset. We use the farthest point sampling algorithm to sample $1024$ points uniformly across the object surface. Further, all the point clouds are normalized and scaled to fit in a unit-sphere. For getting renderings of point clouds from multiple views, we place orthographic cameras in a circular rig. We set the number of views to $12$ and the image size as $224 \times 224$. We set points color to $white$, background color to $black$, points radius to $0.008$, and points per pixel to $2$. For getting two augmented versions of the original point cloud used in self-supervised contrastive learning, we compose spatial transformations picked from random point cloud scaling, rotation, and translation. The original point cloud which is passed to the source classifier is only transformed with random jittering and random rotation about its $Z$ axis.

For a fair comparison with recent works \cite{achituve2021self, Shen_2022_CVPR, zou2021geometry} we use DGCNN \cite{dgcnn2019} as our 3D encoder for extracting global point cloud features. We choose a pre-trained ResNet-50 \cite{he2015deep} as our image feature extractor.
Both 3D and 2D encoders embed their respective modality into a $256$ dimensional feature space for contrastive learning. Whereas, for the classification task, $1024$ dimensional feature vector of the original point cloud is used. We use a 3-layer MLP as our classifier, having $(512, 256, 10)$ neurons respectively.
Please note that for testing the classification performance, we do not use 2D features and only use the global features given by the 3D encoder. We set the temperature parameter $\tau$ used in contrastive losses $\mathcal{L}_{3d}$ and $\mathcal{L}_{mm}$ as $0.1$. To solve the optimization problem of the optimal coupling matrix, we use the POT library \cite{pot}. We do a grid search to find the best $\alpha$ and $\beta$ combination from $\mathcal{L}_{ot}$. For most of the dataset combinations, we set the hyperparameters $\alpha$ and $\beta$ to $0.001$ and $0.0001$, respectively. 

We perform all our experiments on NVIDIA RTX-2080Ti GPUs using the Pytorch framework for implementing our models. We set the batch size to $32$, learning rate as $0.001$ with cosine annealing as the learning rate scheduler and use Adam optimizer. We set weight decay to $0.00005$ and momentum to $0.9$. In total, we train our models for $150$ epochs on PointDA-10 and $120$ epochs on GraspNetPC-10 dataset.

\section{Class-wise Performance Analysis}

\begin{figure*}[h!]
     \centering
     \begin{subfigure}[b]{0.33\textwidth}
         \centering
         \includegraphics[height = 5.8cm, width=\textwidth]{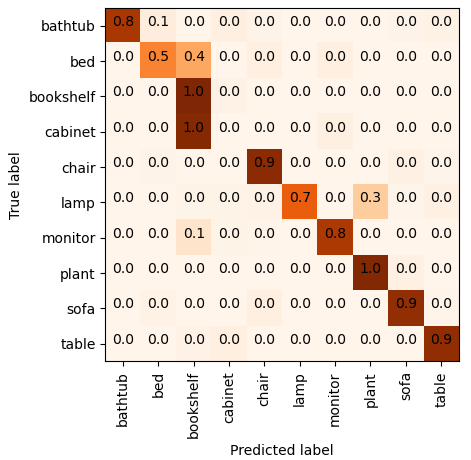}
         \caption{COT with SPST ($M \rightarrow S$)}
         \label{fig:cls_acc_pcm_m_s}
     \end{subfigure}
     \hfill
     \begin{subfigure}[b]{0.33\textwidth}
         \centering
         \includegraphics[height = 5.8cm, width=\textwidth]{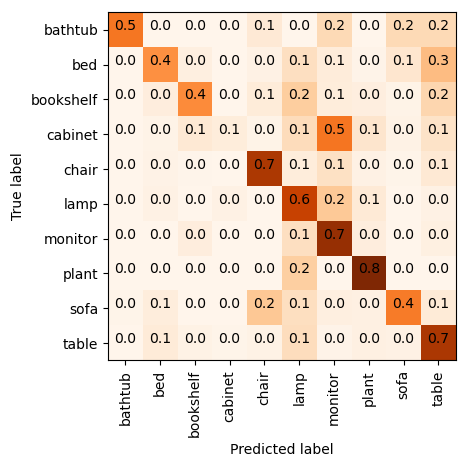}
         \caption{COT with SPST ($M \rightarrow S^{*}$)}
         \label{fig:cls_acc_pcm_m_sc}
     \end{subfigure}
     \hfill
     \centering
      \begin{subfigure}[b]{0.33\textwidth}
          \centering
          \includegraphics[height = 5.8cm, width=\textwidth]{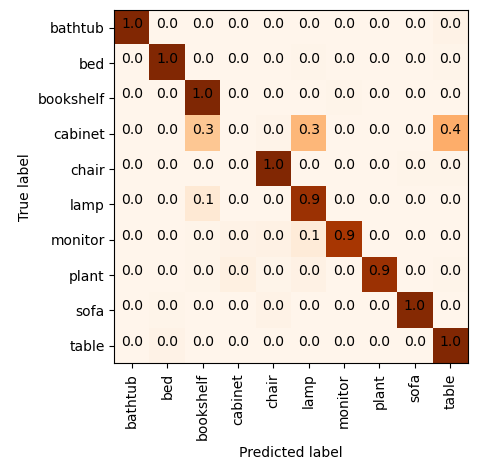}
          \caption{COT with SPST $(S^{*} \rightarrow M)$}
          \label{fig:cls_acc_pcm_sc_m}
      \end{subfigure}
     \\
     \begin{subfigure}[b]{0.33\textwidth}
          \centering
          \includegraphics[height = 5.8cm, width=\textwidth]{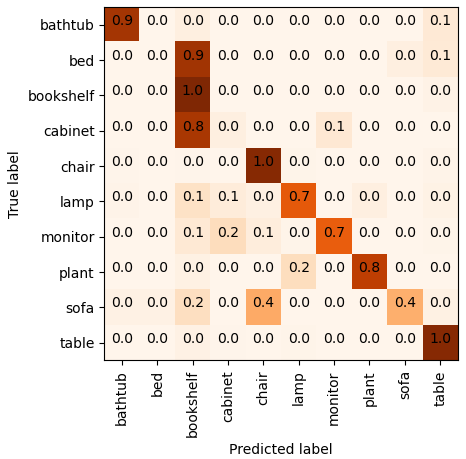}
         \caption{COT with SPST $(S^{*} \rightarrow S)$}
         \label{fig:cls_acc_pcm_sc_s}
      \end{subfigure}
      \hfill
      \begin{subfigure}[b]{0.33\textwidth}
         \centering
         \includegraphics[height = 5.8cm, width=\textwidth]{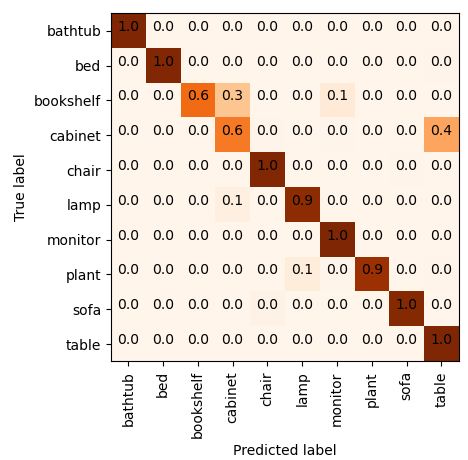}
         \caption{COT with SPST $(S \rightarrow M)$}
         \label{fig:cls_acc_pcm_s_m}
     \end{subfigure}
     \hfill
     \begin{subfigure}[b]{0.33\textwidth}
         \centering
         \includegraphics[height = 5.8cm, width=\textwidth]{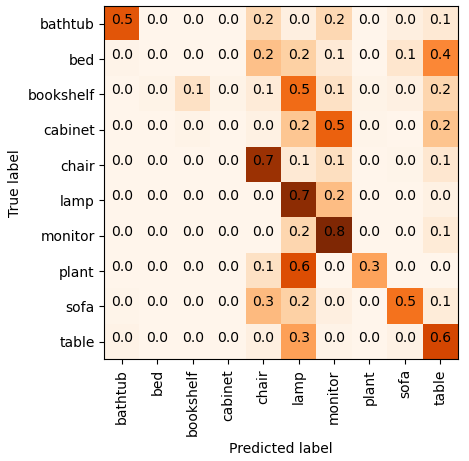}
         \caption{COT with SPST $(S \rightarrow S^{*}$)}
         \label{fig:cls_acc_pcm_s_sc}
     \end{subfigure}
     \caption{Confusion Matrices of Our COT with SPST on PointDA-10 dataset on all Source$\rightarrow$Target experimental settings}
     \label{fig:class_pointda}
\end{figure*}

\begin{figure*}[h!]
     \centering
     \begin{subfigure}[b]{0.45\textwidth}
         \centering
         \includegraphics[width=0.85\textwidth]{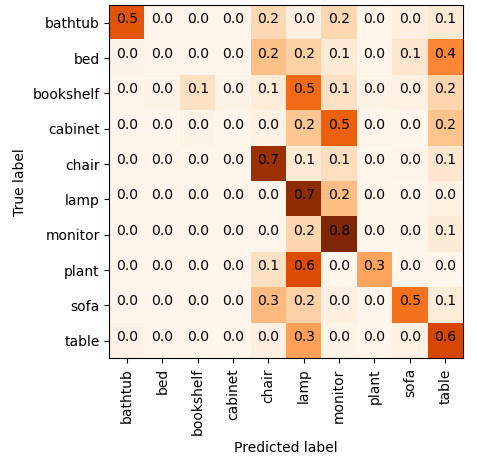}
         \caption{COT with SPST: Syn.$\rightarrow$Kin.}
          \label{fig:cls_acc_syn_kin}
     \end{subfigure}
     \begin{subfigure}[b]{0.45\textwidth}
         \centering
         \includegraphics[width=0.85\textwidth]{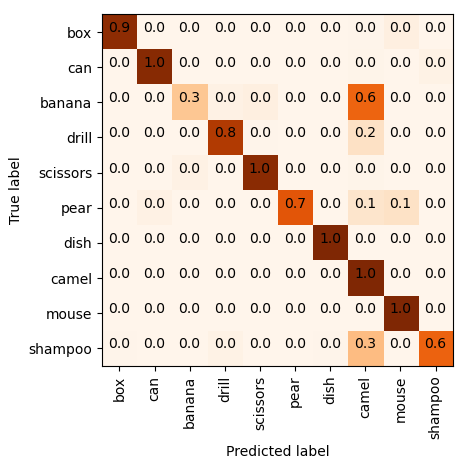}
         \caption{COT with SPST: Syn.$\rightarrow$RS}
         \label{fig:cls_acc_syn_rs}
     \end{subfigure}
     \\
     \begin{subfigure}[b]{0.45\textwidth}
         \centering
         \includegraphics[width=0.85\textwidth]{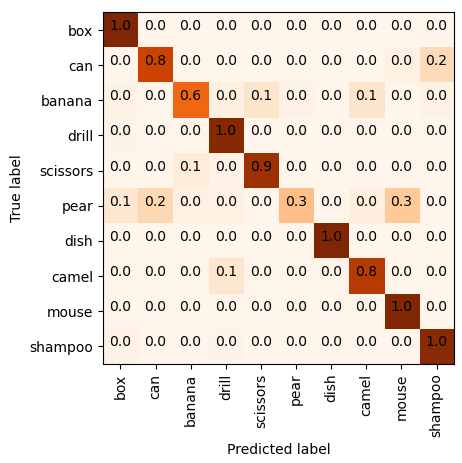}
         \caption{COT with SPST: Kin.$\rightarrow$Rs.}
         \label{fig:cls_acc_kin_rs}
     \end{subfigure}
     \begin{subfigure}[b]{0.45\textwidth}
          \centering
          \includegraphics[width=0.85\textwidth]{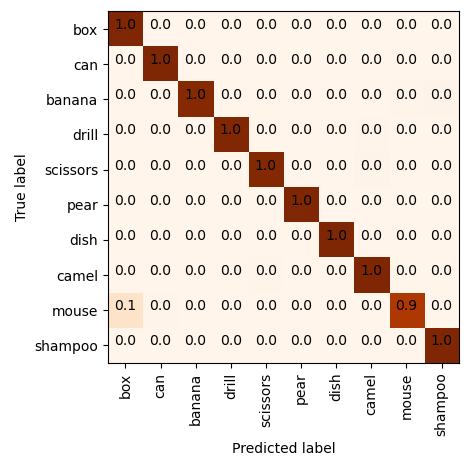}
         \caption{COT with SPST: RS$\rightarrow$Kin.}
         \label{fig:cls_acc_rs_kin}
      \end{subfigure}
     \caption{Confusion Matrices of Our COT with SPST on GraspNetPC-10 dataset on all Source$\rightarrow$Target experimental settings}
     \label{fig:class_graspnet}
\end{figure*}

In this section, we analyse the class-wise accuracy of COT on PointDA-10 and and GraspNetPC-10 datasets. Results for PointDA-10 and GraspNetPC-10 are show  in Tables \ref{tab:class-wise-1} and \ref{tab:class-wise-2}, respectively and the confusion matrices are shown in Figure \ref{fig:class_pointda} for Point DA-10 dataset and in Figure \ref{fig:class_graspnet} for GraspNetPC-10 dataset.
\begin{table}[h!]
\resizebox{\linewidth}{!}{\begin{tabular}{@{}l|cccccccccc@{}}
\toprule
                   & Bathtub & Bed  & Bookshelf & Cabinet & Chair & Lamp & Monitor & Plant & Sofa & Table \\ \midrule
S*$\rightarrow$ M & 0.98    & 0.98 & 0.99      & 0       & 0.98  & 0.9  & 0.86    & 0.9   & 0.97 & 0.98  \\
S*$\rightarrow$ S  & 0.86    & 0    & 0.98      & 0.05    & 0.96  & 0.67 & 0.65    & 0.8   & 0.36 & 0.95    \\
M$\rightarrow$ S  & 0.85    & 0.52 & 0.98      & 0       & 0.94  & 0.65 & 0.84    & 0.97  & 0.92 & 0.93  \\
M$\rightarrow$ S* & 0.46    & 0.39 & 0.4       & 0.05    & 0.69  & 0.63 & 0.74    & 0.8   & 0.45 & 0.69  \\
S$\rightarrow$ S* & 0.54    & 0    & 0.12      & 0       & 0.7   & 0.73 & 0.77    & 0.32  & 0.45 & 0.58  \\
S$\rightarrow$ M  & 1       & 0.99 & 0.62      & 0.57    & 0.99  & 0.95 & 1       & 0.91  & 0.98 & 1     \\ \bottomrule
\end{tabular}}
\caption{Class-wise accuracies of our COT (with SPST) on the PointDA-10 dataset}
\label{tab:class-wise-1}
\end{table}

In the case of PointDA-10, in almost all the cases, the cabinet class is the toughest to classify. In some of the combinations even a single example from this class is not classified correctly. In case of GraspNetPC-10 all the samples belonging to the class Dish are always classified correctly.

\begin{table}[h!]
\resizebox{\linewidth}{!}{\begin{tabular}{@{}l|cccccccccc@{}}
\toprule
                      & Box  & Can  & Banana & Drill & Scissors & Pear & Dish & Camer & Mouse & Shampoo \\ \midrule
Syn$\rightarrow$ Kin & 1    & 1    & 0.98   & 0.99  & 0.92     & 1    & 1    & 1     & 1     & 1       \\
Syn$\rightarrow$ RS  & 0.95 & 0.97 & 0.28   & 0.84  & 0.96     & 0.69 & 1    & 1     & 1     & 0.65    \\
Kin$\rightarrow$ RS  & 1    & 0.78 & 0.63   & 0.98  & 0.9      & 0.31 & 1    & 0.83  & 0.99  & 0.96    \\
Rs$\rightarrow$ Kin  & 1    & 1    & 0.98   & 0.99  & 0.98     & 1    & 1    & 1     & 0.85  & 1       \\ \bottomrule
\end{tabular}}
\caption{Class-wise accuracies of our COT (with SPST) on the GraspNetPC-10 dataset}
\label{tab:class-wise-2}
\end{table}

\section{Domain Alignment Analysis}
\begin{figure*}[h!]
     \centering
     \begin{subfigure}[b]{0.45\textwidth}
         \centering
         \includegraphics[trim={0 0 2cm 0},clip,width=0.76\textwidth]{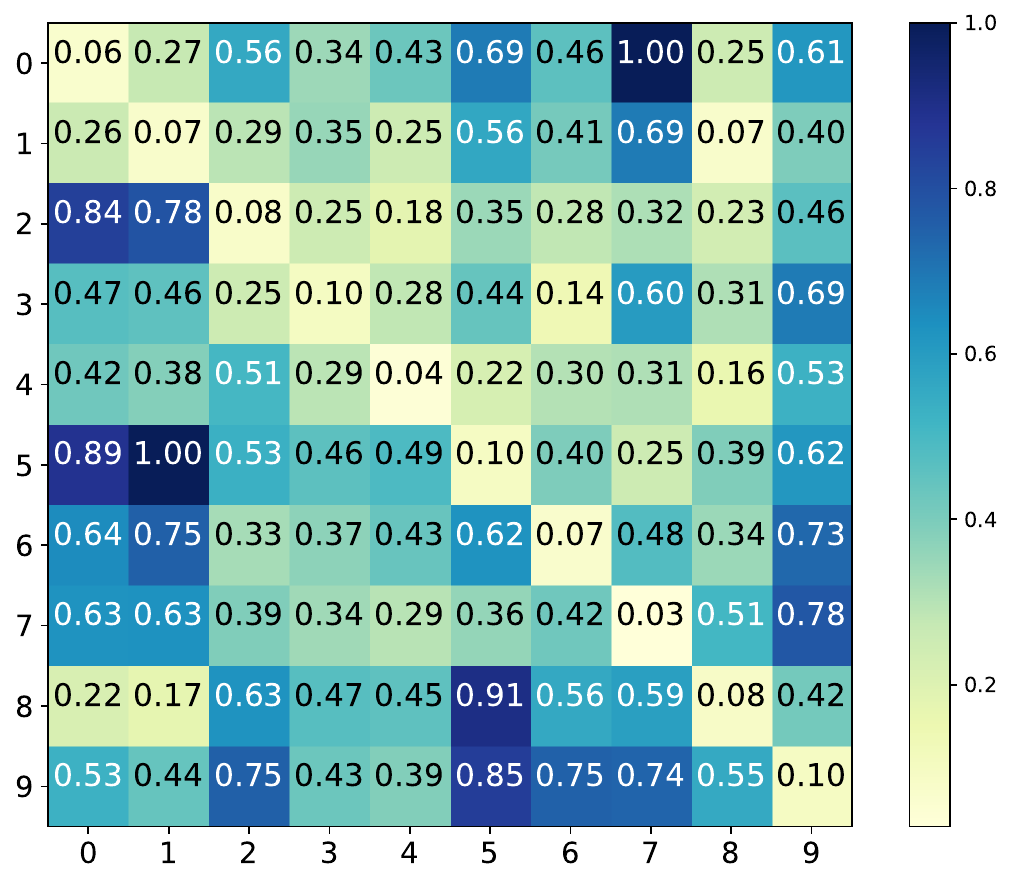}
         \caption{Baseline (PCM without Adaptation): $S^{\ast} \rightarrow M$}
         \label{fig:mmd_pcm_scan_model}
     \end{subfigure}
     \begin{subfigure}[b]{0.45\textwidth}
         \centering
         \includegraphics[trim={0 0 2cm 0},clip,width=0.76\textwidth]{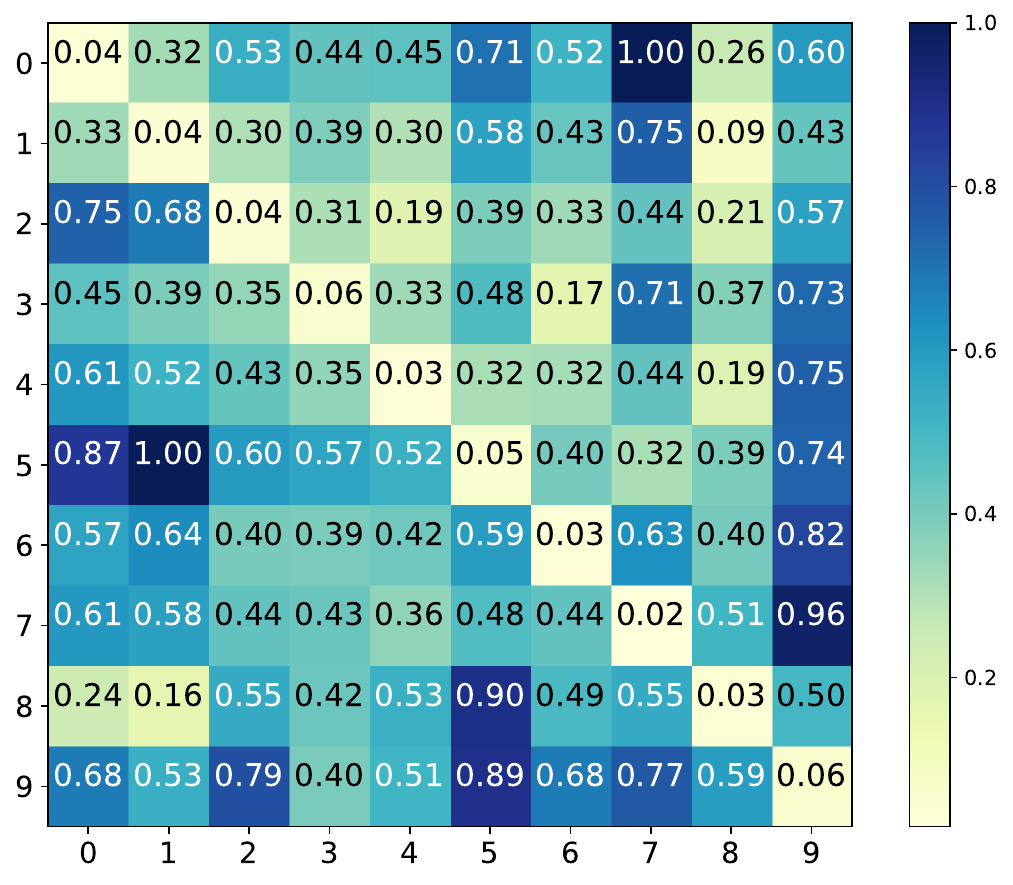}
         \caption{COT with SPST: $S^{\ast} \rightarrow M$}
        \label{fig:mmd_spst_scan_model}
     \end{subfigure}
    \caption{Class-wise MMD plots for Baseline (PCM without Adaptation) and Our COT with SPST for $S^{\ast} \rightarrow M$.}
     \label{fig:mmd_pointda_scan_model}
\end{figure*}

\begin{figure*}[h!]
    \centering
     \begin{subfigure}[b]{0.45\textwidth}
         \centering
         \includegraphics[trim={0 0 2cm 0},clip,width=0.76\textwidth]{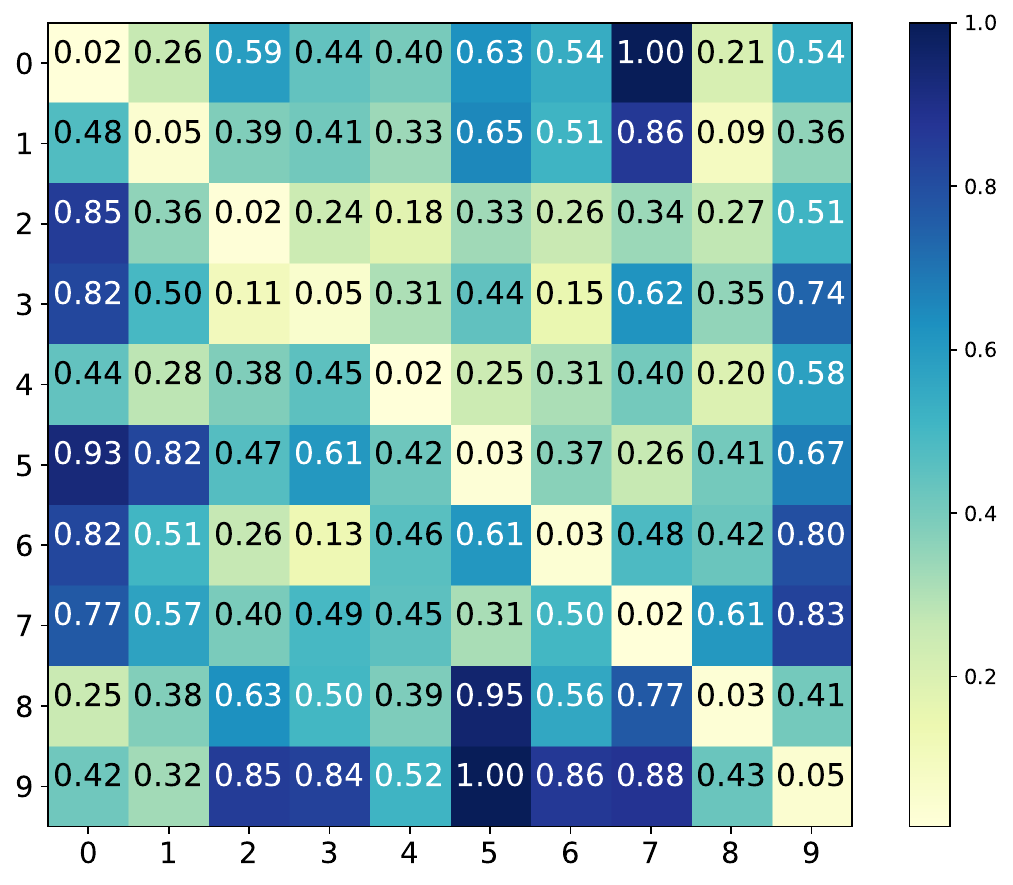}
         \caption{Baseline (PCM without Adaptation): $S^{\ast} \rightarrow S$}
         \label{fig:mmd_pcm_scan_shape}
     \end{subfigure}
     \begin{subfigure}[b]{0.45\textwidth}
         \centering
         \includegraphics[trim={0 0 2cm 0},clip,width=0.76\textwidth]{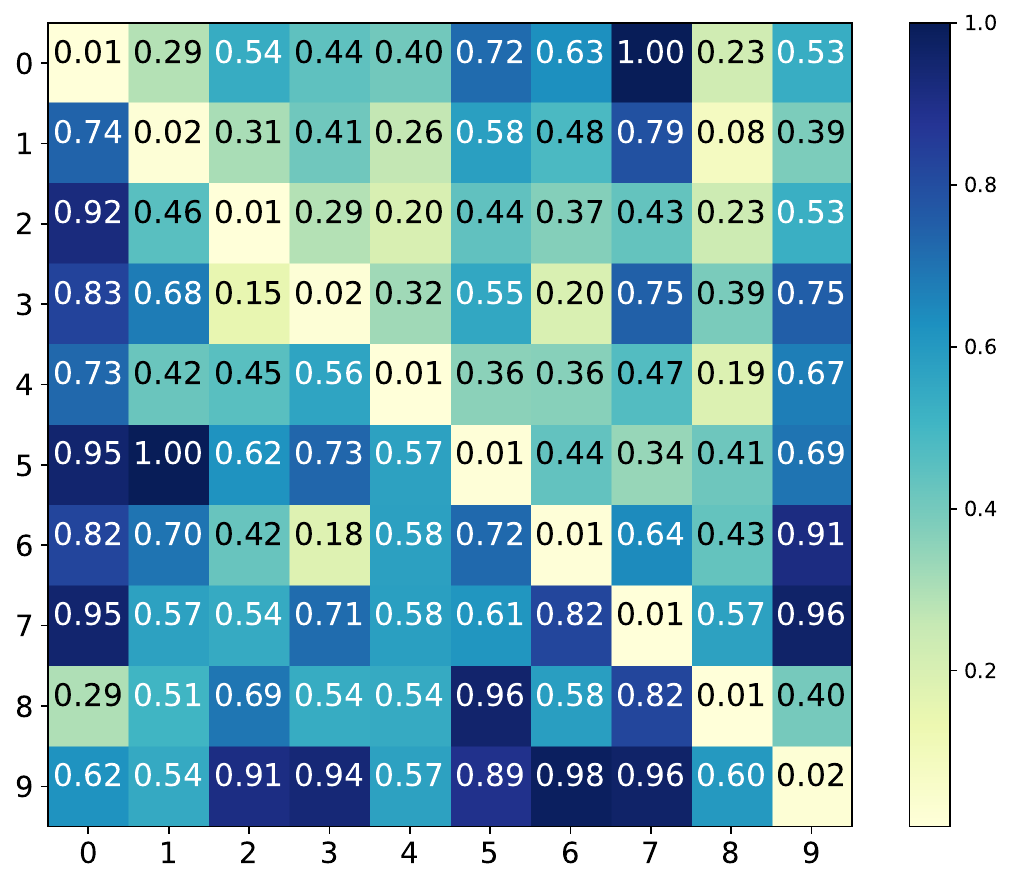}
         \caption{COT with SPST: $S^{\ast} \rightarrow S$}
        \label{fig:mmd_spst_scan_shape}
     \end{subfigure}
    \caption{Class-wise MMD plots for Baseline (PCM without Adaptation) and Our COT with SPST for $S^{\ast} \rightarrow S$.}
     \label{fig:mmd_pointda_scan_shape}
\end{figure*}

\begin{figure*}[h!]
    \centering
     \begin{subfigure}[b]{0.45\textwidth}
         \centering
         \includegraphics[trim={0 0 2cm 0},clip,width=0.76\textwidth]{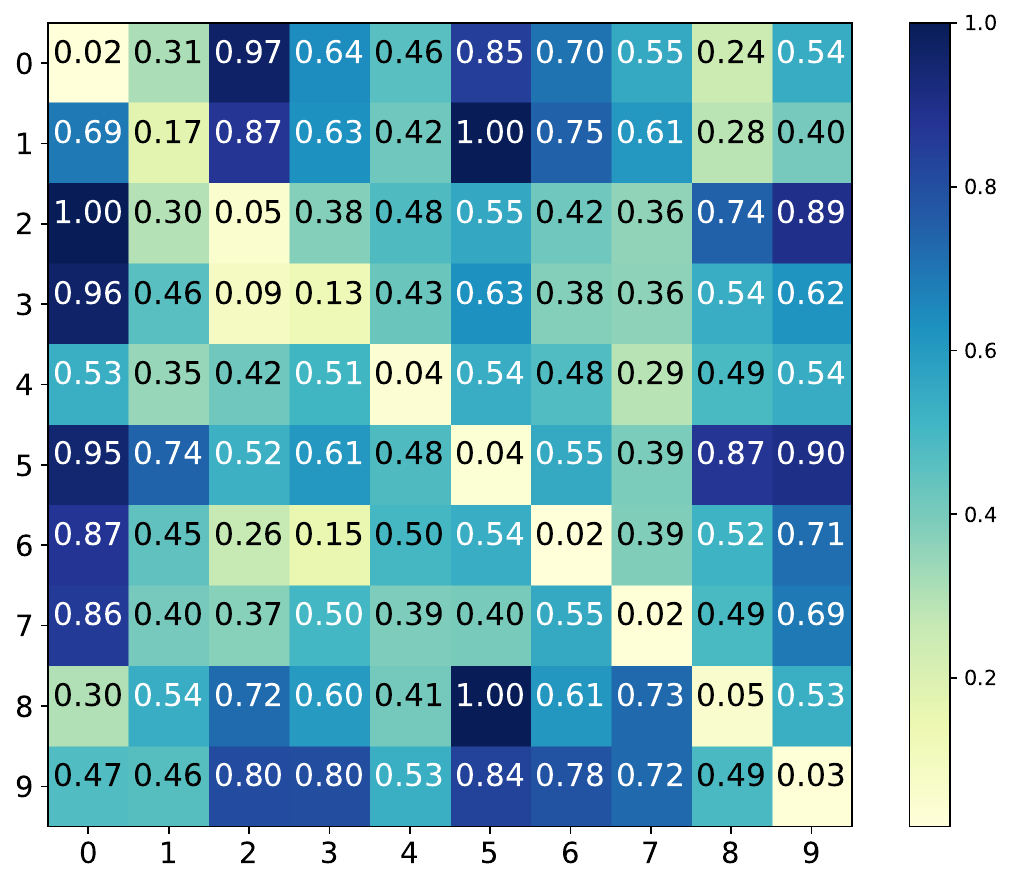}
         \caption{Baseline (PCM without Adaptation): $M \rightarrow S$}
         \label{fig:mmd_pcm_model_shape}
     \end{subfigure}
     \begin{subfigure}[b]{0.45\textwidth}
         \centering
         \includegraphics[trim={0 0 2cm 0},clip,width=0.76\textwidth]{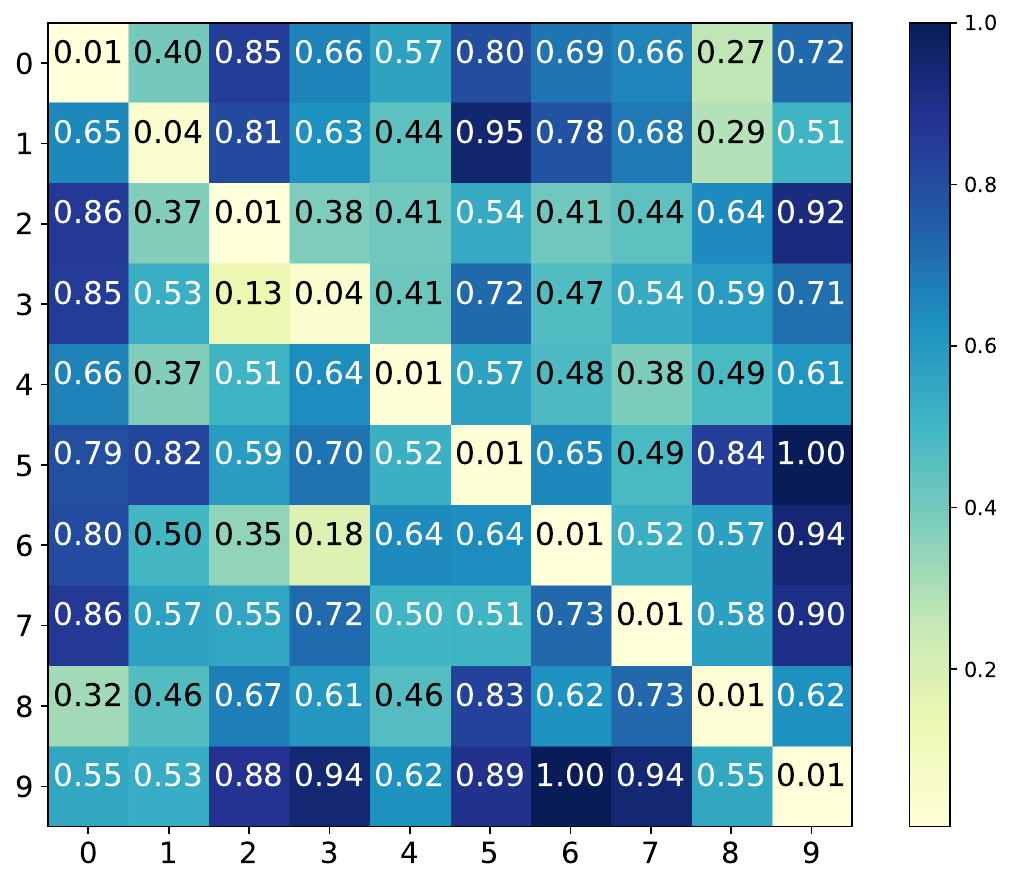}
         \caption{COT with SPST: $M \rightarrow S$}
        \label{fig:mmd_spst_model_shape}
     \end{subfigure}
     \caption{Class-wise MMD plots for Baseline (PCM without Adaptation) and Our COT with SPST for $M \rightarrow S$.}
     \label{fig:mmd_pointda_model_shape}
\end{figure*}

\begin{figure*}[t]
     \centering
     \begin{subfigure}[b]{0.45\textwidth}
         \centering
         \includegraphics[trim={0 0 2cm 0},clip,width=0.95\textwidth]{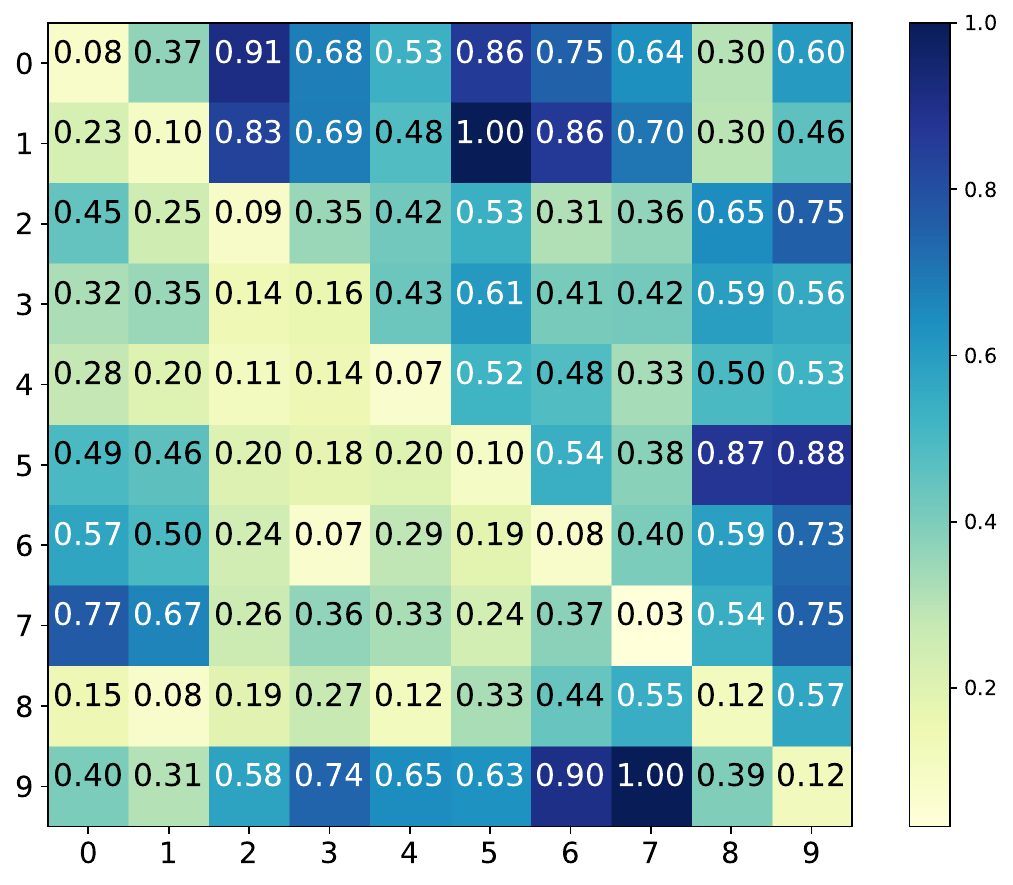}
         \caption{Baseline (PCM without Adaptation): $M \rightarrow S^{\ast}$}
         \label{fig:mmd_pcm_model_scan}
     \end{subfigure}
     \begin{subfigure}[b]{0.45\textwidth}
         \centering
         \includegraphics[trim={0 0 2cm 0},clip,width=0.95\textwidth]{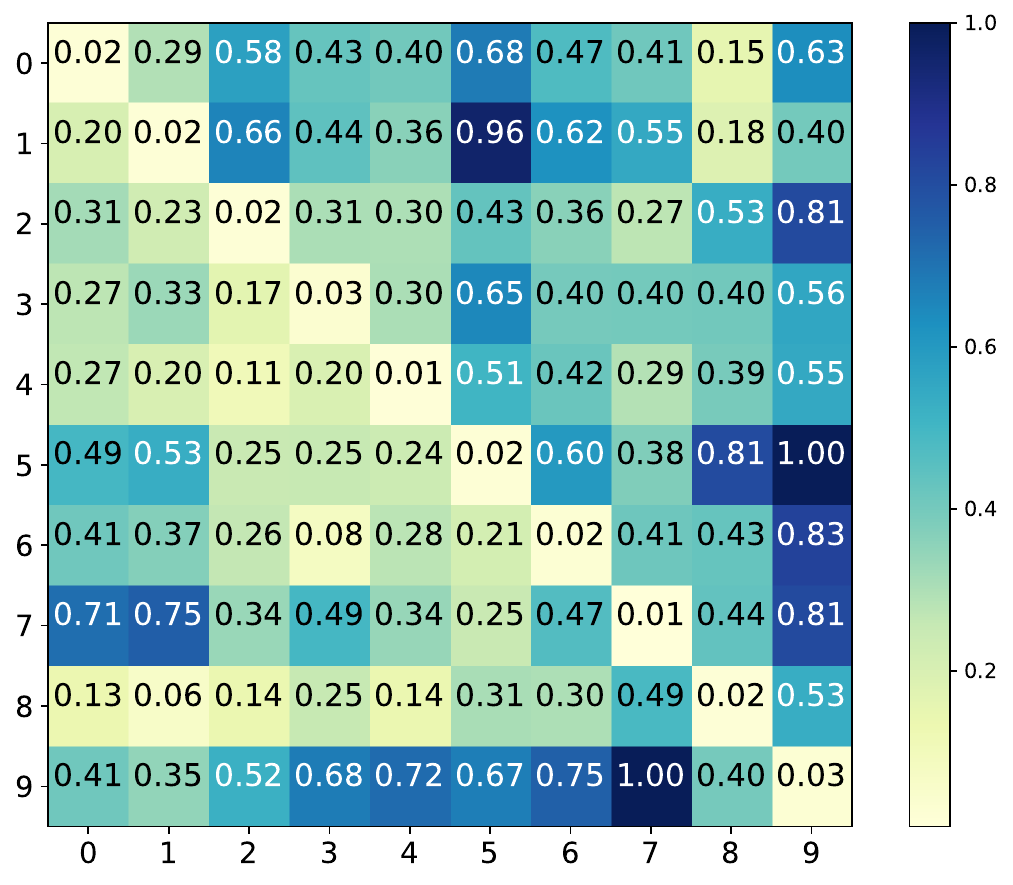}
         \caption{COT with SPST: $M \rightarrow S^{\ast}$}
        \label{fig:mmd_spst_model_scan}
     \end{subfigure}
    \caption{Class-wise MMD plots for Baseline (PCM without Adaptation) and Our COT with SPST for $M \rightarrow S^{\ast}$.}
     \label{fig:mmd_pointda_model_scan}
\end{figure*}

\begin{figure*}[t]
     \centering
     \begin{subfigure}[b]{0.45\textwidth}
         \centering
         \includegraphics[trim={0 0 2cm 0},clip,width=0.95\textwidth]{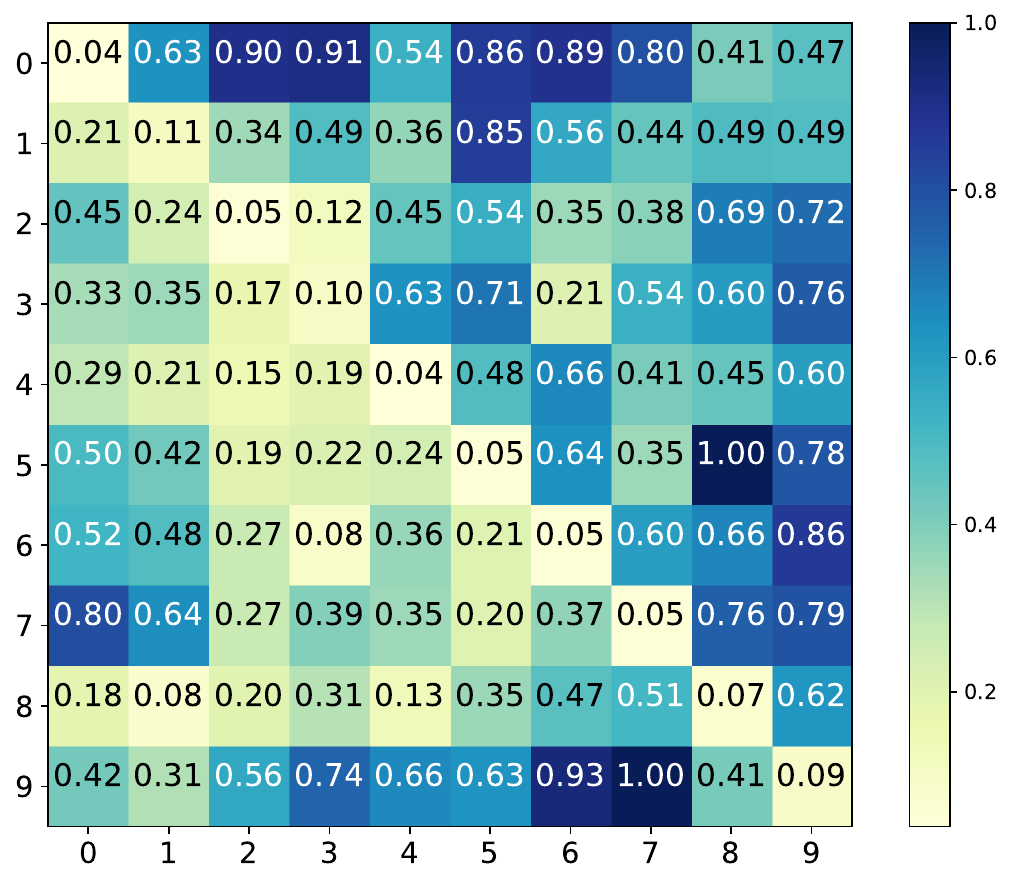}
         \caption{Baseline (PCM without Adaptation): $S \rightarrow S^{\ast}$}
         \label{fig:mmd_pcm_shape_scan}
     \end{subfigure}
     \begin{subfigure}[b]{0.45\textwidth}
         \centering
         \includegraphics[trim={0 0 2cm 0},clip,width=0.95\textwidth]{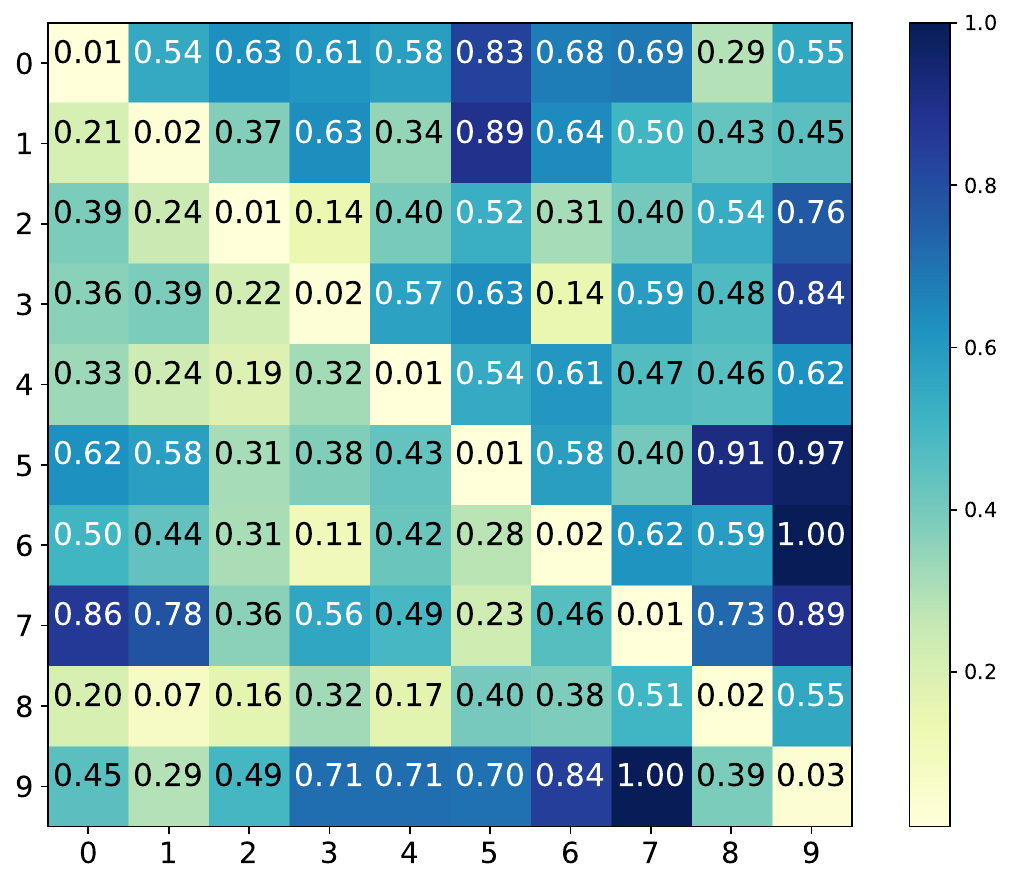}
         \caption{COT with SPST: $S \rightarrow S^{\ast}$}
        \label{fig:mmd_spst_shape_scan}
     \end{subfigure}
     \caption{Class-wise MMD plots for Baseline (PCM without Adaptation) and Our COT with SPST for $S \rightarrow S^{\ast}$.}
     \label{fig:mmd_pointda_shape_scan}
\end{figure*}

\begin{figure*}[h!]
     \centering
     \begin{subfigure}[b]{0.45\textwidth}
         \centering
         \includegraphics[trim={0 0 2cm 0},clip,width=0.95\textwidth]{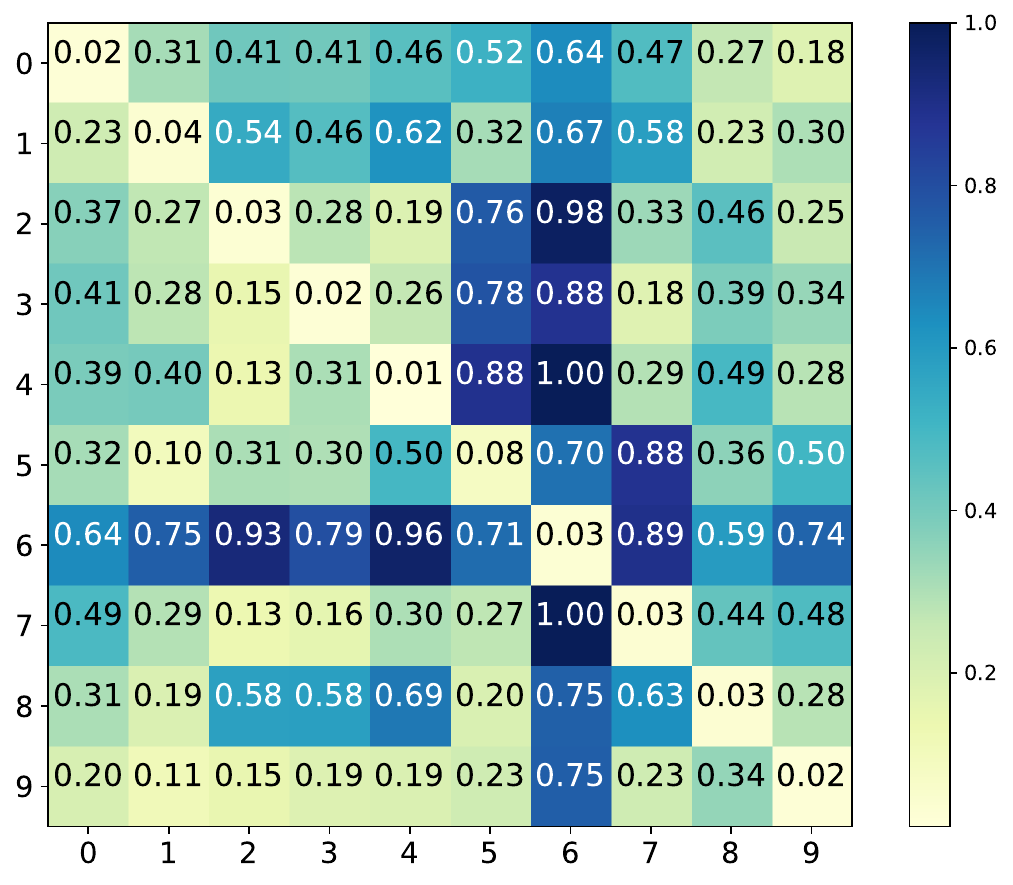}
         \caption{Baseline (PCM without Adaptation): Kin.$\rightarrow$ RS}
         \label{fig:mmd_pcm_kin_real}
     \end{subfigure}
     \begin{subfigure}[b]{0.45\textwidth}
         \centering
         \includegraphics[trim={0 0 2cm 0},clip,width=0.95\textwidth]{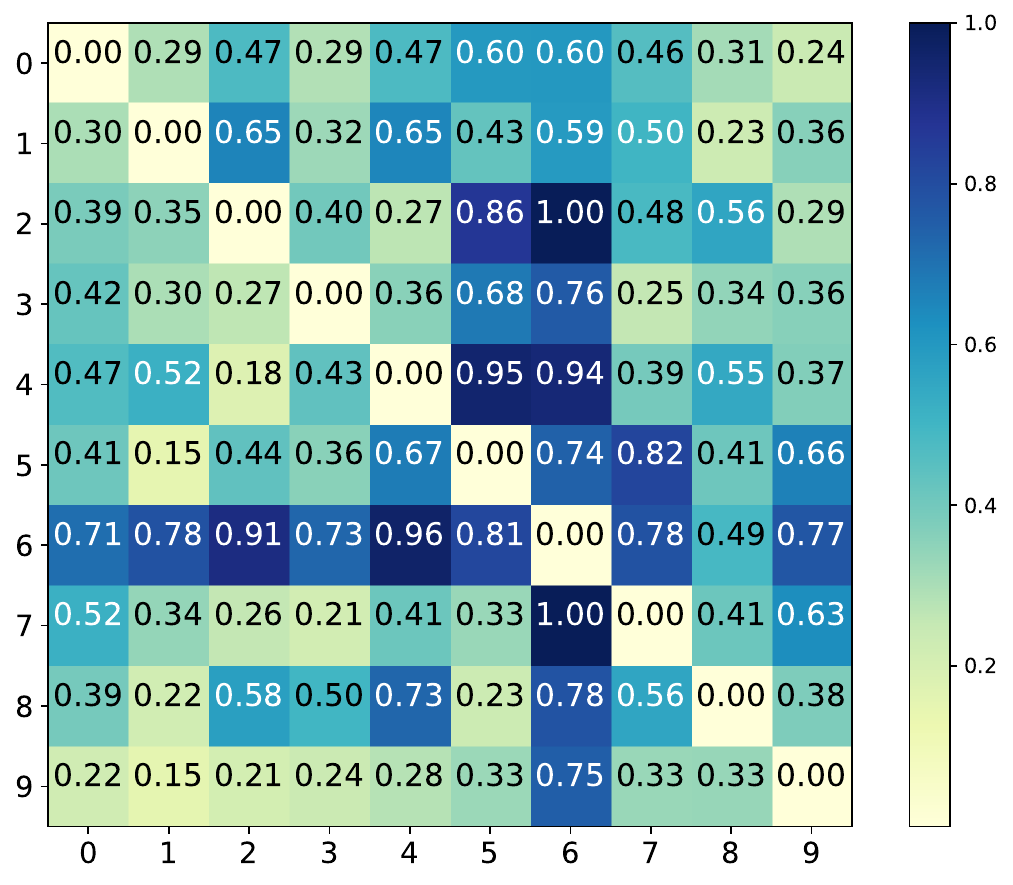}
         \caption{COT with SPST: Kin.$\rightarrow$ RS}
         \label{fig:mmd_spst_kin_real}
     \end{subfigure}
     \\
     \centering
     \begin{subfigure}[b]{0.45\textwidth}
         \centering
         \includegraphics[trim={0 0 2cm 0},clip,width=0.95\textwidth]{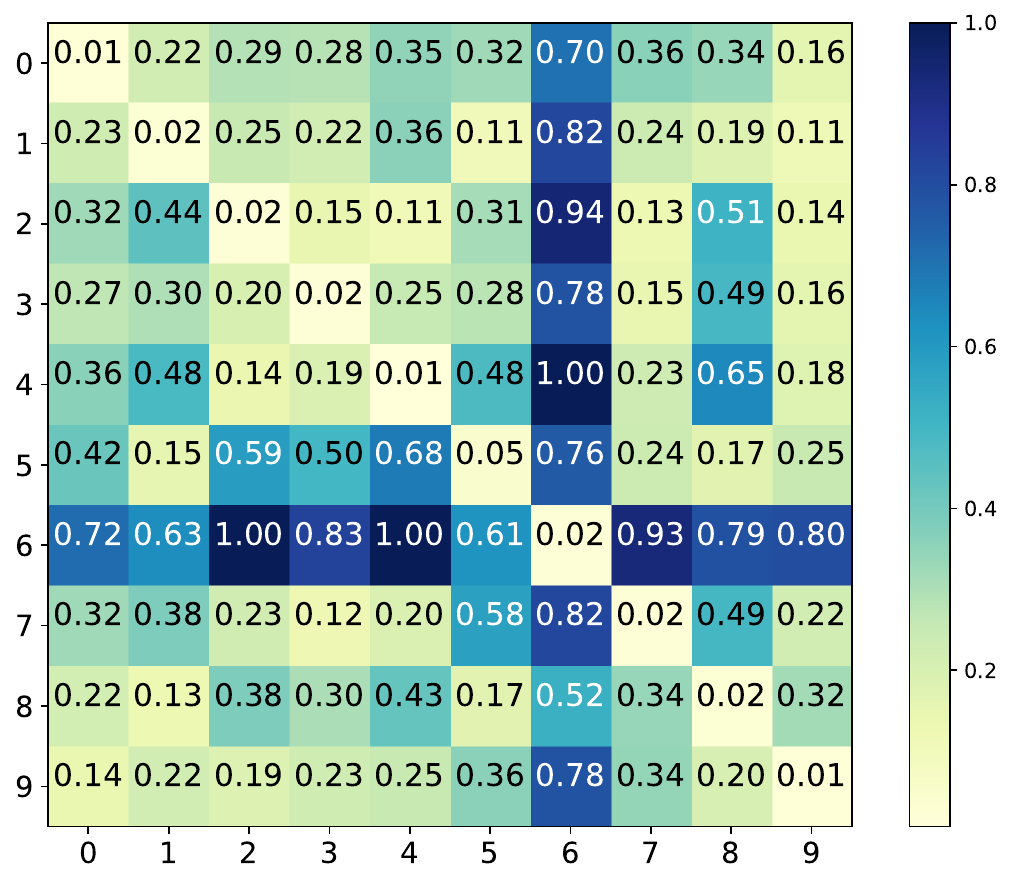}
         \caption{Baseline (PCM without Adaptation): RS.$\rightarrow$ Kin}
         \label{fig:mmd_pcm_real_kin}
     \end{subfigure}
     \begin{subfigure}[b]{0.45\textwidth}
         \centering
         \includegraphics[trim={0 0 2cm 0},clip,width=0.95\textwidth]{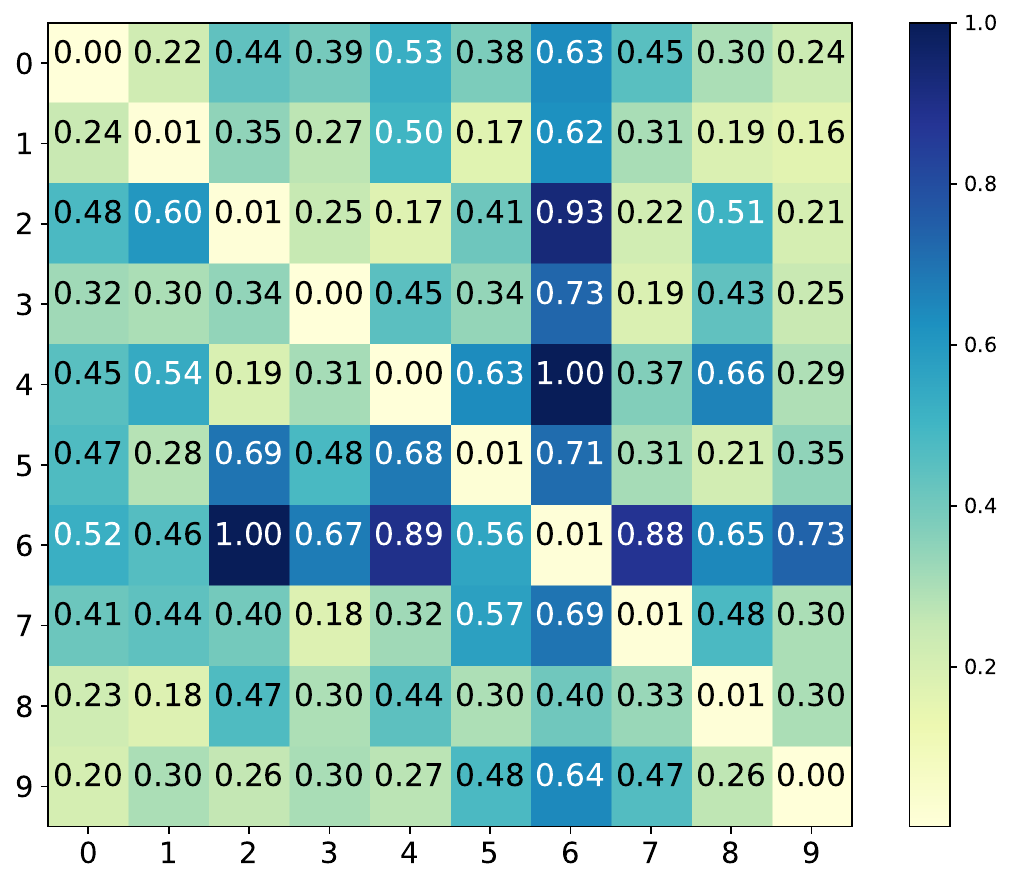}
         \caption{COT with SPST: RS.$\rightarrow$ Kin}
         \label{fig:mmd_spst_real_kin}
     \end{subfigure}
    \caption{Class-wise MMD on GraspNetPC-10 dataset for (a), (c) baseline (only PCM without adaptation), and (b), (d) our COT with SPST for two experimental settings (Kin$\rightarrow$RS and RS$\rightarrow$Kin.)}
    \label{fig:grasp_mmd}
\end{figure*}

We compute and plot Maximum-Mean-Discrepancy (MMD) between learned point cloud features for the rest of the dataset combinations. Figures \ref{fig:mmd_pointda_scan_model}, \ref{fig:mmd_pointda_scan_shape}, \ref{fig:mmd_pointda_model_shape}, \ref{fig:mmd_pointda_model_scan}, \ref{fig:mmd_pointda_shape_scan} contain MMD plots for all source-target dataset combinations from PointDA-10. Figure \ref{fig:grasp_mmd} contains MMD plots for two experimental settings (Kin.$\rightarrow$RS. and  RS.$\rightarrow$Kin.) from GraspNetPC-10.

We can observe that the diagonals have lower values for our method, which indicates better class alignment across source and target. Also in the plot from our method, the upper and lower triangular matrices have higher values than without adaptation ones, which indicates better inter-class distance between source and target classes individually.

\section{Decision Boundary Analysis}
\begin{figure*}[h!]
     \centering
     \begin{subfigure}[b]{0.24\textwidth}
         \centering
         \includegraphics[width=0.84\textwidth]{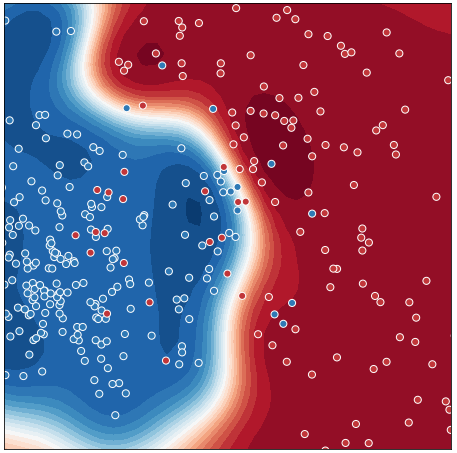}
         \caption{Baseline: (M$\rightarrow$S)}
         \label{fig:pointda_pcm_m_s}
     \end{subfigure}
     \hfill
     \centering
     \begin{subfigure}[b]{0.24\textwidth}
         \centering
         \includegraphics[width=0.84\textwidth]{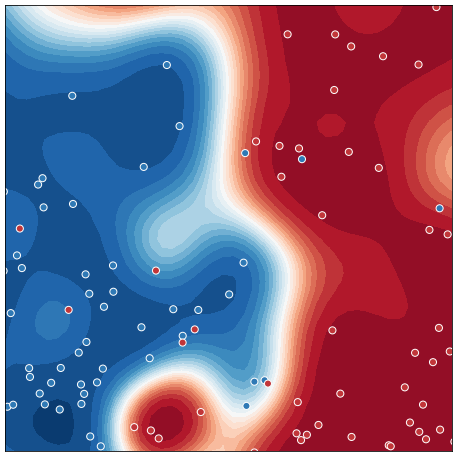}
         \caption{Baseline+CL:(M$\rightarrow$S)}
         \label{fig:pointda_pcm_cl_m_s}
     \end{subfigure}
     \hfill
     \centering
     \begin{subfigure}[b]{0.24\textwidth}
         \centering
         \includegraphics[width=0.84\textwidth]{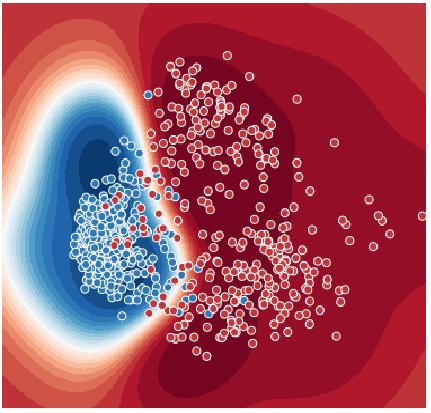}
         \caption{COT: (M$\rightarrow$S)}
         \label{fig:pointda_pcm_ot_m_s}
     \end{subfigure}
     \hfill
     \begin{subfigure}[b]{0.24\textwidth}
         \centering
         \includegraphics[width=0.84\textwidth]{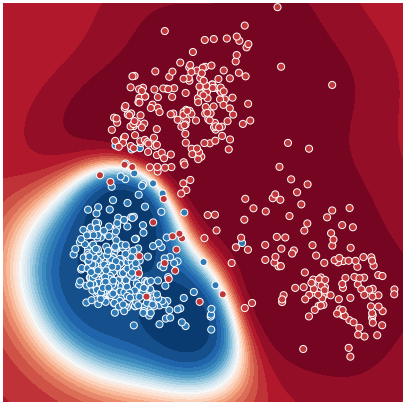}
         \caption{COT+SPST: (M$\rightarrow$S)}
         \label{fig:pointda_spst_m_s}
     \end{subfigure}
     \\
     \begin{subfigure}[b]{0.24\textwidth}
         \centering
         \includegraphics[width=0.84\textwidth]{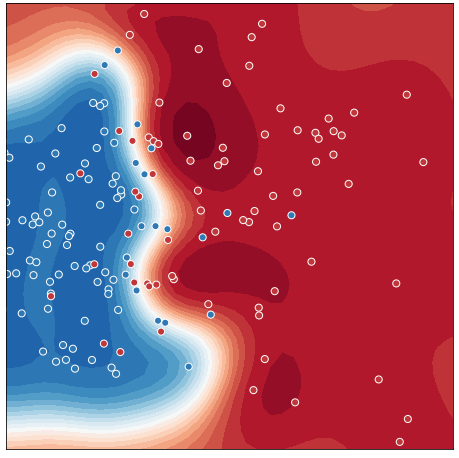}
         \caption{Baseline: (M$\rightarrow$S$^{\ast}$)}
         \label{fig:pointda_pcm_m_sc}
     \end{subfigure}
     \hfill
     \begin{subfigure}[b]{0.24\textwidth}
         \centering
         \includegraphics[width=0.84\textwidth]{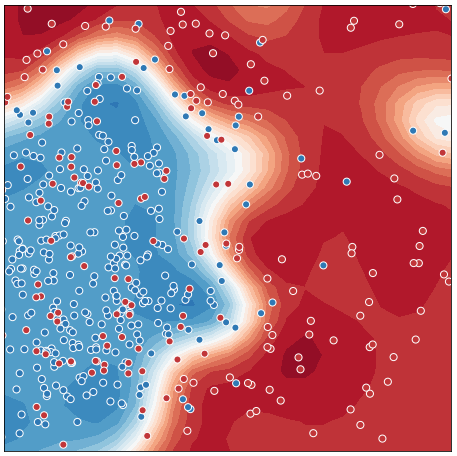}
         \caption{Baseline+CL: (M$\rightarrow$S$^{\ast}$)}
         \label{fig:pointda_pcm_cl_m_sc}
     \end{subfigure}
     \hfill
     \begin{subfigure}[b]{0.24\textwidth}
         \centering
         \includegraphics[width=0.84\textwidth]{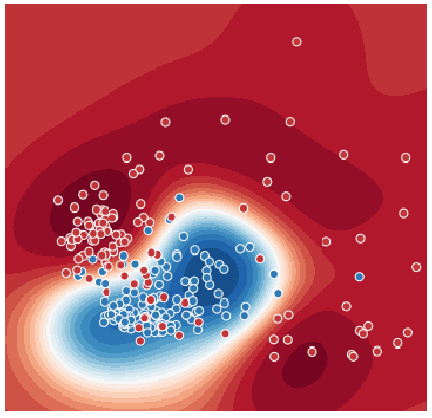}
         \caption{COT: (M$\rightarrow$S$^{\ast}$)}
         \label{fig:pointda_pcm_ot_m_sc}
     \end{subfigure}
     \hfill
     \begin{subfigure}[b]{0.24\textwidth}
         \centering
         \includegraphics[width=0.84\textwidth]{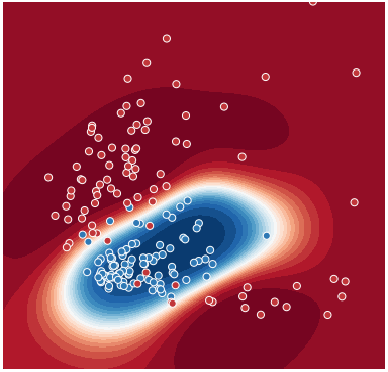}
         \caption{COT+SPST: (M$\rightarrow$S$^{\ast}$)}
         \label{fig:pointda_spst_sc}
     \end{subfigure}
     \\
     \begin{subfigure}[b]{0.24\textwidth}
         \centering
         \includegraphics[width=0.84\textwidth]{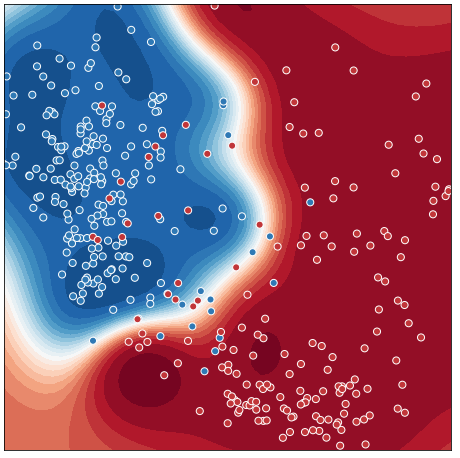}
         \caption{Baseline: (S$\rightarrow$S$^{\ast}$)}
         \label{fig:pointda_pcm_s_sc}
     \end{subfigure}
     \hfill
     \centering
     \begin{subfigure}[b]{0.24\textwidth}
         \centering
         \includegraphics[width=0.84\textwidth]{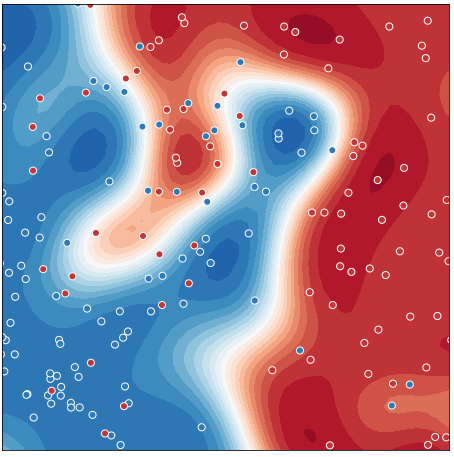}
         \caption{Baseline+CL:(S$\rightarrow$S$^{\ast}$)}
         \label{fig:pointda_pcm_cl_s_sc}
     \end{subfigure}
     \hfill
     \centering
     \begin{subfigure}[b]{0.24\textwidth}
         \centering
         \includegraphics[width=0.84\textwidth]{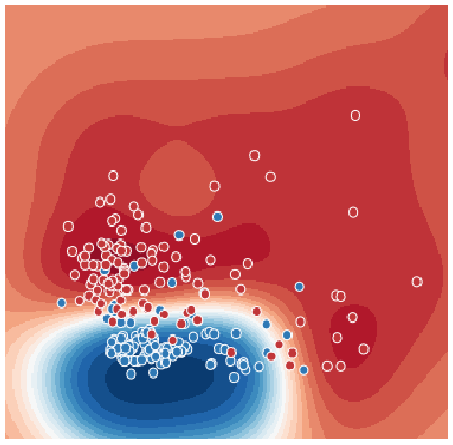}
         \caption{COT: (S$\rightarrow$S$^{\ast}$)}
         \label{fig:pointda_pcm_ot_s_sc}
     \end{subfigure}
     \hfill
    \centering
     \begin{subfigure}[b]{0.24\textwidth}
         \centering
         \includegraphics[width=0.84\textwidth]{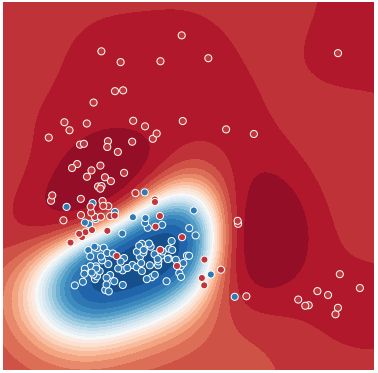}
         \caption{COT+SPST: (S$\rightarrow$S$^{\ast}$)}
         \label{fig:pointda_spst_s_sc}
     \end{subfigure}
     \\
     \begin{subfigure}[b]{0.24\textwidth}
         \centering
         \includegraphics[width=0.84\textwidth]{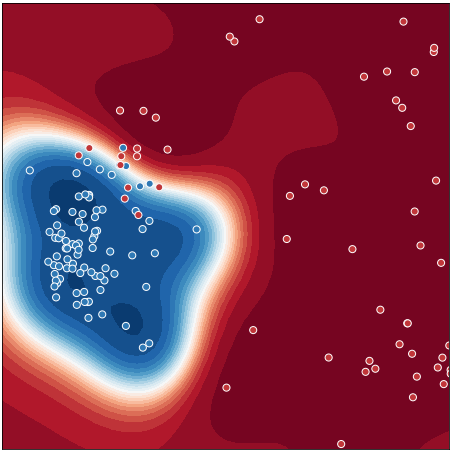}
         \caption{Baseline: (S$^{\ast}$$\rightarrow$M)}
         \label{fig:pointda_pcm_sc_m}
     \end{subfigure}
     \hfill
     \begin{subfigure}[b]{0.24\textwidth}
         \centering
         \includegraphics[width=0.84\textwidth]{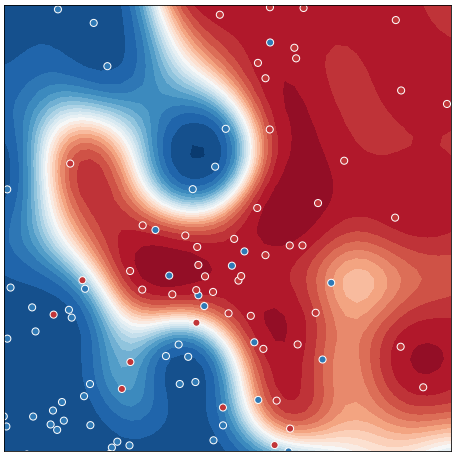}
         \caption{Baseline+CL:(S$^{\ast}$ $\rightarrow$M)}
         \label{fig:pointda_pcm_cl_sc_m}
     \end{subfigure}
     \hfill
     \begin{subfigure}[b]{0.24\textwidth}
         \centering
         \includegraphics[width=0.84\textwidth]{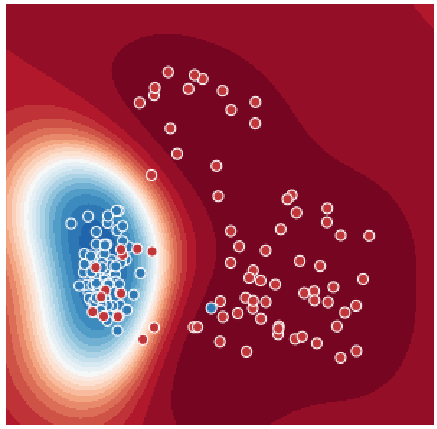}
         \caption{COT: (S$^{\ast}$ $\rightarrow$M)}
         \label{fig:pointda_pcm_ot_sc_m}
     \end{subfigure}
     \hfill
     \begin{subfigure}[b]{0.24\textwidth}
         \centering
         \includegraphics[width=0.84\textwidth]{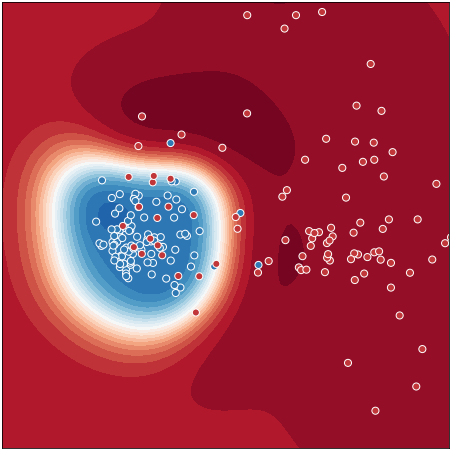}
         \caption{COT+SPST: (S$^{\ast}$$\rightarrow$M)}
         \label{fig:pointda_spst_sc_m}
     \end{subfigure}
     \\
     \begin{subfigure}[b]{0.24\textwidth}
         \centering
         \includegraphics[width=0.84\textwidth]{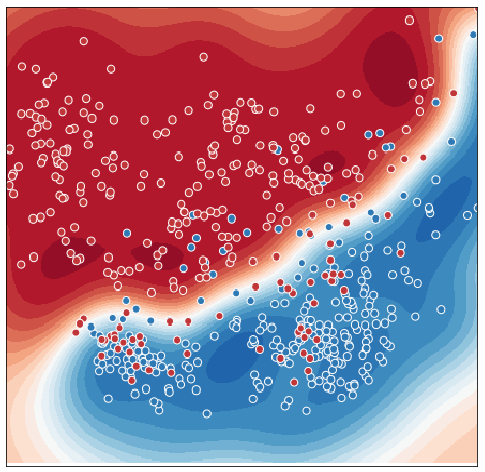}
         \caption{Baseline: (S$^{\ast}$$\rightarrow$S)}
         \label{fig:pointda_pcm_sc_s}
     \end{subfigure}
     \hfill
     \begin{subfigure}[b]{0.24\textwidth}
         \centering
         \includegraphics[width=0.84\textwidth]{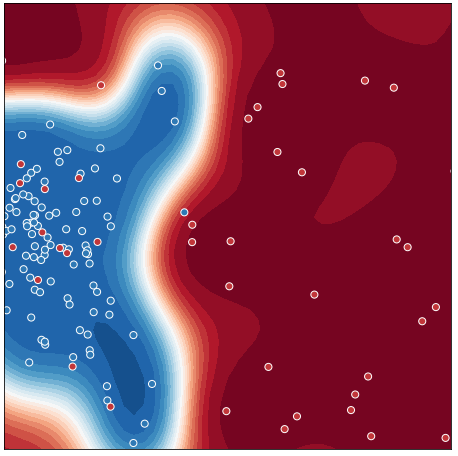}
         \caption{Baseline+CL:(S$^{\ast}$ $\rightarrow$S)}
         \label{fig:pointda_pcm_cl_sc_s}
     \end{subfigure}
     \hfill
     \begin{subfigure}[b]{0.24\textwidth}
         \centering
         \includegraphics[width=0.84\textwidth]{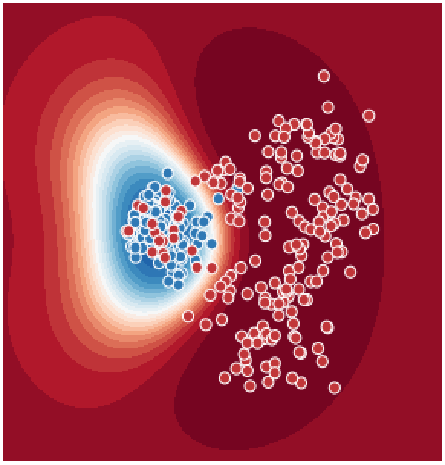}
         \caption{COT: (S$^{\ast}$ $\rightarrow$S)}
         \label{fig:pointda_pcm_ot_sc_s}
     \end{subfigure}
     \hfill
     \begin{subfigure}[b]{0.24\textwidth}
         \centering
         \includegraphics[width=0.84\textwidth]{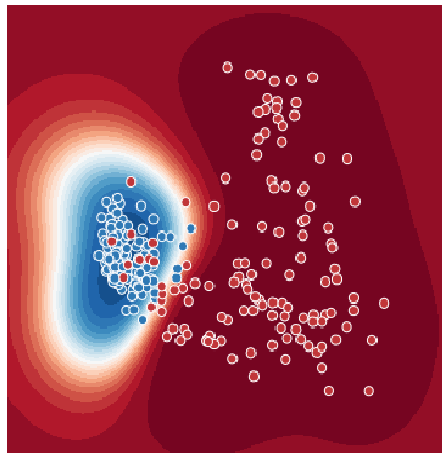}
         \caption{COT+SPST: (S$^{\ast}$$\rightarrow$S)}
         \label{fig:pointda_spst_sc_s}
     \end{subfigure}
\caption{Decision boundaries on target samples for One-vs-Rest (Class: Monitor) for all experiment setups (Row-wise, except S$\rightarrow$M which is in the main paper) of PointDA-10 for Baseline, Baseline with Contrastive Learning (CL), Our COT and Our COT with SPST (Column-wise)}
\label{fig:pointda_boundary}
\end{figure*}

\begin{figure*}[h!]
     \centering
     \begin{subfigure}[b]{0.23\textwidth}
         \centering
         \includegraphics[height= 3.8cm, width=\textwidth]{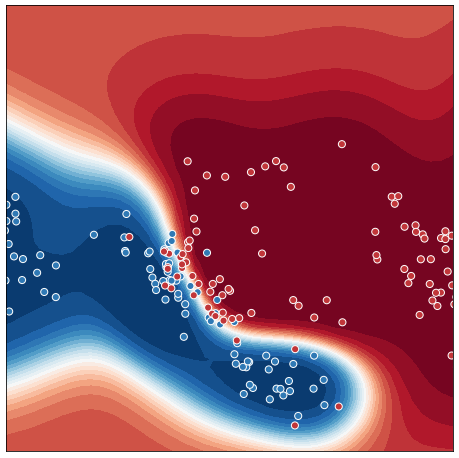}
         \caption{Baseline: (Syn.$\rightarrow$Kin.)}
         \label{fig:graspnet_pcm_syn_kin}
     \end{subfigure}
     \hfill
     \begin{subfigure}[b]{0.23\textwidth}
         \centering
         \includegraphics[height= 3.8cm, width=\textwidth]{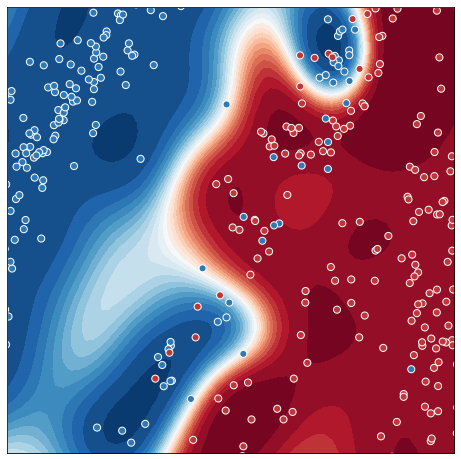}
         \caption{Baseline+CL: (Syn.$\rightarrow$Kin.)}
         \label{fig:graspnet_pcm_cl_syn_kin}
     \end{subfigure}
     \hfill
      \begin{subfigure}[b]{0.23\textwidth}
         \centering
         \includegraphics[height= 3.8cm, width=\textwidth]{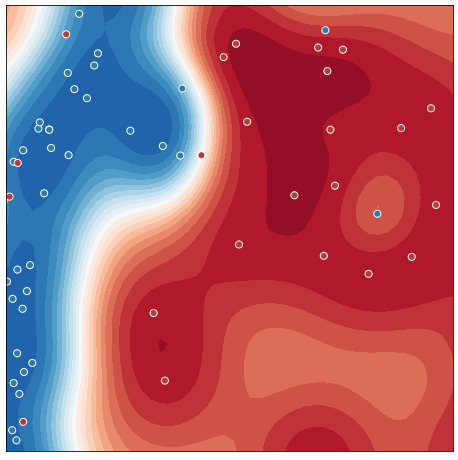}
         \caption{COT: (Syn.$\rightarrow$Kin.)}
         \label{fig:graspnet_cot_syn_kin}
     \end{subfigure}
     \hfill
      \begin{subfigure}[b]{0.23\textwidth}
         \centering
         \includegraphics[height= 3.8cm, width=\textwidth]{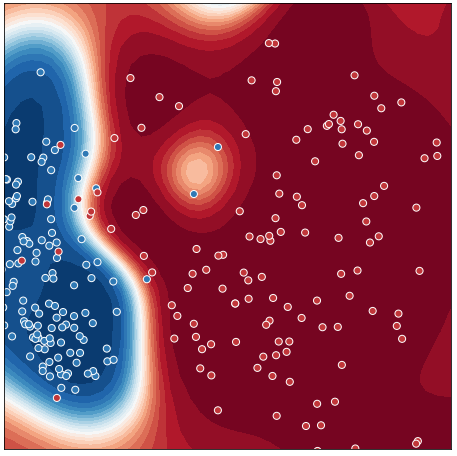}
         \caption{COT+SPST: (Syn.$\rightarrow$Kin.)}
         \label{fig:graspnet_spst_syn_kin}
     \end{subfigure}
     \\
     \begin{subfigure}[b]{0.23\textwidth}
         \centering
         \includegraphics[height= 3.8cm, width=\textwidth]{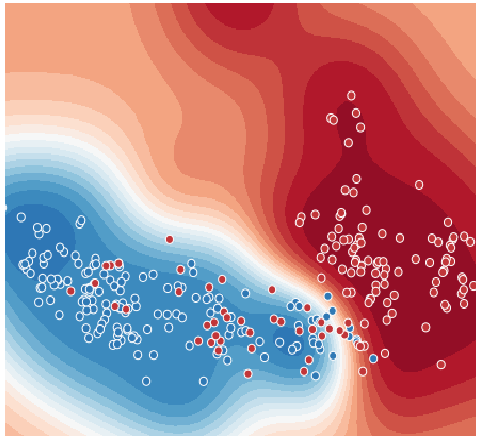}
         \caption{Baseline: (Syn.$\rightarrow$RS.)}
         \label{fig:graspnet_pcm_syn_real}
     \end{subfigure}
     \hfill
     \begin{subfigure}[b]{0.23\textwidth}
         \centering
         \includegraphics[height= 3.8cm, width=\textwidth]{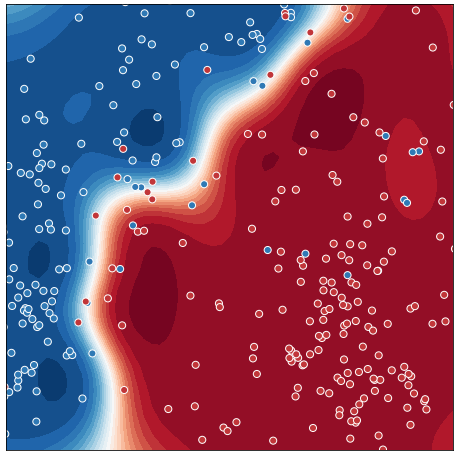}
         \caption{Baseline+CL: (Syn.$\rightarrow$RS.)}
         \label{fig:graspnet_pcm_cl_syn_real}
     \end{subfigure}
     \hfill
     \begin{subfigure}[b]{0.23\textwidth}
         \centering
         \includegraphics[height= 3.8cm, width=\textwidth]{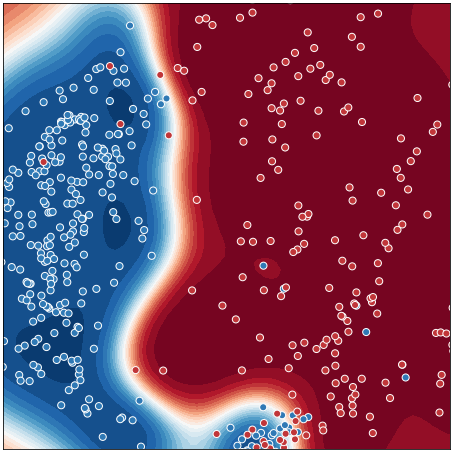}
         \caption{COT: (Syn.$\rightarrow$RS.)}
         \label{fig:graspnet_cot_syn_real}
     \end{subfigure}
     \hfill
     \begin{subfigure}[b]{0.23\textwidth}
         \centering
         \includegraphics[height= 3.8cm, width=\textwidth]{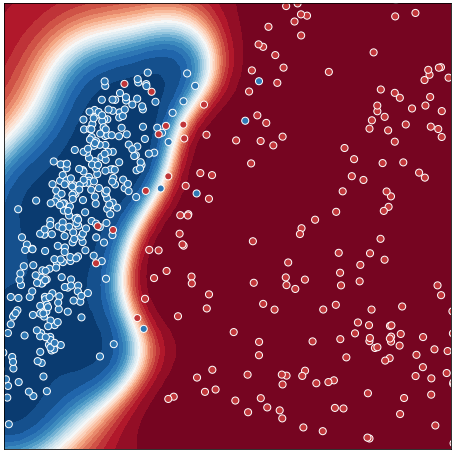}
         \caption{COT+SPST: (Syn.$\rightarrow$RS.)}
         \label{fig:graspnet_spst_syn_real}
     \end{subfigure}
     \\
     \centering
     \begin{subfigure}[b]{0.23\textwidth}
         \centering
         \includegraphics[height= 3.8cm, width=\textwidth]{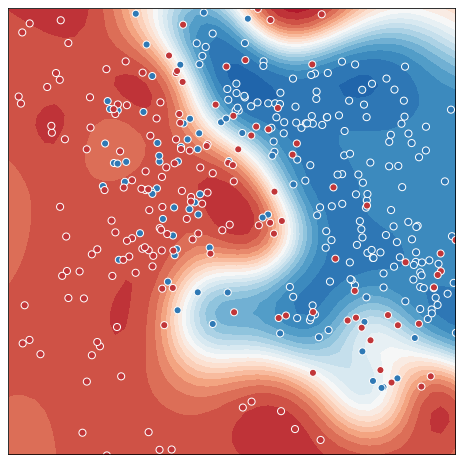}
         \caption{Baseline: (Kin.$\rightarrow$RS.)}
         \label{fig:graspnet_pcm_kin_real}
     \end{subfigure}
     \hfill
     \centering
     \begin{subfigure}[b]{0.23\textwidth}
         \centering
         \includegraphics[height= 3.8cm, width=\textwidth]{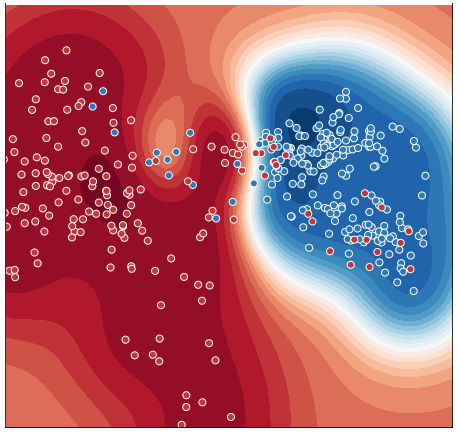}
         \caption{Baseline+CL: (Kin.$\rightarrow$RS.)}
         \label{fig:graspnet_pcm_cl_kin_real}
     \end{subfigure}
     \hfill
     \begin{subfigure}[b]{0.23\textwidth}
         \centering
         \includegraphics[height= 3.8cm, width=\textwidth]{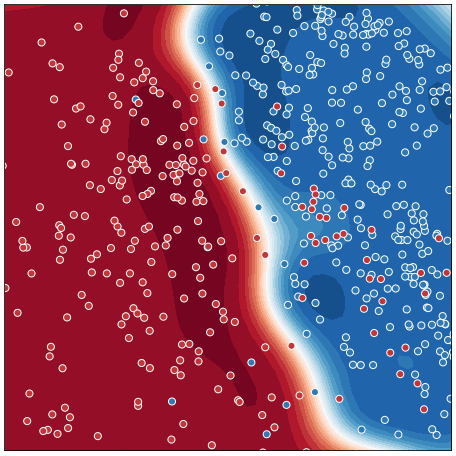}
         \caption{COT: (Kin.$\rightarrow$RS.)}
         \label{fig:graspnet_cot_kin_real}
     \end{subfigure}
     \hfill
     \centering
     \begin{subfigure}[b]{0.23\textwidth}
         \centering
         \includegraphics[height= 3.8cm, width=\textwidth]{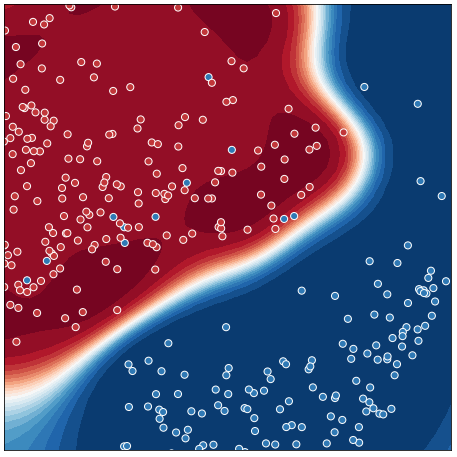}
         \caption{COT+SPST: (Kin.$\rightarrow$RS.)}
         \label{fig:graspnet_spst_kin_real}
     \end{subfigure}
     \\
     \begin{subfigure}[b]{0.23\textwidth}
         \centering
         \includegraphics[height= 3.8cm, width=\textwidth]{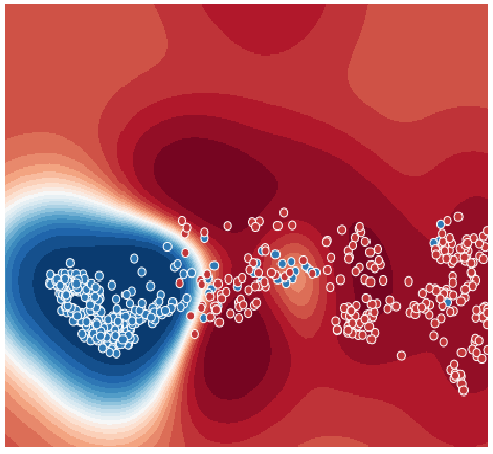}
         \caption{Baseline: (RS.$\rightarrow$Kin)}
         \label{fig:graspnet_pcm_real_kin}
     \end{subfigure}   
     \hfill
     \begin{subfigure}[b]{0.23\textwidth}
         \centering
         \includegraphics[height= 3.8cm, width=\textwidth]{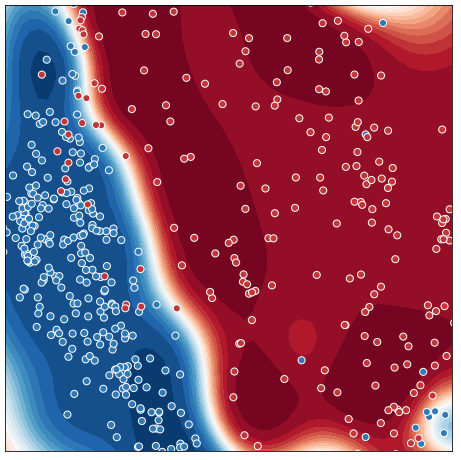}
         \caption{Baseline+CL: (RS.$\rightarrow$Kin)}
         \label{fig:graspnet_pcm_cl_real_kin}
     \end{subfigure}
     \hfill
     \centering
     \begin{subfigure}[b]{0.23\textwidth}
         \centering
         \includegraphics[height= 3.8cm, width=\textwidth]{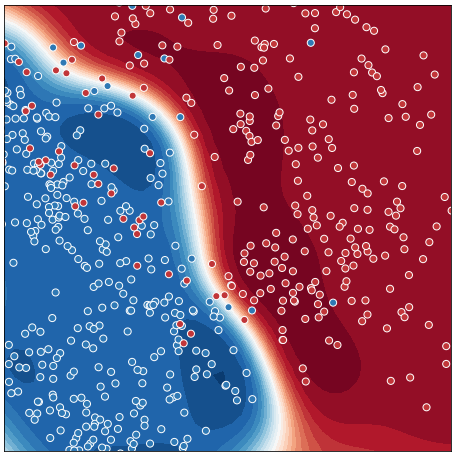}
         \caption{COT: (RS.$\rightarrow$Kin)}
         \label{fig:graspnet_cot_real_kin}
     \end{subfigure}
     \begin{subfigure}[b]{0.23\textwidth}
         \centering
         \includegraphics[height= 3.8cm, width=\textwidth]{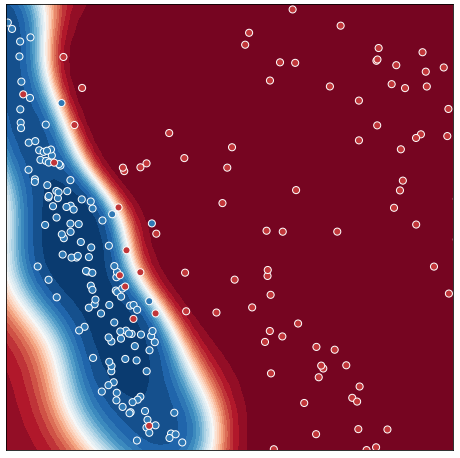}
         \caption{COT+SPST: (RS.$\rightarrow$Kin)}
         \label{fig:graspnet_spst_real_kin}
     \end{subfigure} 
     \hfill
\caption{Decision boundaries on target samples for One-vs-Rest (Class: Dish) for all experiment setups (Row-wise) of GraspNetPC-10 for Baseline, Baseline with Contrastive Learning (CL), Our COT and Our COT with SPST (Column-wise)}
\label{fig:graspnet_boundary}
\end{figure*}

In this section, we analyze the effectiveness of our proposed architectures (COT and COT with SPST) by examining decision boundaries of the baseline variants (PCM and PCM with Contrastive Learning) and our proposed architectures learned on all the experimental settings of PointDA-10 and GraspNetPC-10 datasets. For this analysis, we create decision boundaries by extracting learned representations of the target dataset from 3D encoder and the corresponding predicted target labels. We finally fit an SVM by considering ``One-vs-Rest" strategy for a class of interest (here Class: Monitor for PointDA-10 and Class: Dish for GraspNetPC-10). Figure \ref{fig:pointda_boundary} and \ref{fig:graspnet_boundary} shows the corresponding decision boundaries of all possible settings of PointDA-10 (except ShapeNet-ModelNet i.e., S$\rightarrow$M which is in the main paper) and GraspNetPC-10 dataset, respectively.
In both of the datasets, it is clearly visible from the decision boundaries that our proposed COT produces stronger and more robust decision boundaries. For all PointDA-10 combinations, the decision boundaries of our method has compact 
decision boundary. For the GraspNet-10 combinations, our methods classifier has high confidence (either dark red or dark blue regions) compared to the baselines.
These decision boundary plots show the effectiveness of our domain alignment method endowed by contrastive learning and optimal transport.

\begin{figure*}[t] 
     \centering
     \begin{subfigure}[b]{0.24\textwidth}
         \centering
         \includegraphics[width=0.85\textwidth]{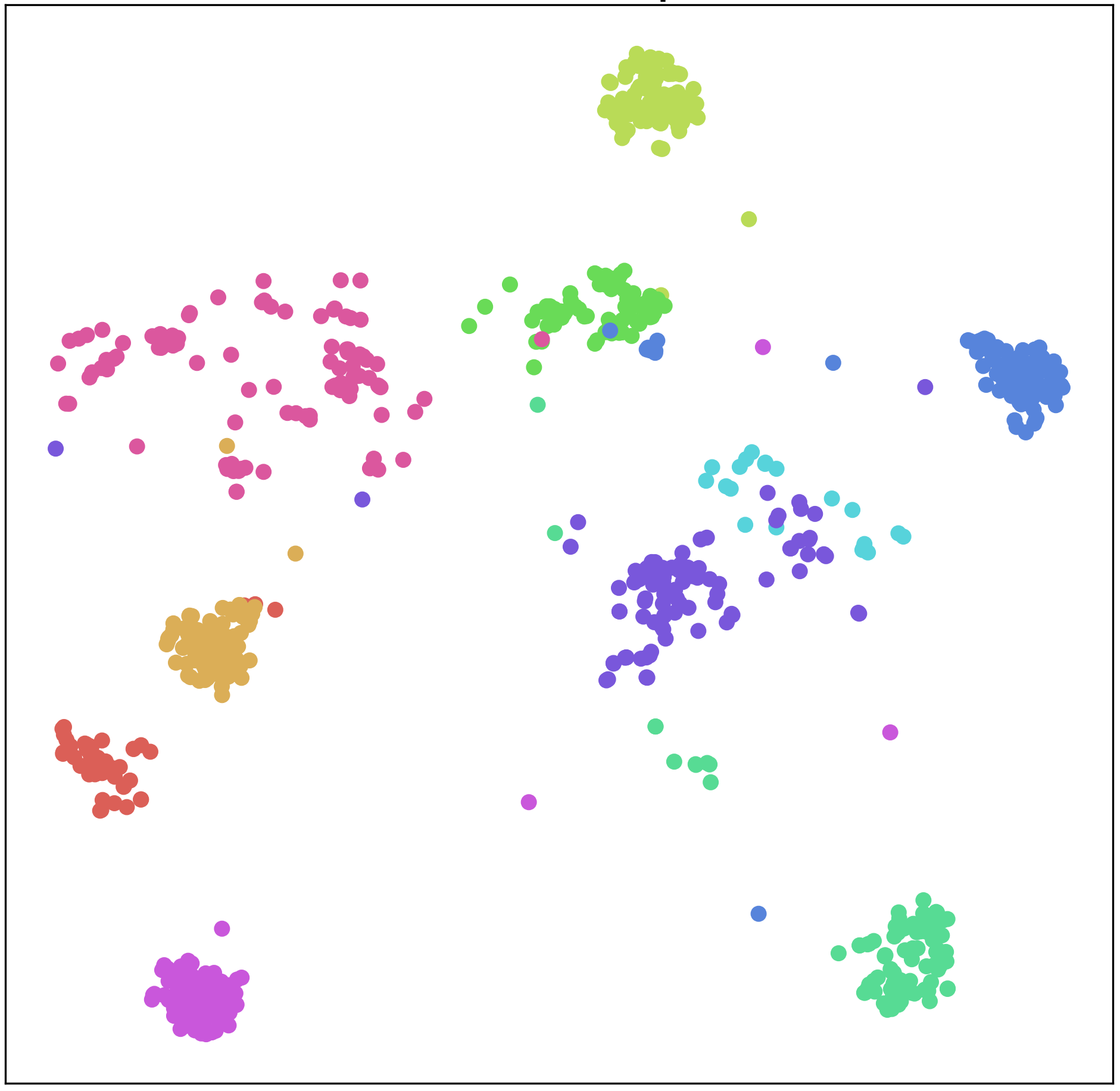}
         \caption{ Source Baseline: (M$\rightarrow$S)}
         \label{Point_src_pcm_M_S}
     \end{subfigure}
     \hfill
     \centering
     \begin{subfigure}[b]{0.24\textwidth}
         \centering
         \includegraphics[width=0.85\textwidth]{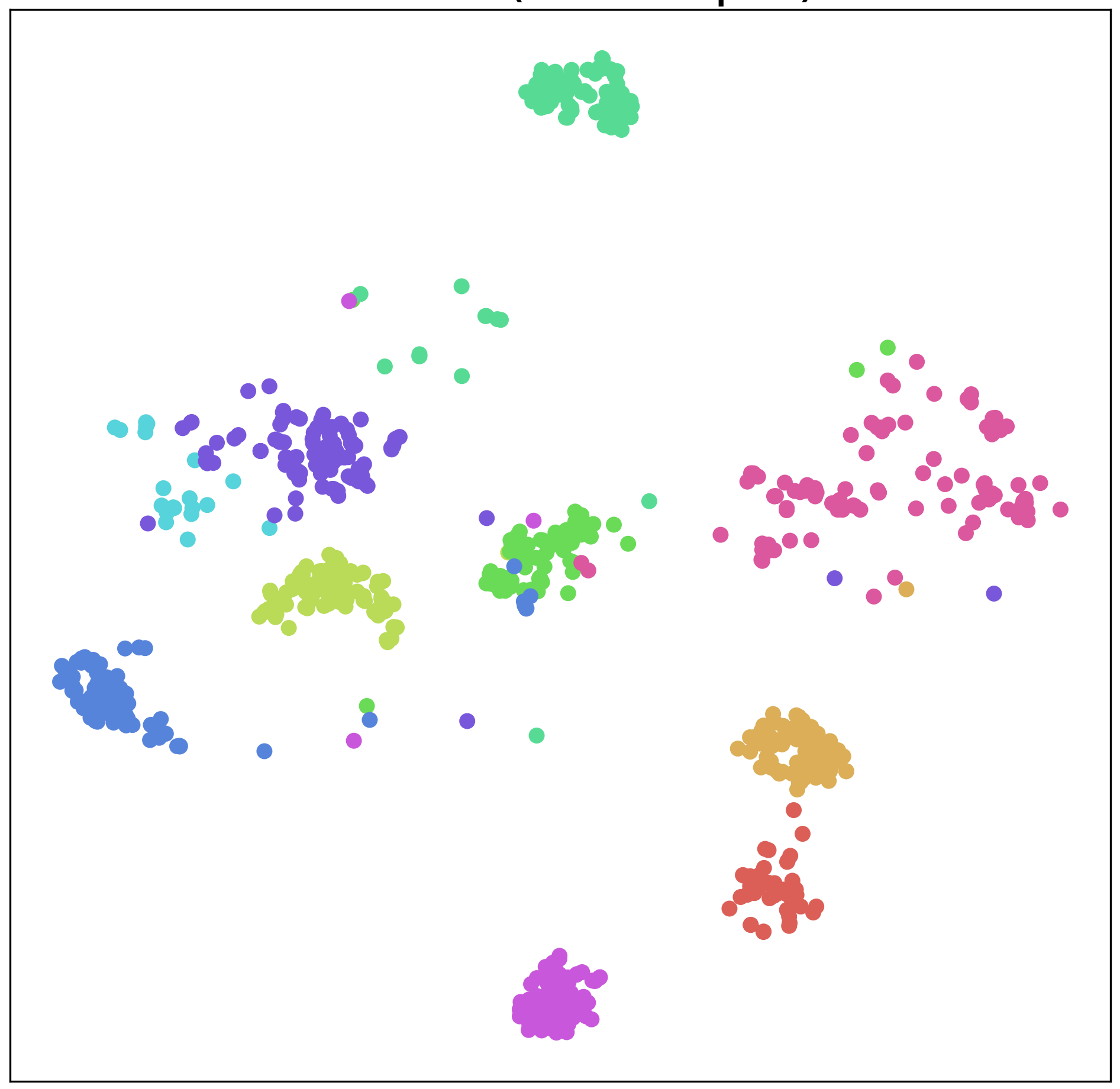}
         \caption{Source COT+SPST: (M$\rightarrow$S) }
         \label{Point_src_COT_SPST_M_S}
     \end{subfigure}
    \hfill
    \centering
     \begin{subfigure}[b]{0.24\textwidth}
         \centering
         \includegraphics[width=0.85\textwidth]{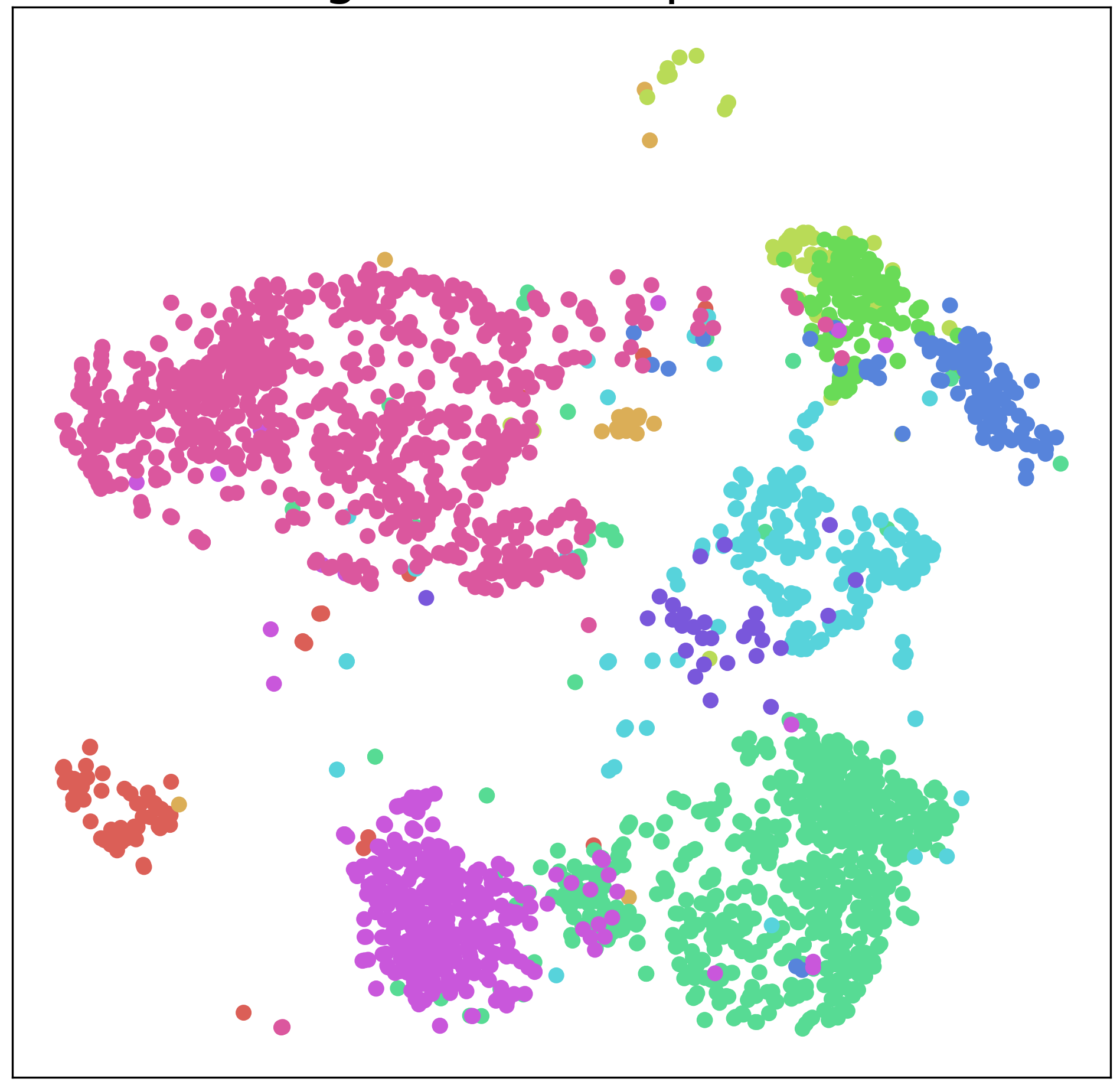}
         \caption{Target Baseline: (M$\rightarrow$S) }
         \label{Point_trgt_pcm_M_S}
     \end{subfigure}
     \hfill
     \centering
     \begin{subfigure}[b]{0.24\textwidth}
         \centering
         \includegraphics[width=0.85\textwidth]{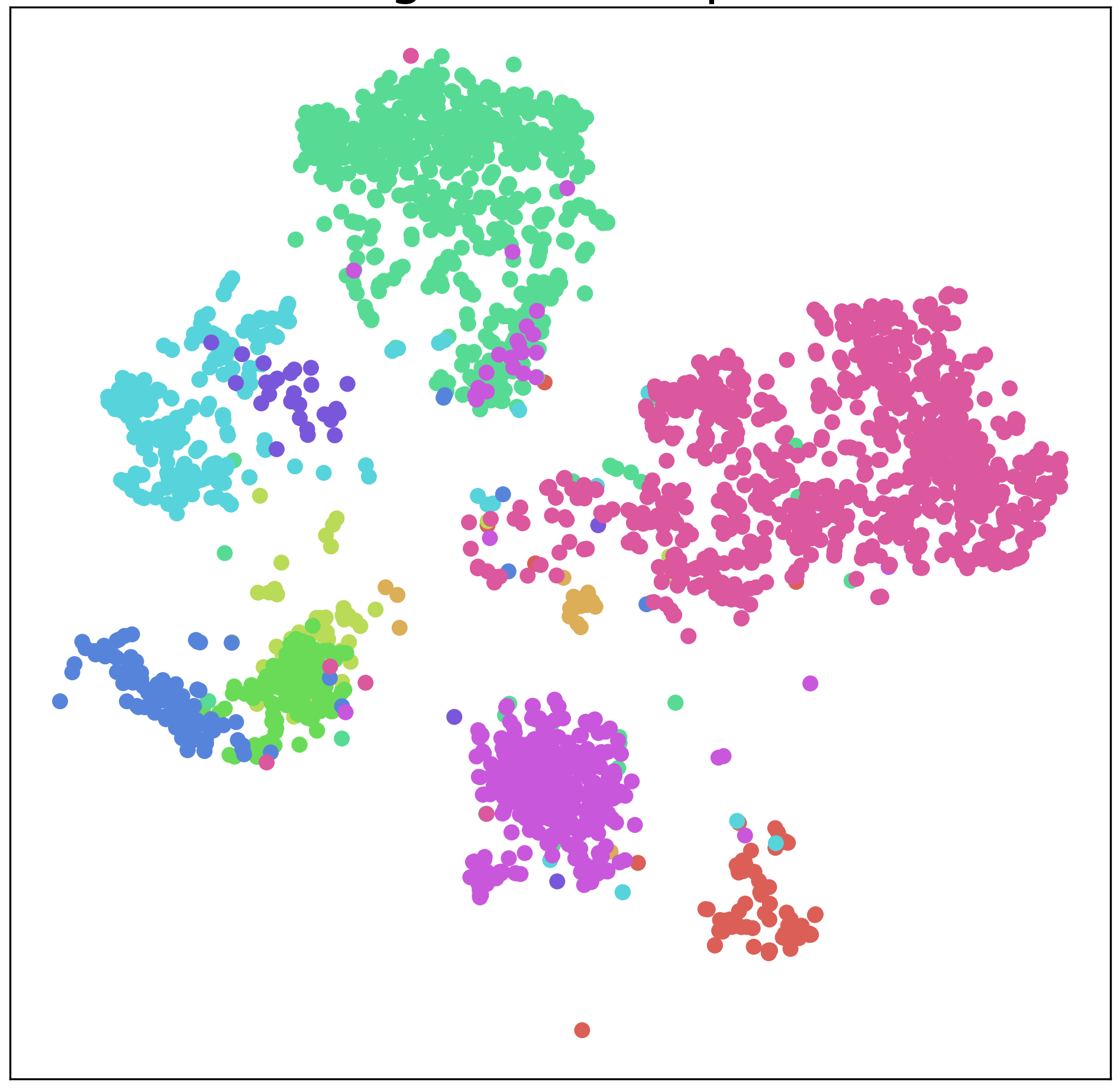}
         \caption{Target COT+SPST: (M$\rightarrow$S) }
         \label{Point_trgt_COT_SPST_M_S}
     \end{subfigure}
     \\
     \centering
     \begin{subfigure}[b]{0.24\textwidth}
         \centering
         \includegraphics[width=0.85\textwidth]{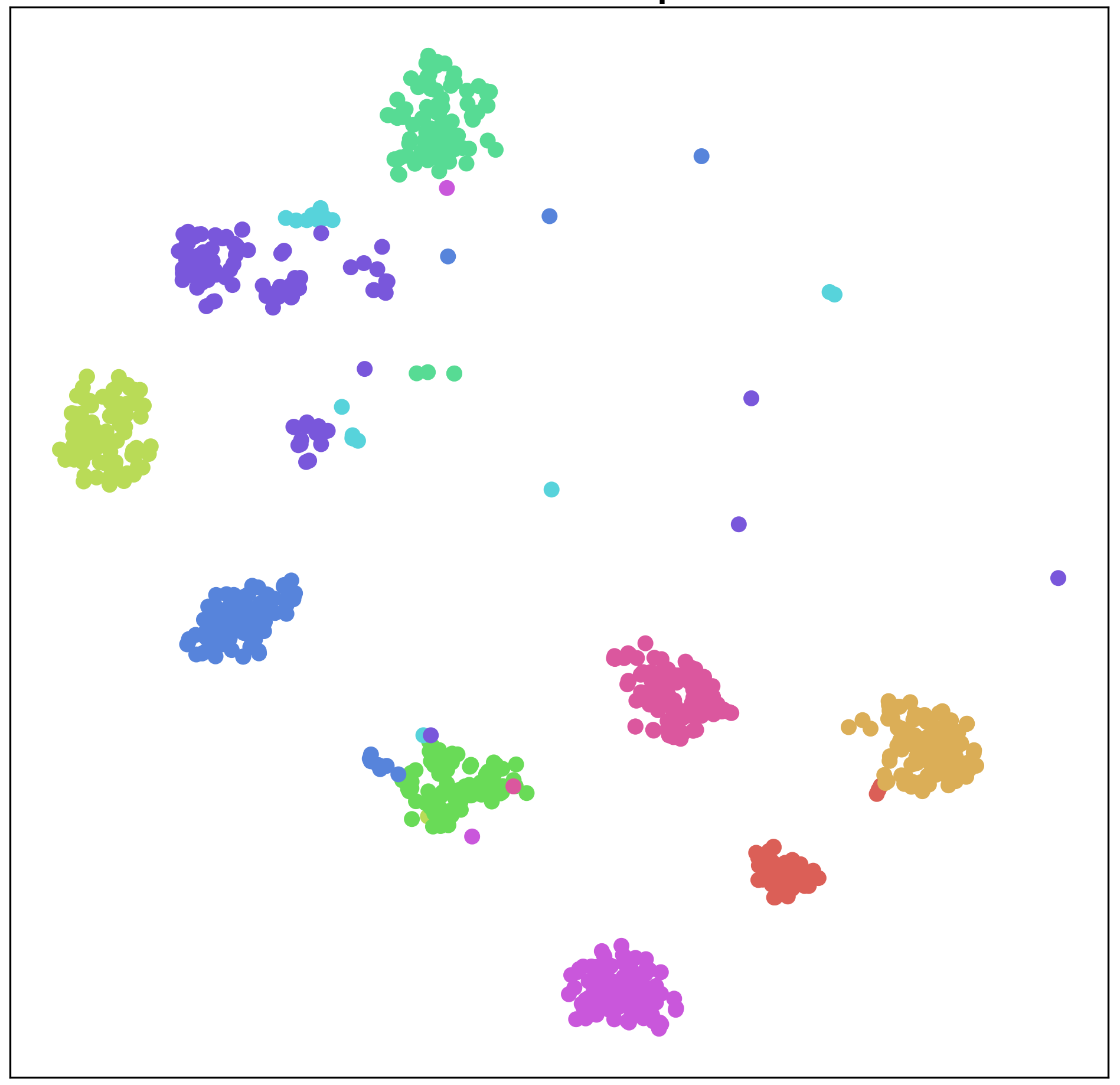}
         \caption{Source Baseline: (M$\rightarrow$S*) }
         \label{Point_src_pcm_M_S*}
     \end{subfigure}
     \hfill
     \centering
     \begin{subfigure}[b]{0.24\textwidth}
         \centering
         \includegraphics[width=0.85\textwidth]{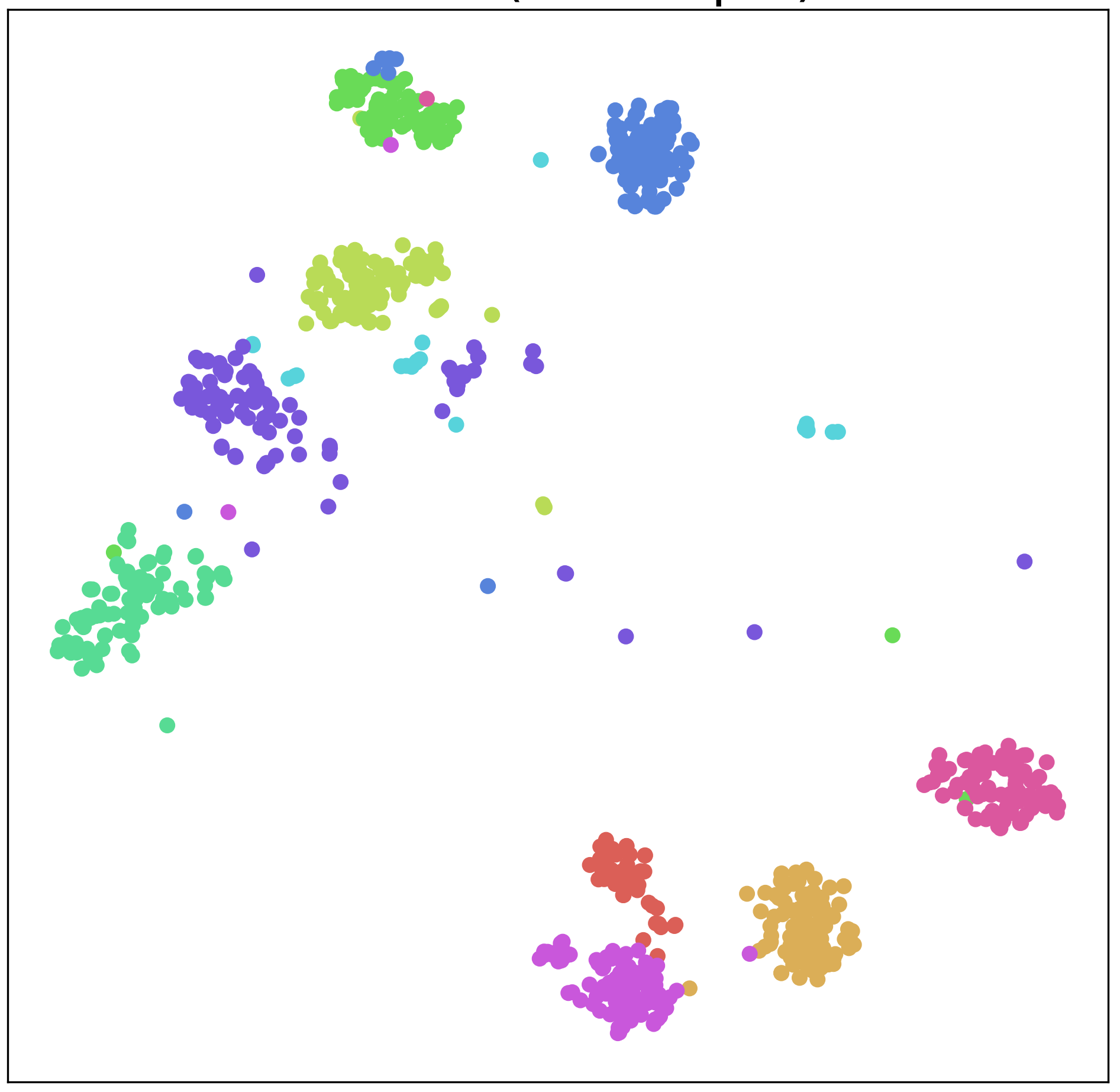}
         \caption{Source COT+SPST: (M$\rightarrow$S*) }
         \label{Point_src_COT_SPST_M_S*}
     \end{subfigure}
     \hfill
     \centering
     \begin{subfigure}[b]{0.24\textwidth}
         \centering
         \includegraphics[width=0.85\textwidth]{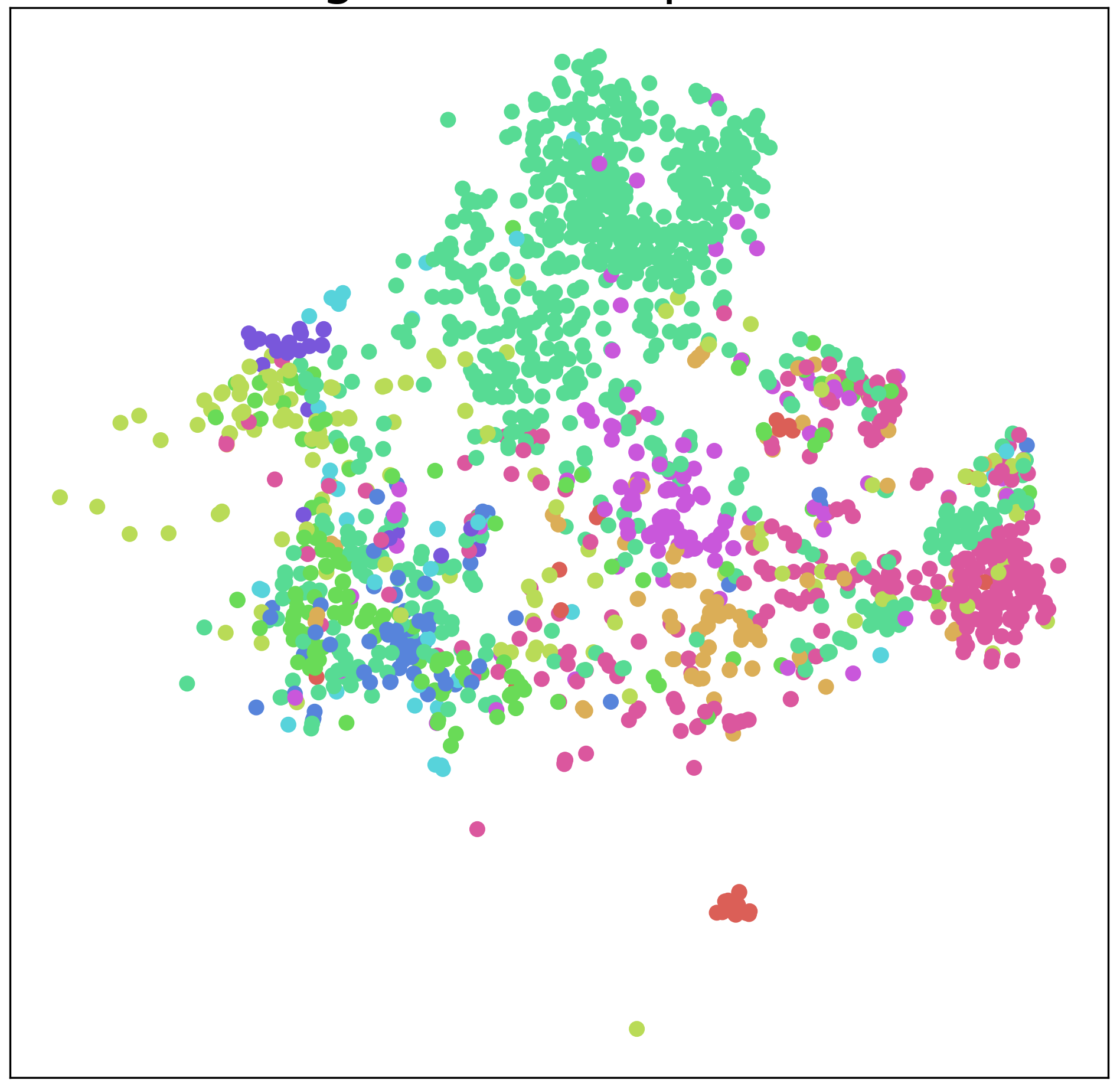}
         \caption{Target Baseline: (M$\rightarrow$S*) }
         \label{Point_trgt_pcm_M_S*}
     \end{subfigure}
     \hfill
     \centering
     \begin{subfigure}[b]{0.24\textwidth}
         \centering
         \includegraphics[width=0.85\textwidth]{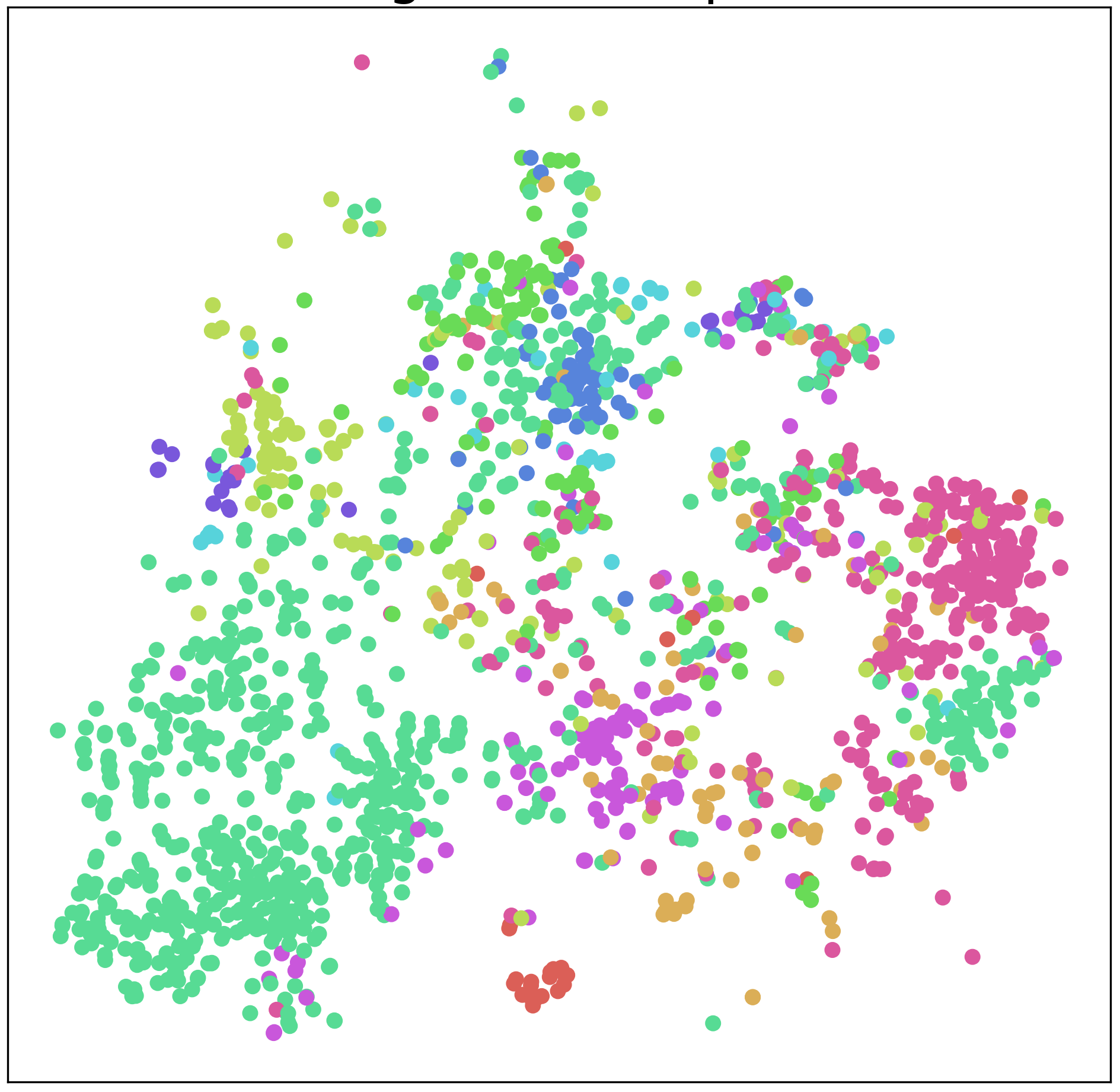}
         \caption{Target COT+SPST: (M$\rightarrow$S*) }
         \label{Point_trgt_COT_SPST_M_S*}
     \end{subfigure}
     \\
     \centering
     \begin{subfigure}[b]{0.24\textwidth}
         \centering
         \includegraphics[width=0.85\textwidth]{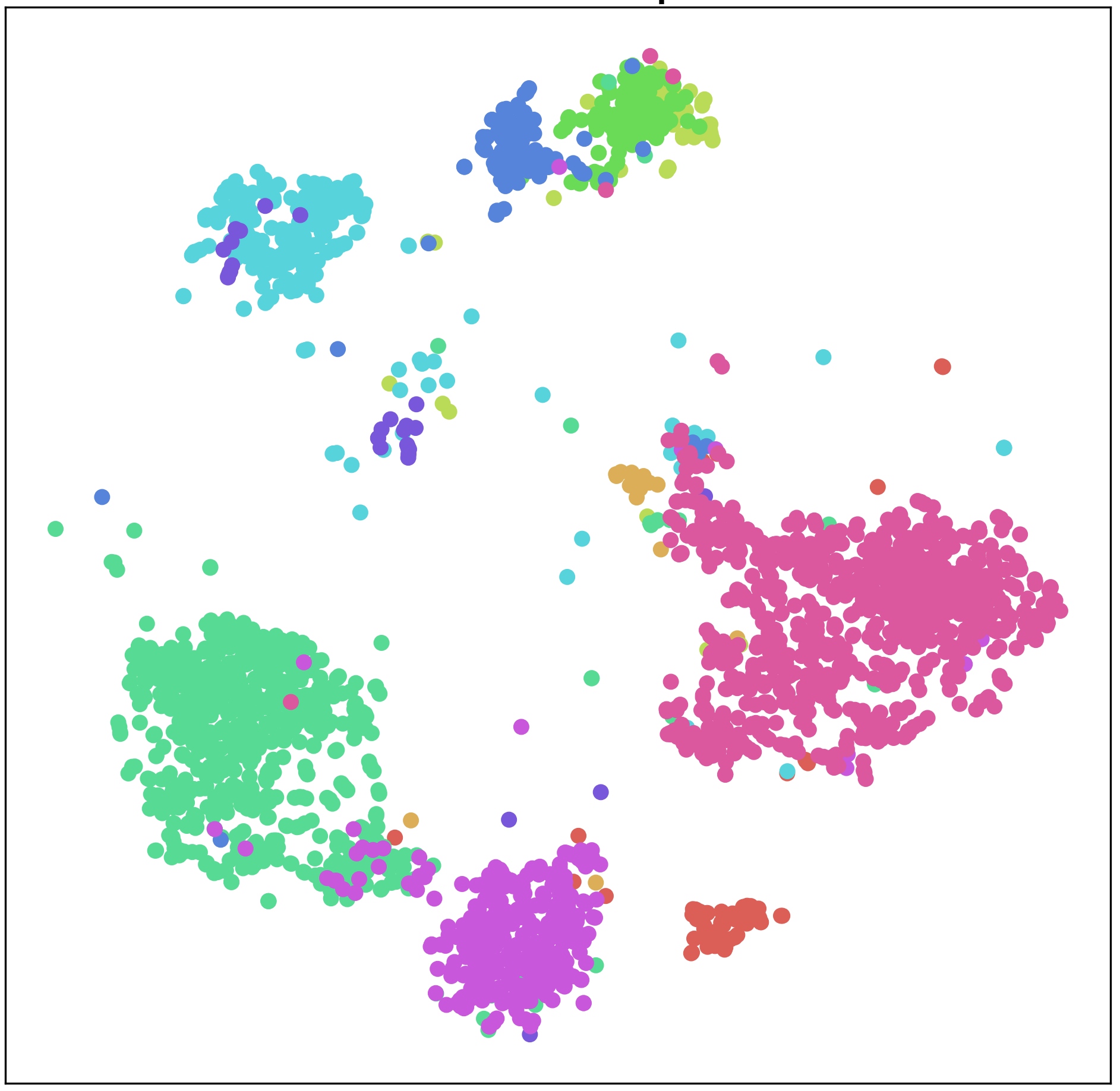}
         \caption{Source Baseline: (S$\rightarrow$S*) }
         \label{Point_src_pcm_S_S*}
     \end{subfigure}
     \hfill
     \centering
     \begin{subfigure}[b]{0.24\textwidth}
         \centering
         \includegraphics[width=0.85\textwidth]{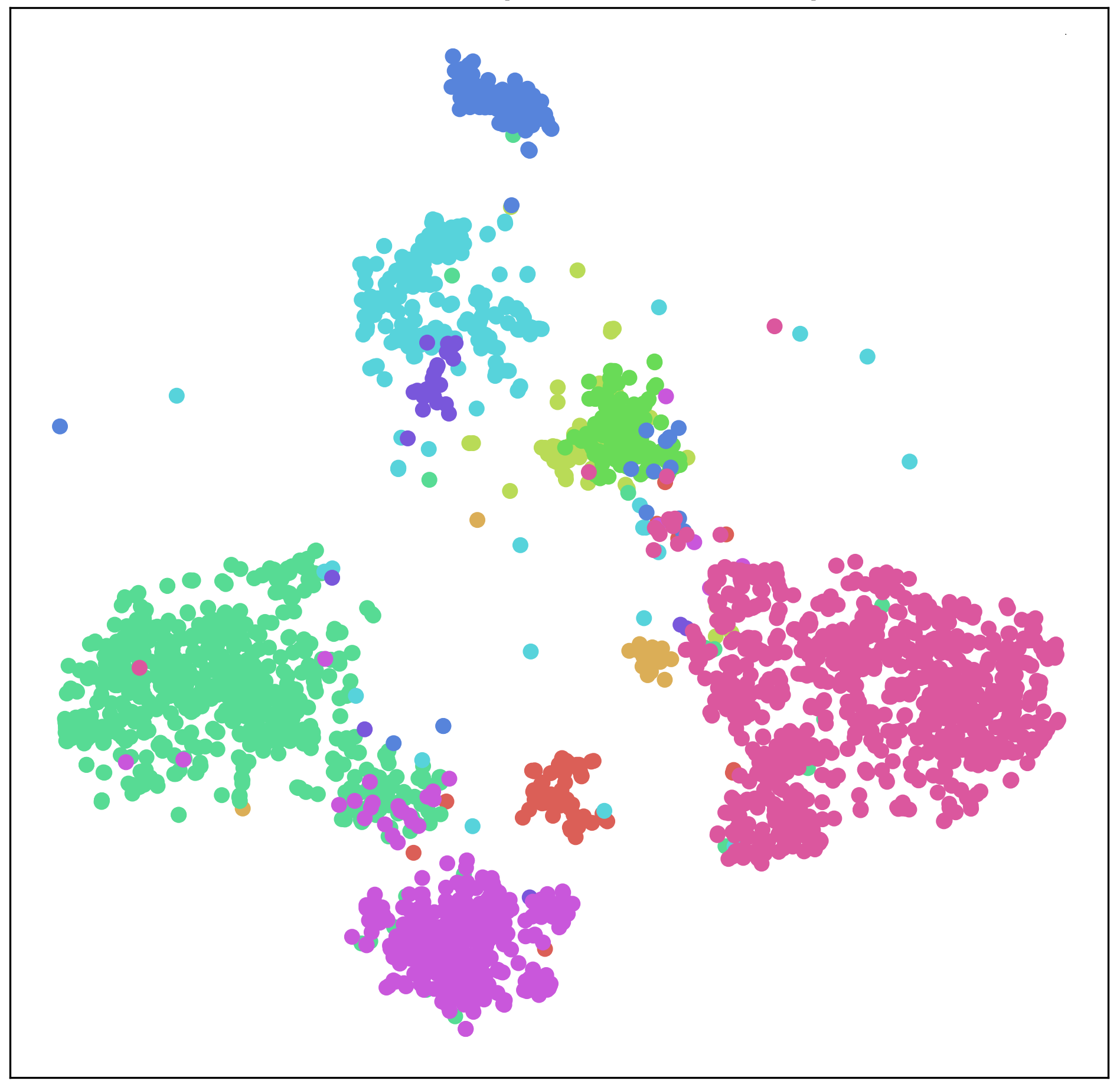}
         \caption{Source COT+SPST: (S$\rightarrow$S*) }
         \label{Point_src_COT_SPST_S_S*}
     \end{subfigure}
     \hfill
     \centering
     \begin{subfigure}[b]{0.24\textwidth}
         \centering
         \includegraphics[width=0.85\textwidth]{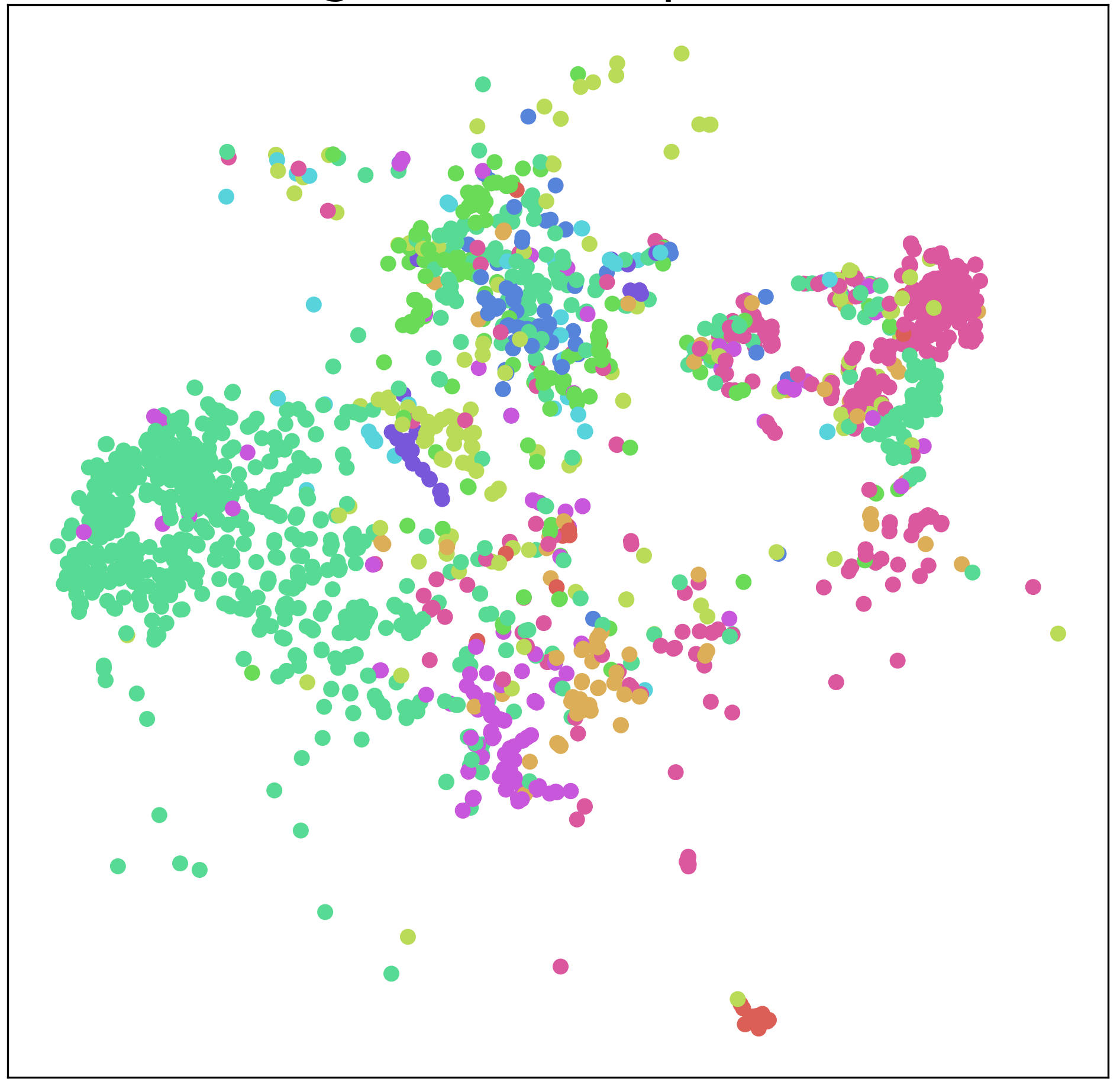}
         \caption{Target Baseline: (S$\rightarrow$S*) }
         \label{Point_trgt_pcm_S_S*}
     \end{subfigure}
     \hfill
     \centering
     \begin{subfigure}[b]{0.24\textwidth}
         \centering
         \includegraphics[width=0.85\textwidth]{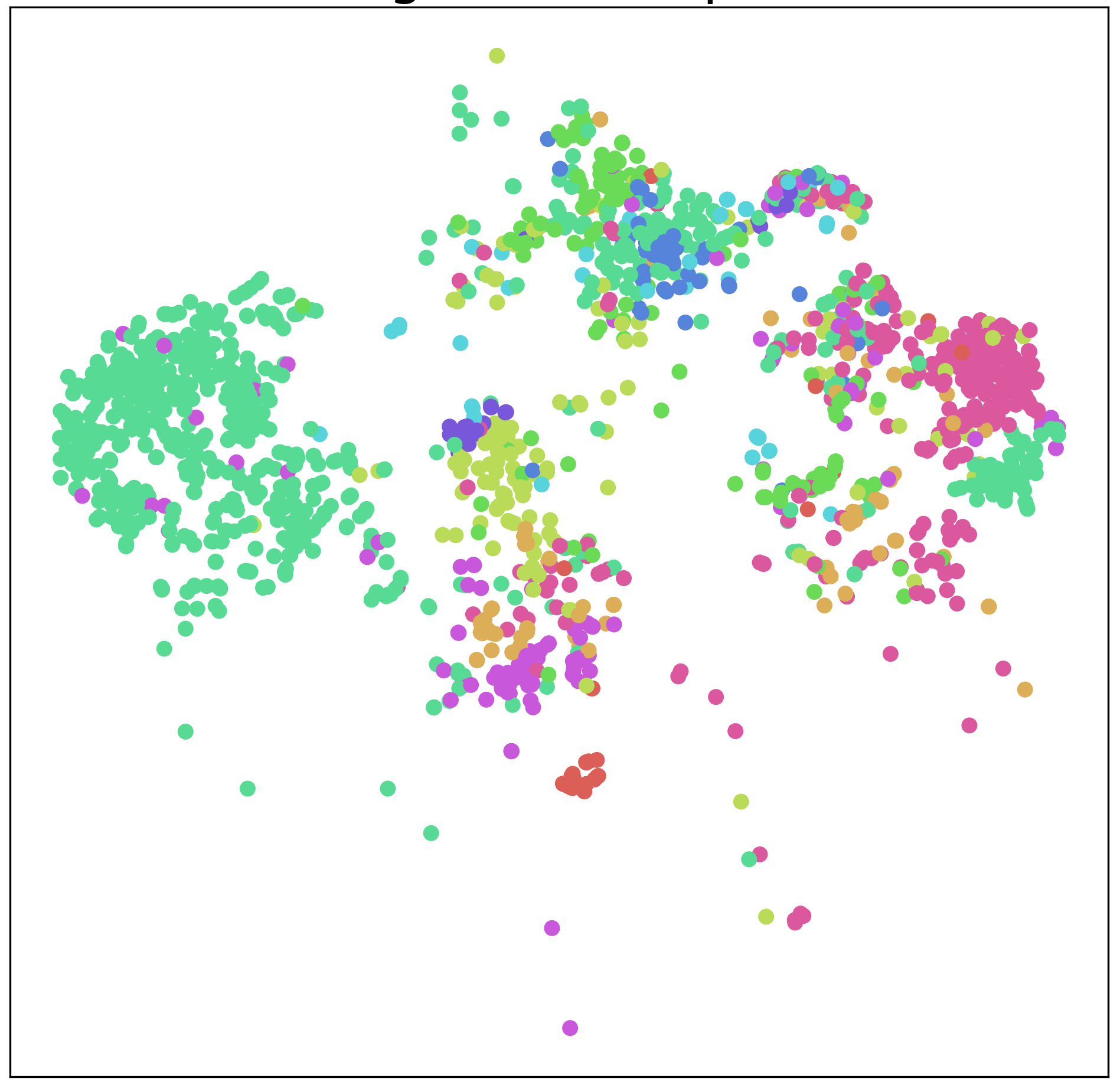}
         \caption{Target COT+SPST: (S$\rightarrow$S*) }
         \label{Point_trgt_COT_SPST_S_S*}
     \end{subfigure}
     \\
     \centering
     \begin{subfigure}[b]{0.24\textwidth}
         \centering
         \includegraphics[width=0.85\textwidth]{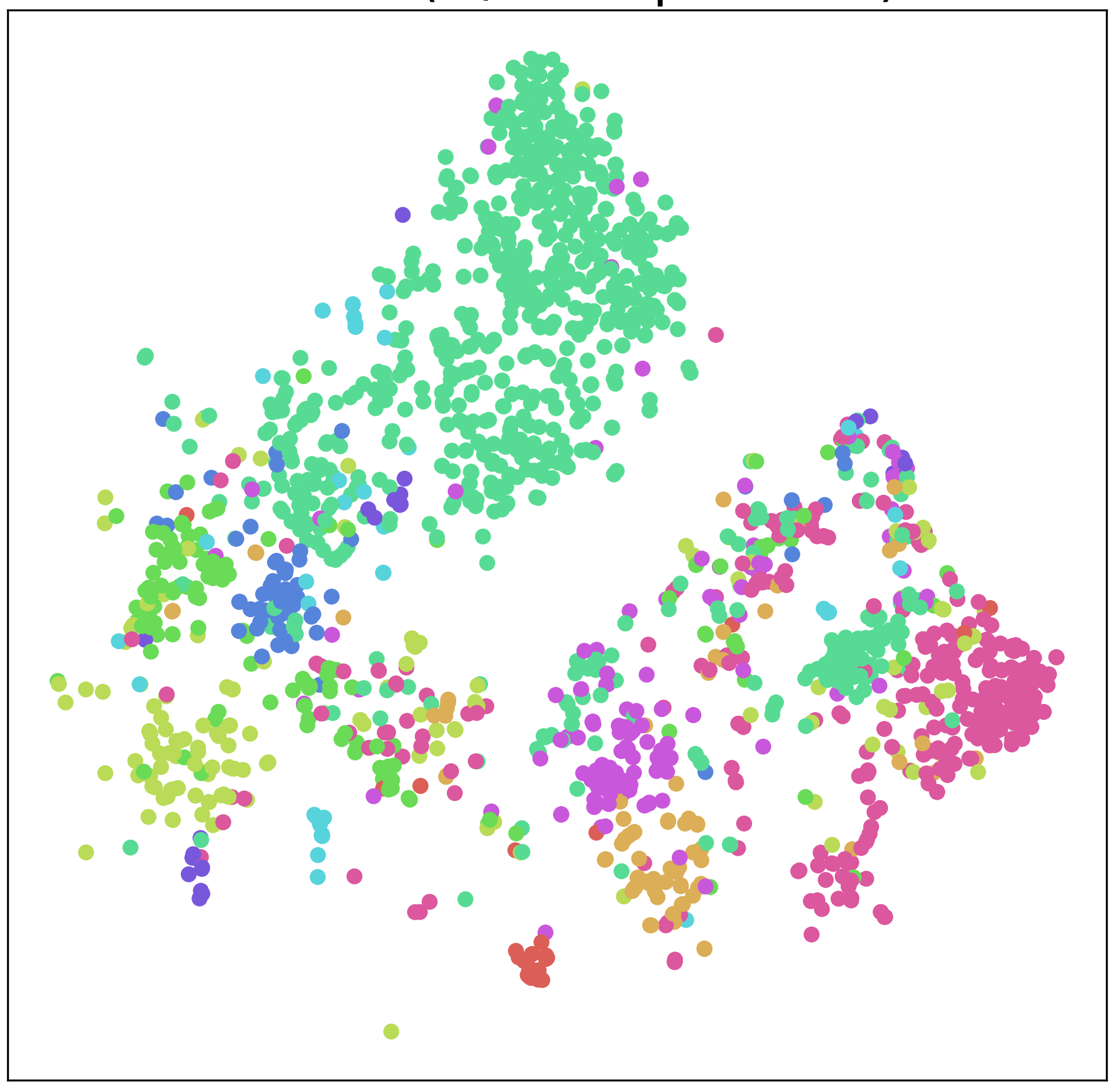}
         \caption{Source Baseline: (S*$\rightarrow$M) }
         \label{Point_src_pcm_S*_M}
     \end{subfigure}
     \hfill
     \centering
     \begin{subfigure}[b]{0.24\textwidth}
         \centering
         \includegraphics[width=0.85\textwidth]{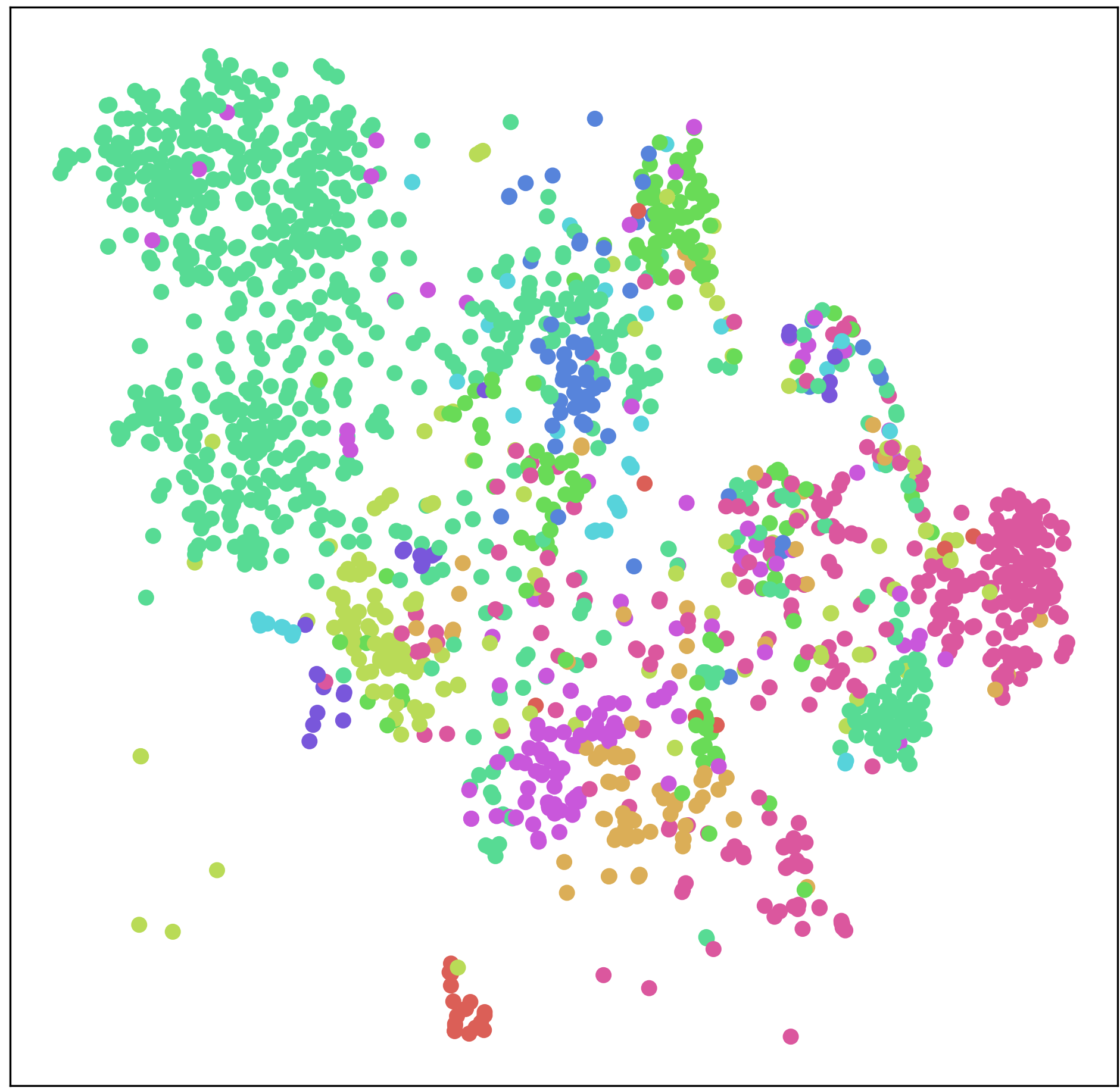}
         \caption{Source COT+SPST: (S*$\rightarrow$M) }
         \label{Point_src_COT_SPST_S*_M}
     \end{subfigure}
    \hfill
    \centering
     \begin{subfigure}[b]{0.24\textwidth}
         \centering
         \includegraphics[width=0.85\textwidth]{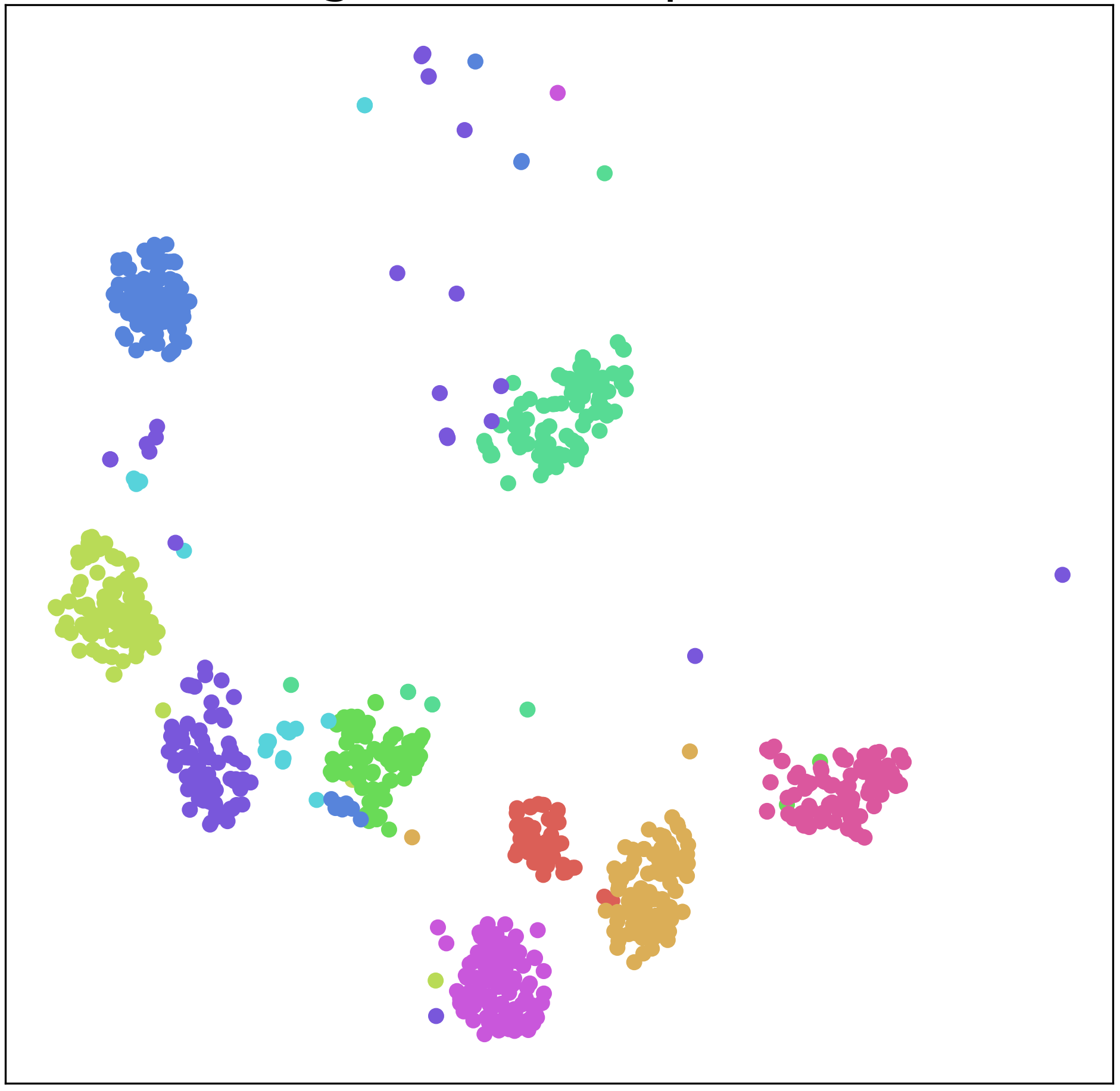}
         \caption{Target Baseline: (S*$\rightarrow$M) }
         \label{Point_trgt_pcm_S*_M}
     \end{subfigure}
     \hfill
     \centering
     \begin{subfigure}[b]{0.24\textwidth}
         \centering
         \includegraphics[width=0.85\textwidth]{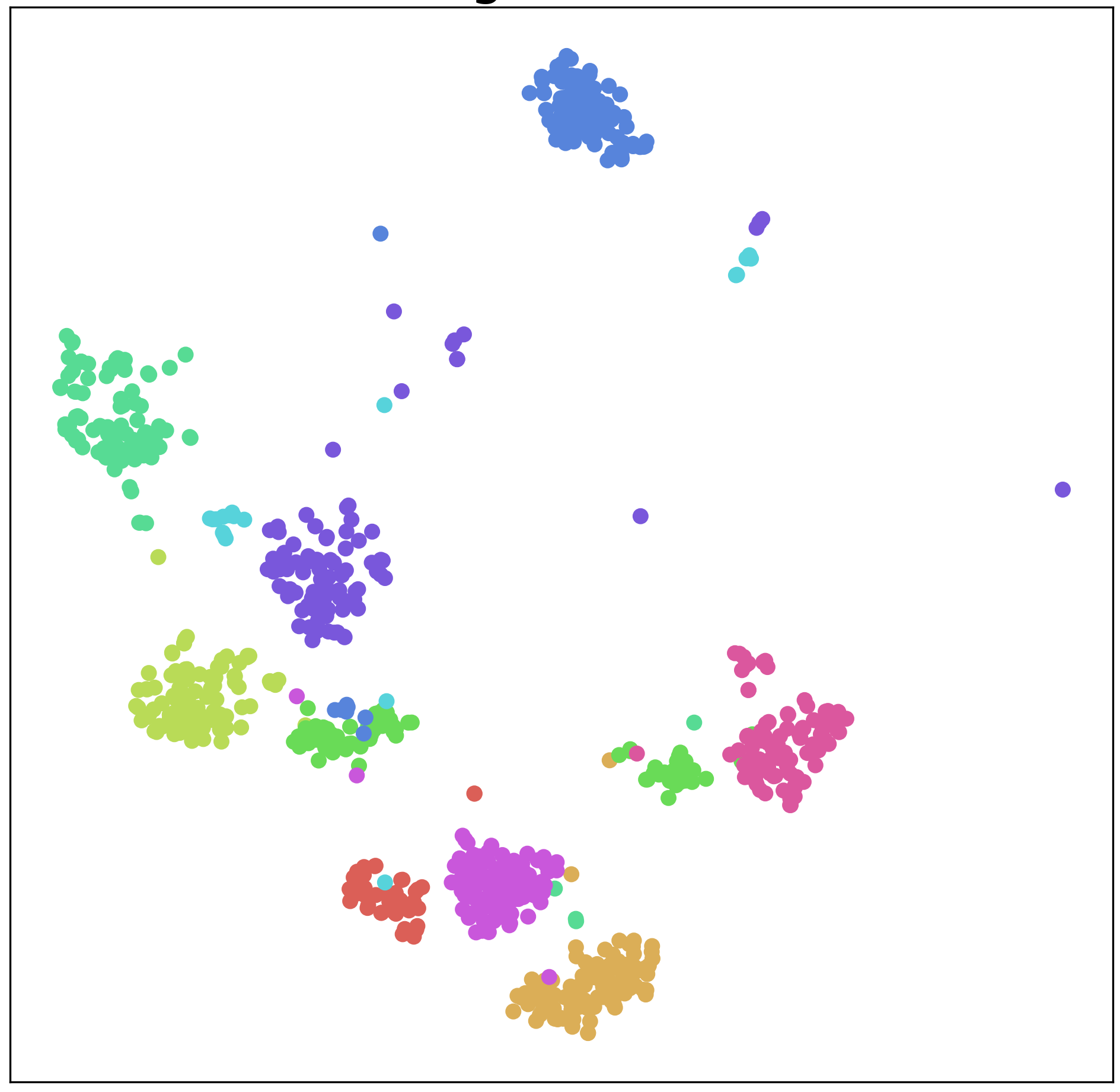}
         \caption{Target COT+SPST: (S*$\rightarrow$M) }
         \label{Point_trgt_COT_SPST_S*_M}
     \end{subfigure}
          \\
     \centering
     \begin{subfigure}[b]{0.24\textwidth}
         \centering
         \includegraphics[width=0.85\textwidth]{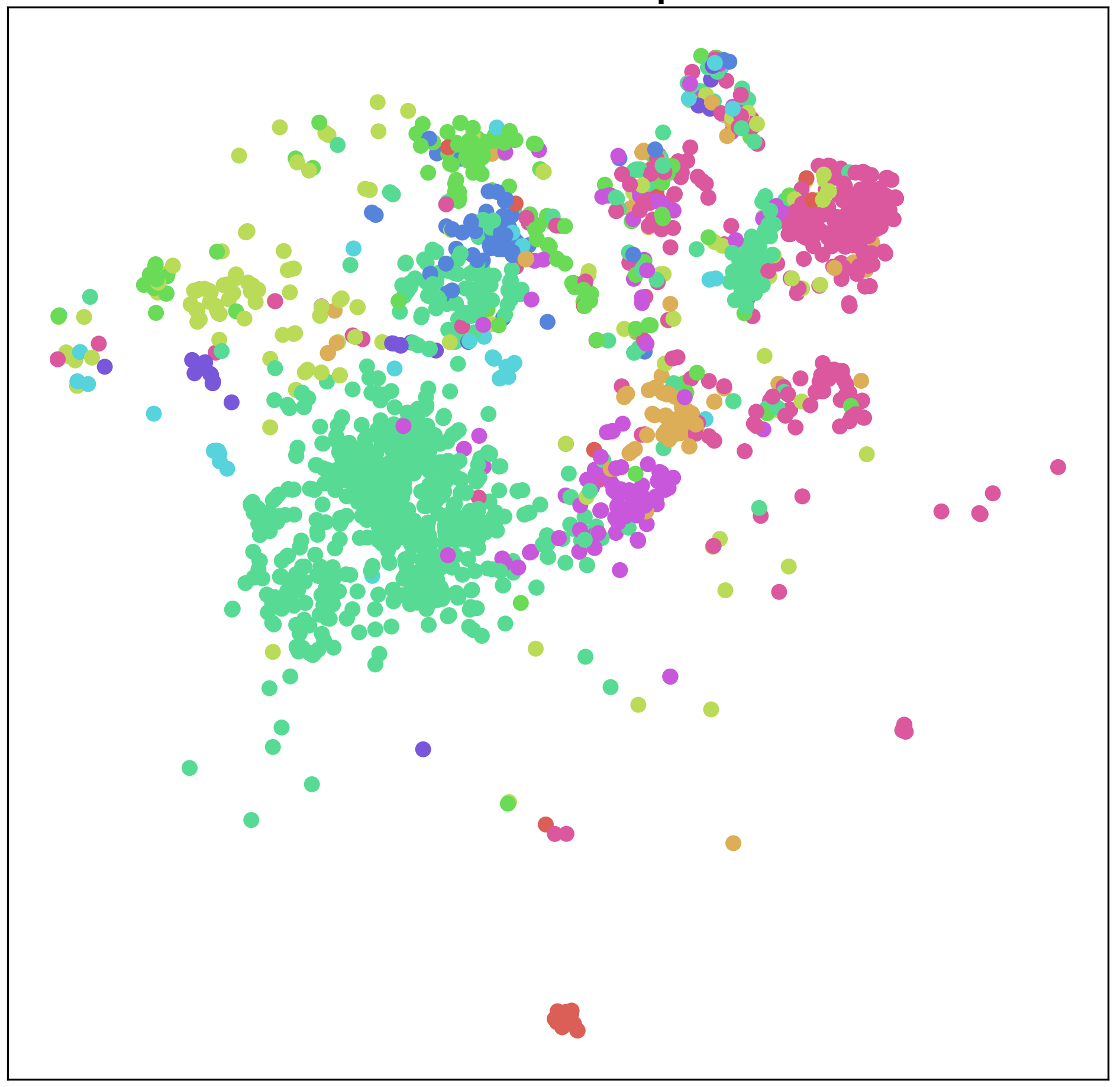}
         \caption{Source Baseline: (S*$\rightarrow$S) }
         \label{Point_src_pcm_S*_S}
     \end{subfigure}
     \hfill
     \centering
     \begin{subfigure}[b]{0.24\textwidth}
         \centering
         \includegraphics[width=0.85\textwidth]{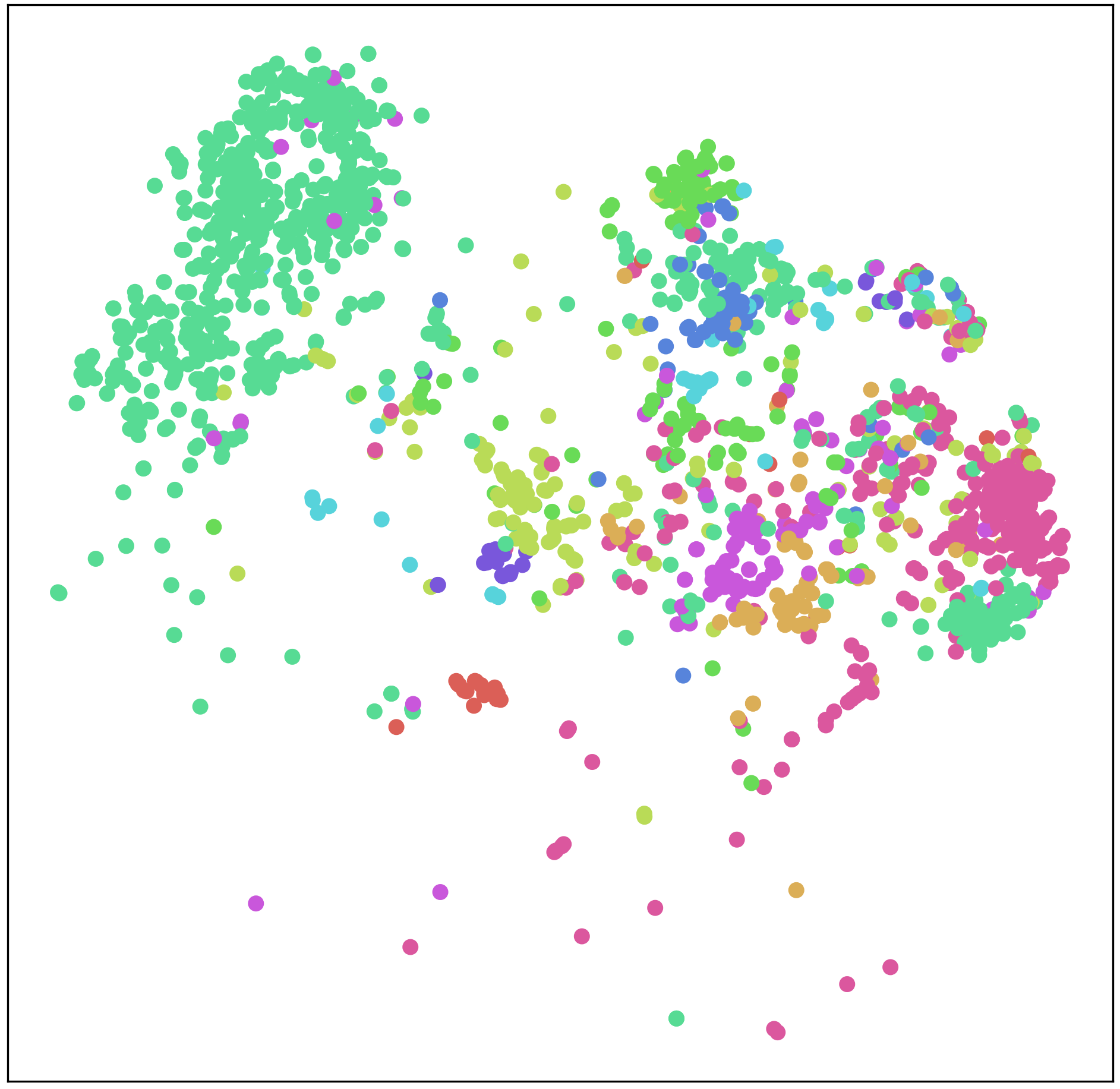}
         \caption{Source COT+SPST: (S*$\rightarrow$S) }
         \label{Point_src_COT_SPST_S*_S}
     \end{subfigure}
    \hfill
    \centering
     \begin{subfigure}[b]{0.24\textwidth}
         \centering
         \includegraphics[width=0.85\textwidth]{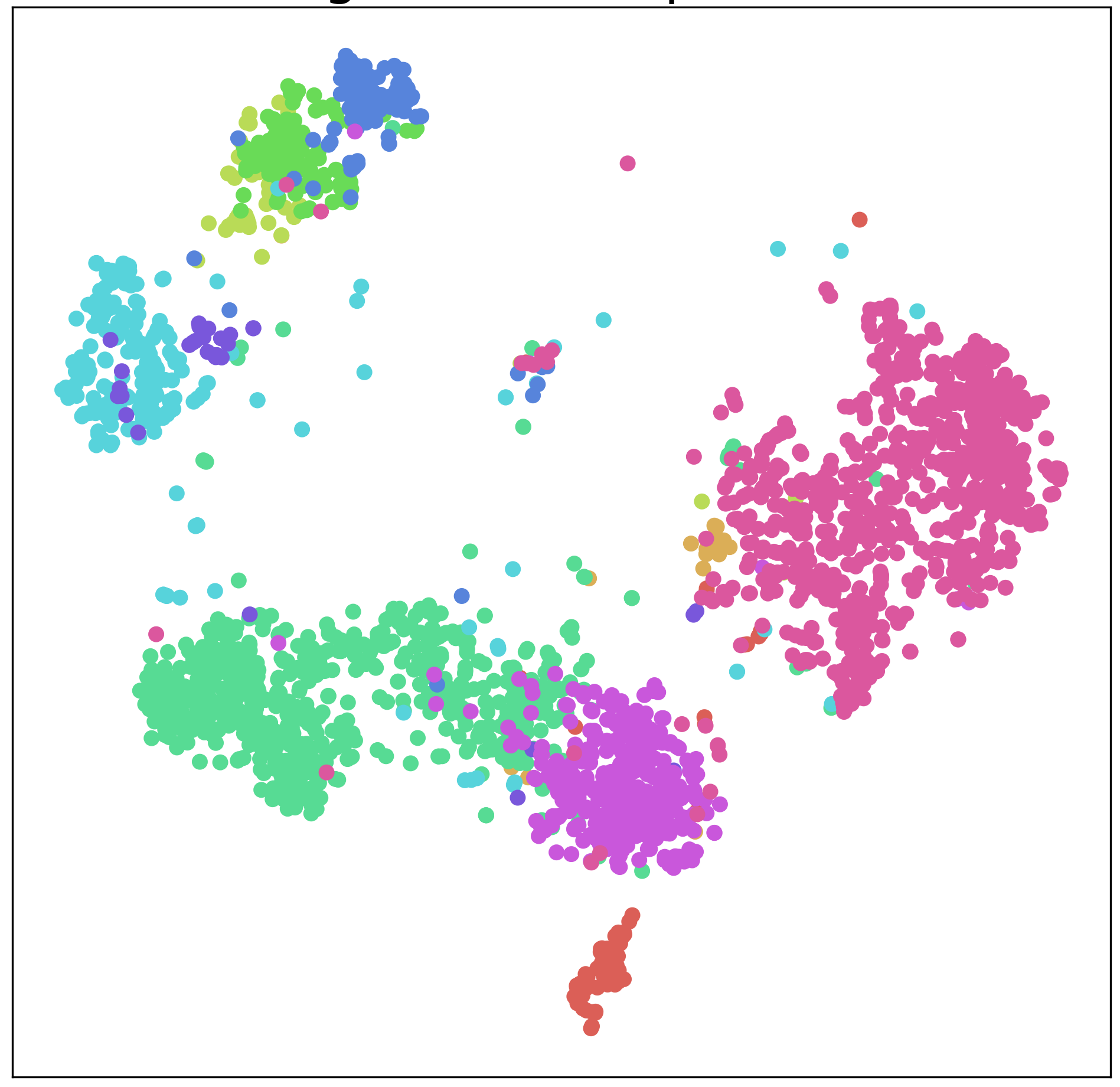}
         \caption{Target Baseline: (S*$\rightarrow$S) }
         \label{Point_trgt_pcm_S*_S}
     \end{subfigure}
     \hfill
     \centering
     \begin{subfigure}[b]{0.24\textwidth}
         \centering
         \includegraphics[width=0.85\textwidth]{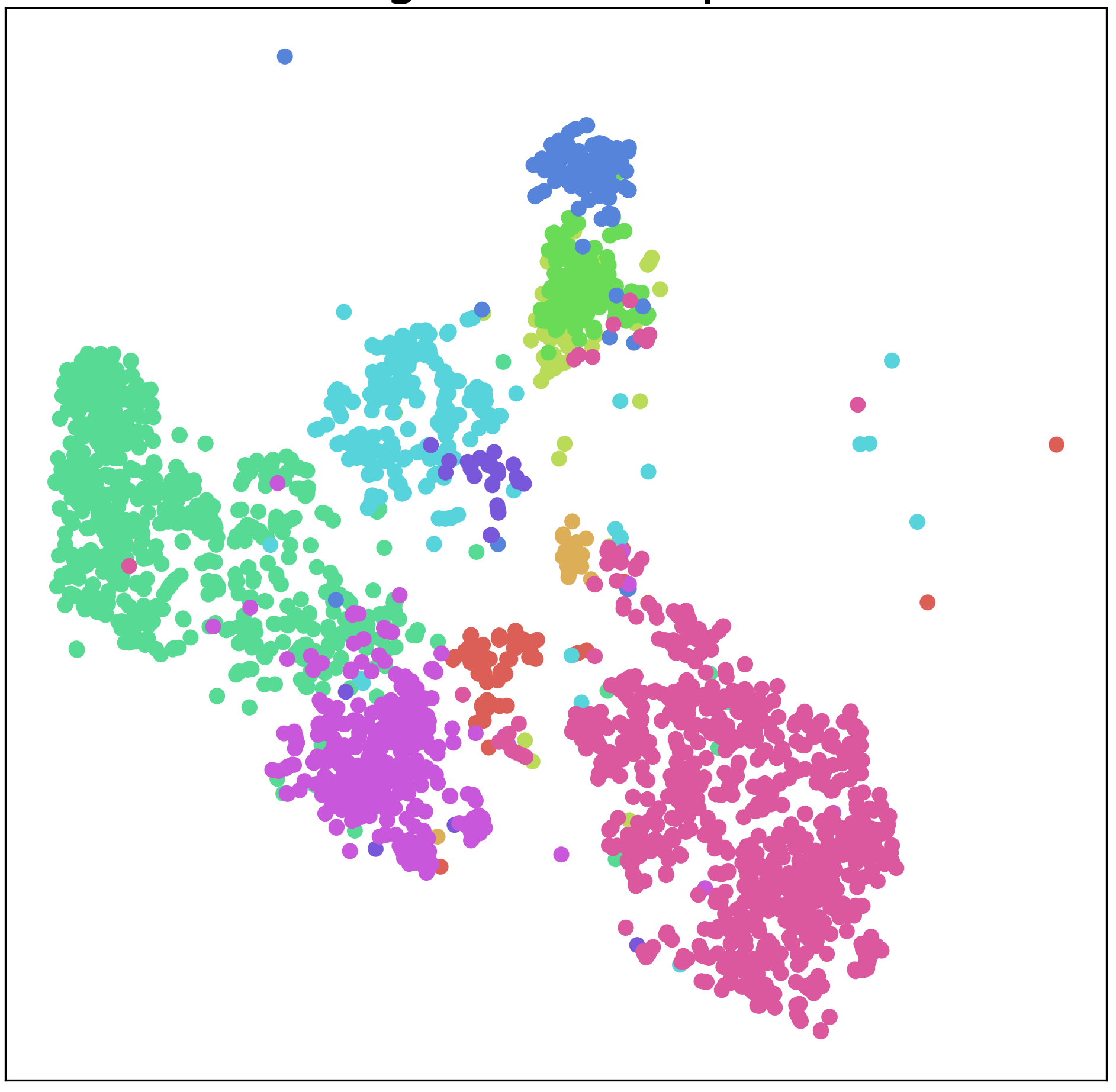}
         \caption{Target COT+SPST: (S*$\rightarrow$S) }
         \label{Point_trgt_COT_SPST_S*_S}
     \end{subfigure}
\caption{$t$-SNE visualization of Source (first two columns) and Target (last two columns) test sets (10 classes) for baseline (only PCM w/o adaptation) and our COT with SPST on all experimental setups of PointDA-10.}
\label{fig: Pointnet tSNE plots}
\end{figure*}

\begin{figure*}[t] 
     \centering
     \begin{subfigure}[b]{0.24\textwidth}
         \centering
         \includegraphics[width=0.85\textwidth]{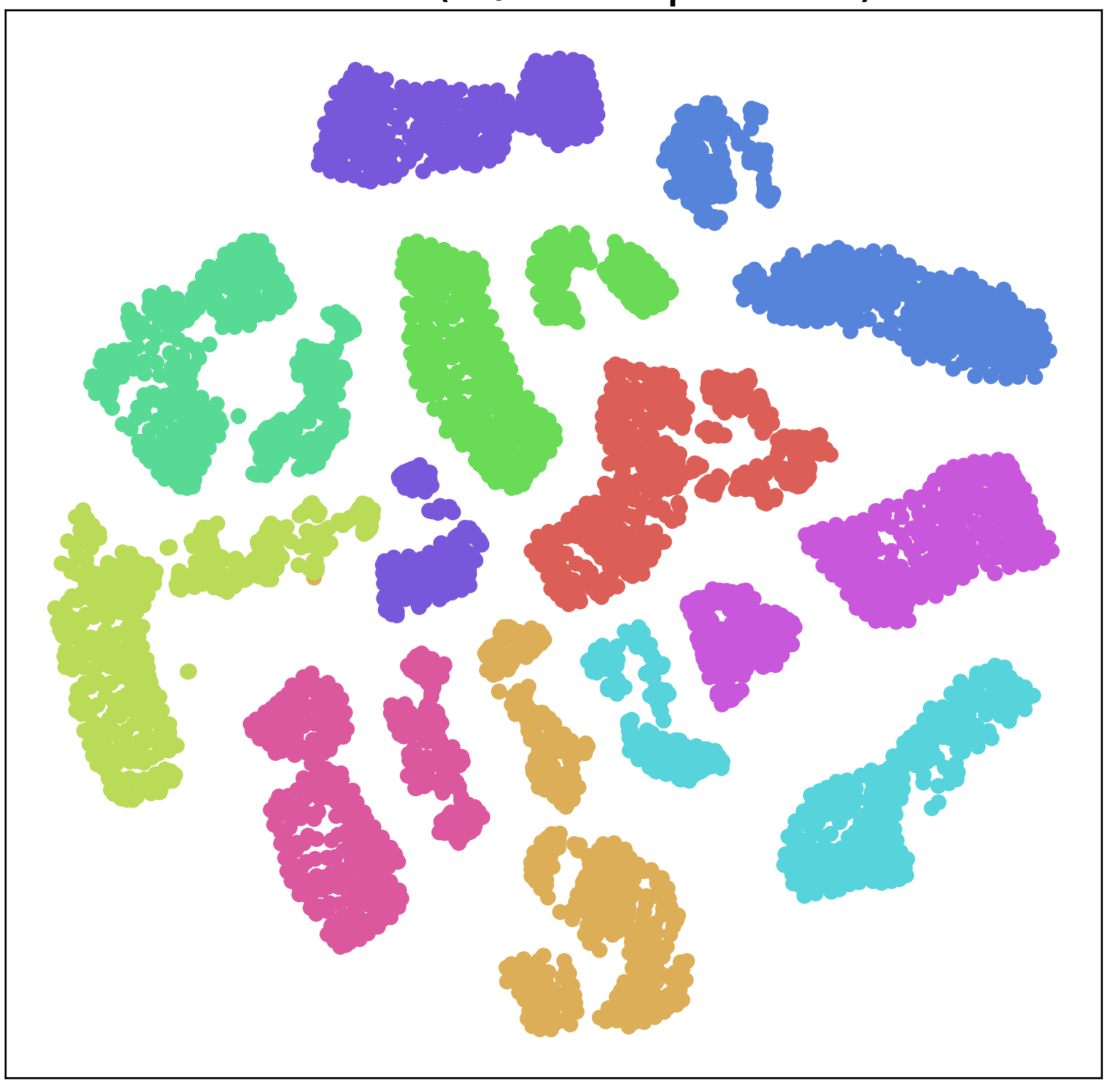}
         \caption{Src Baseline: (Syn.$\rightarrow$Kin.) \vspace{\baselineskip}}
         \label{Grasp_src_pcm_syn_kin}
     \end{subfigure}
     \hfill
     \centering
     \begin{subfigure}[b]{0.25\textwidth}
         \centering
         \includegraphics[width=0.85\textwidth]{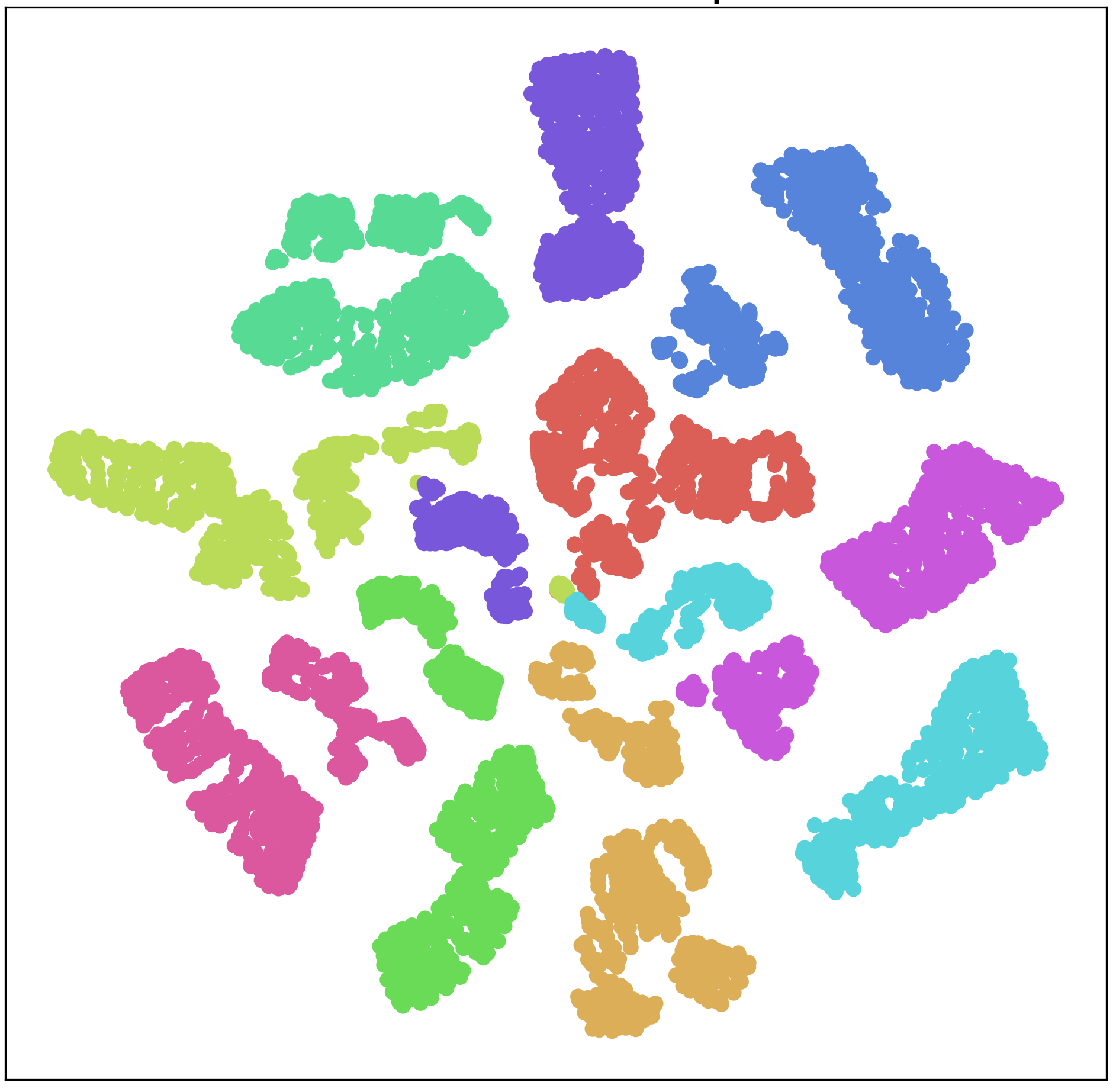}
         \caption{Src COT+SPST: (Syn.$\rightarrow$Kin.) \vspace{\baselineskip}}
         \label{Grasp_src_COT_SPST_syn_kin}
     \end{subfigure}
    \hfill
    \centering
     \begin{subfigure}[b]{0.24\textwidth}
         \centering
         \includegraphics[width=0.85\textwidth]{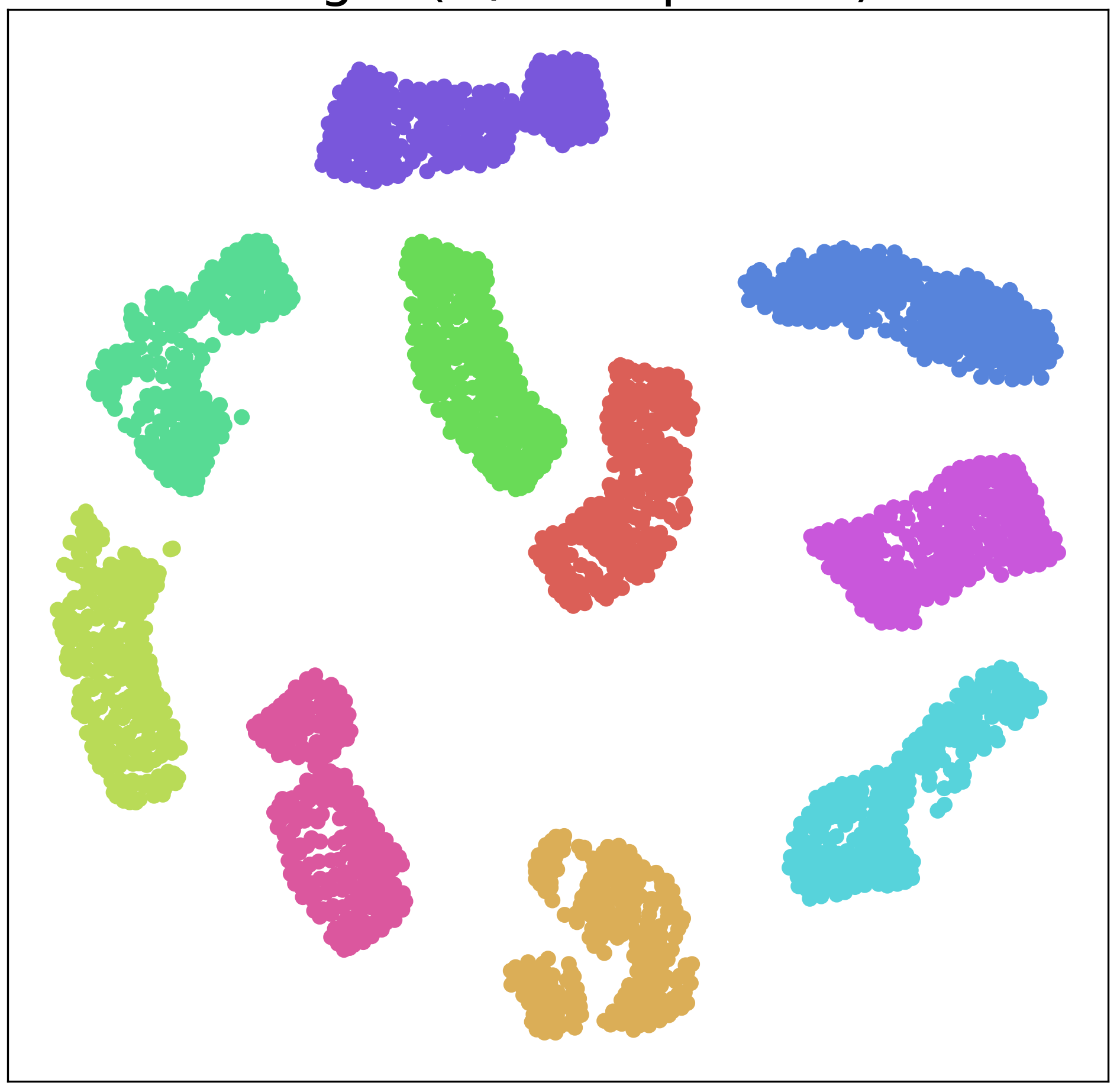}
         \caption{Trgt Baseline: (Syn.$\rightarrow$Kin.) \vspace{\baselineskip} \vspace{\baselineskip}}
         \label{Grasp_trgt_pcm_syn_kin}
     \end{subfigure}
     \hfill
     \centering
     \begin{subfigure}[b]{0.25\textwidth}
         \centering
         \includegraphics[width=0.85\textwidth]{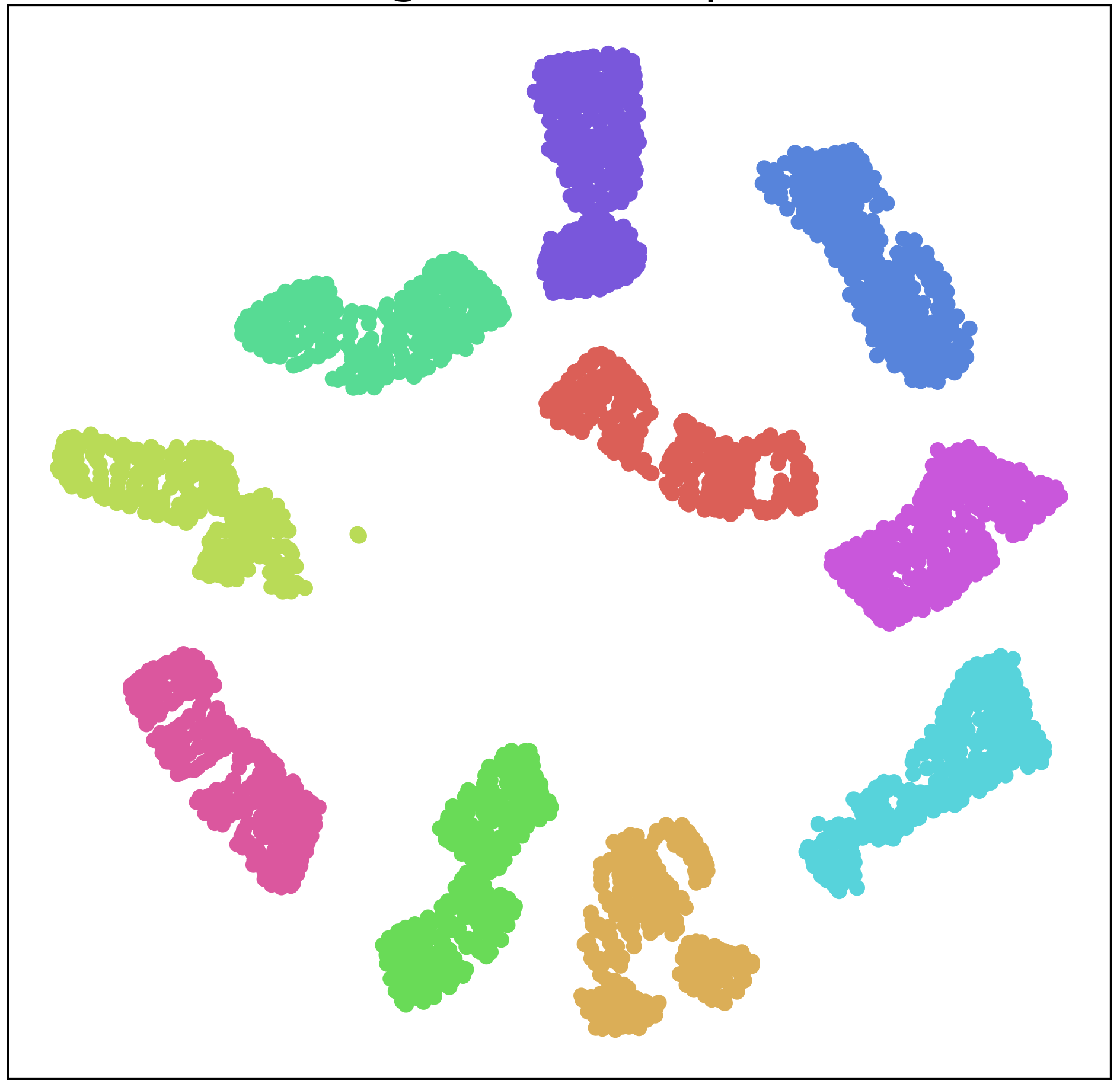}
         \caption{Trgt COT+SPST: (Syn.$\rightarrow$Kin.) \vspace{\baselineskip}}
         \label{Grasp_trgt_COT_SPST_syn_kin}
     \end{subfigure}
     \\
     \centering
     \begin{subfigure}[b]{0.24\textwidth}
         \centering
         \includegraphics[width=0.85\textwidth]{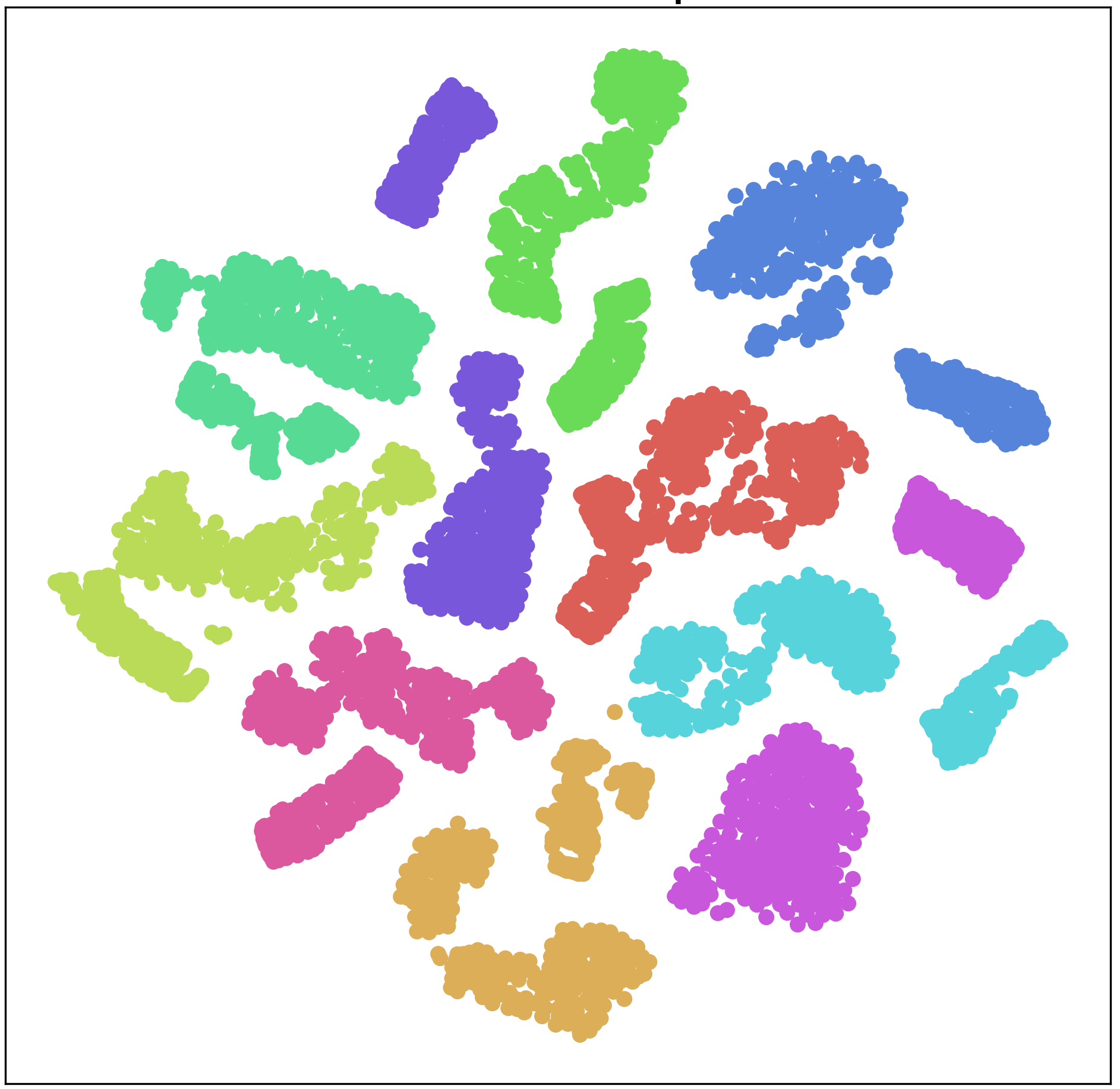}
         \caption{Src Baseline: (Syn.$\rightarrow$RS.) \vspace{\baselineskip} \vspace{\baselineskip}}
         \label{Grasp_src_pcm_syn_real}
     \end{subfigure}
     \hfill
     \centering
     \begin{subfigure}[b]{0.25\textwidth}
         \centering
         \includegraphics[width=0.85\textwidth]{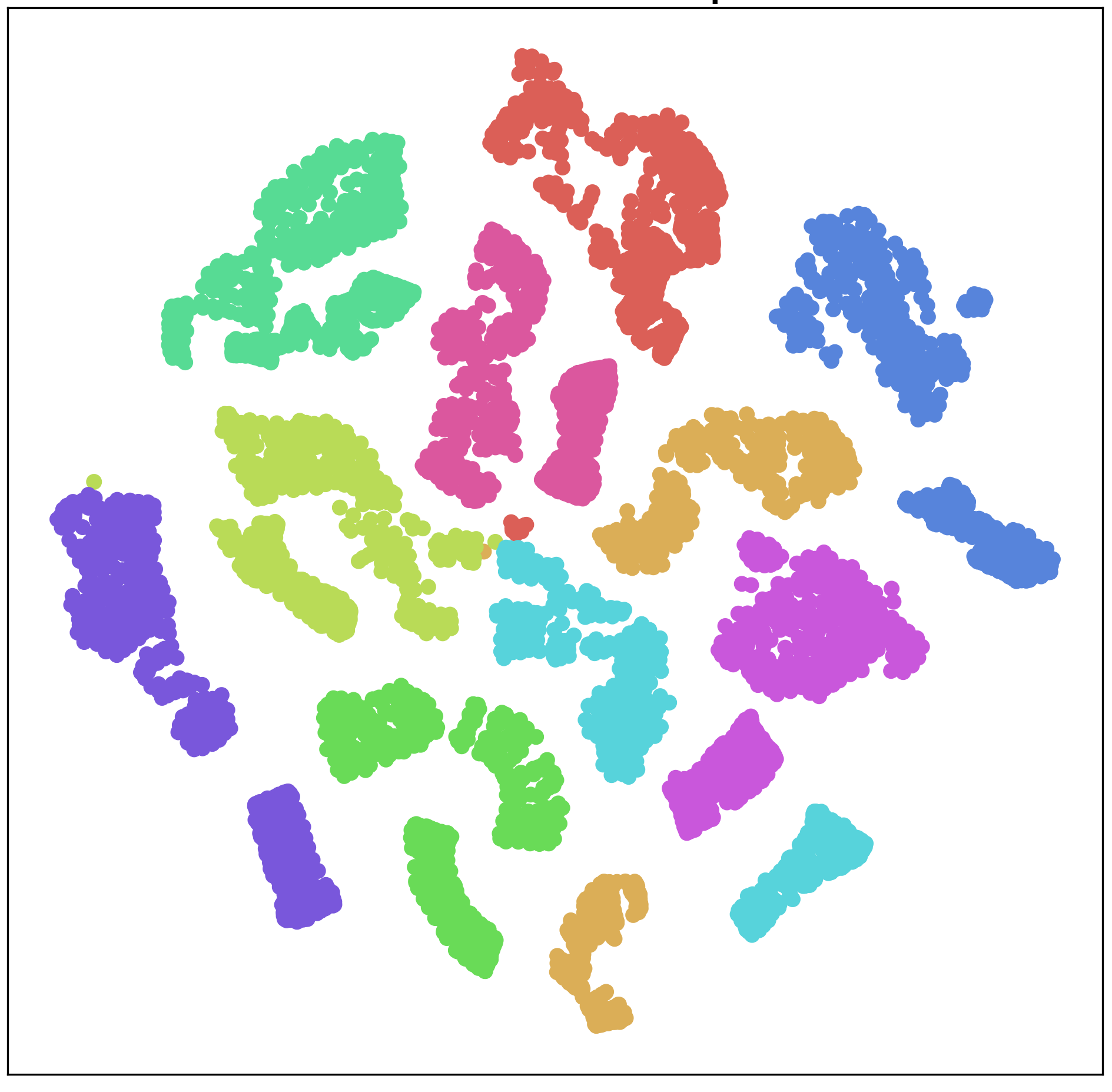}
         \caption{Src COT+SPST: (Syn.$\rightarrow$RS.) \vspace{\baselineskip}}
         \label{Grasp_src_COT_SPST_syn_real}
     \end{subfigure}
     \hfill
     \centering
     \begin{subfigure}[b]{0.24\textwidth}
         \centering
         \includegraphics[width=0.85\textwidth]{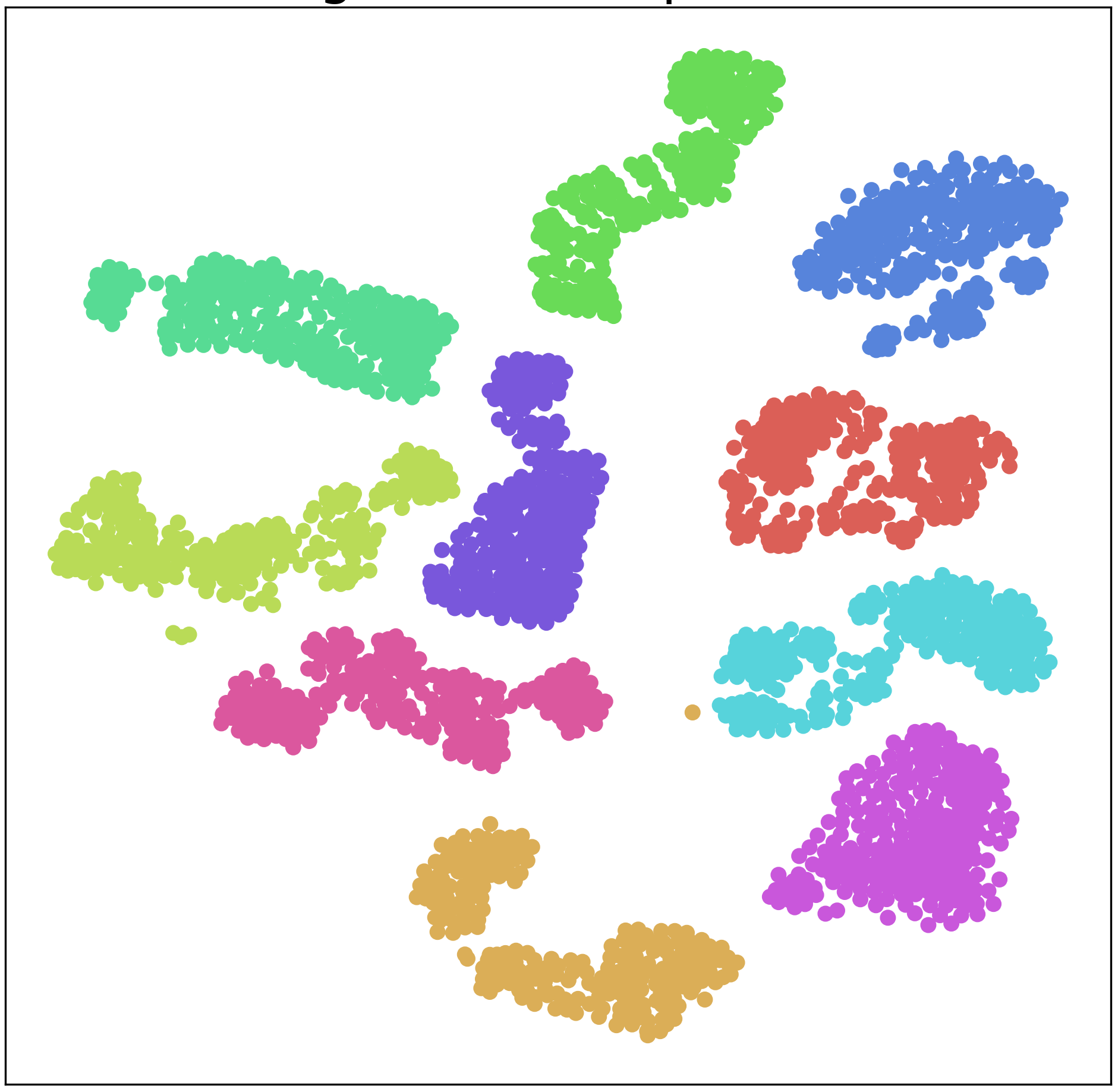}
         \caption{Trgt Baseline: (Syn.$\rightarrow$RS.) \vspace{\baselineskip} \vspace{\baselineskip}}
         \label{Grasp_trgt_pcm_syn_real}
     \end{subfigure}
     \hfill
     \centering
     \begin{subfigure}[b]{0.25\textwidth}
         \centering
         \includegraphics[width=0.85\textwidth]{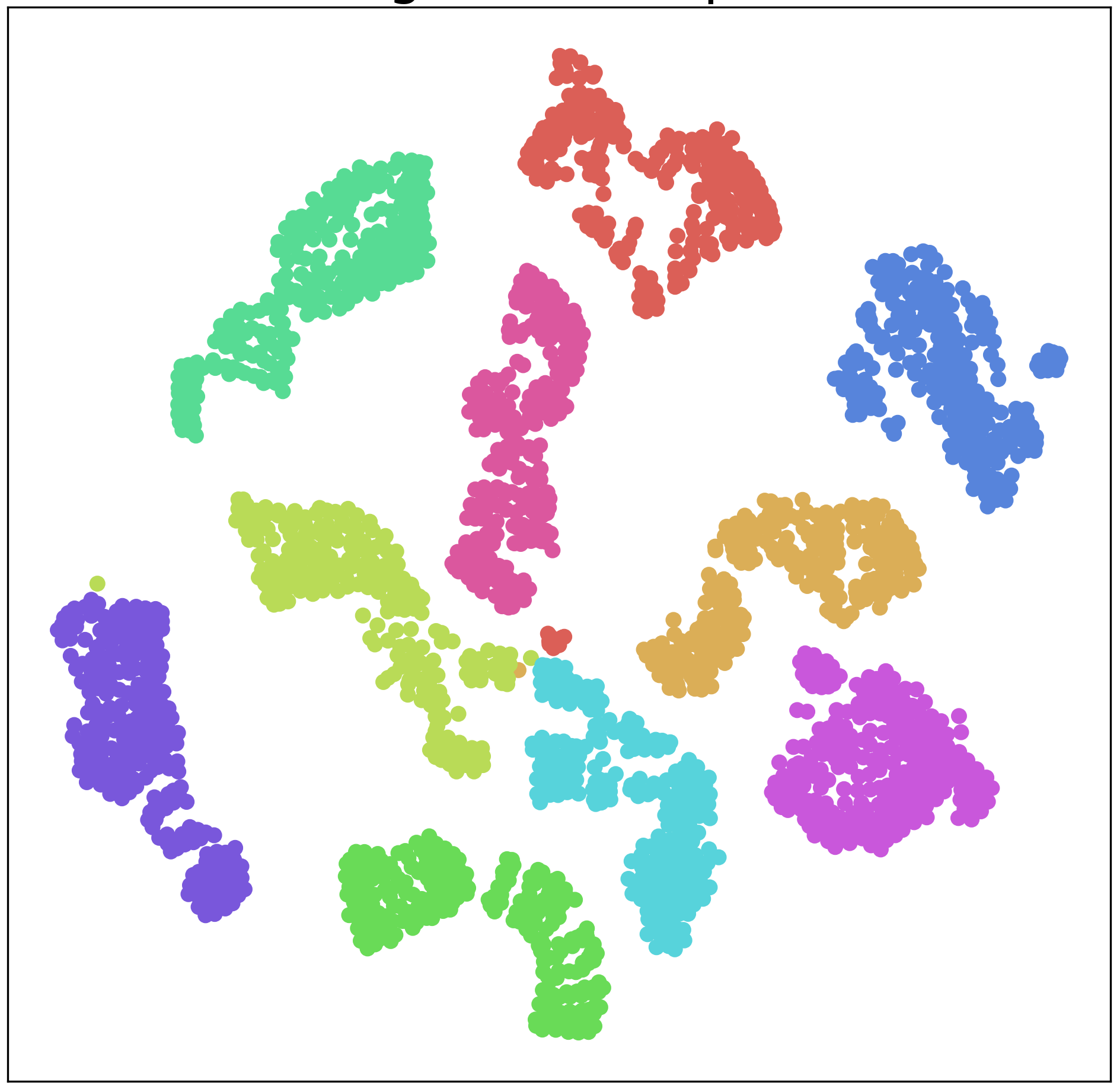}
         \caption{Trgt COT+SPST: (Syn.$\rightarrow$RS.) \vspace{\baselineskip}}
         \label{Grasp_trgt_COT_SPST_syn_real}
     \end{subfigure}
     \\
     \centering
     \begin{subfigure}[b]{0.24\textwidth}
         \centering
         \includegraphics[width=0.85\textwidth]{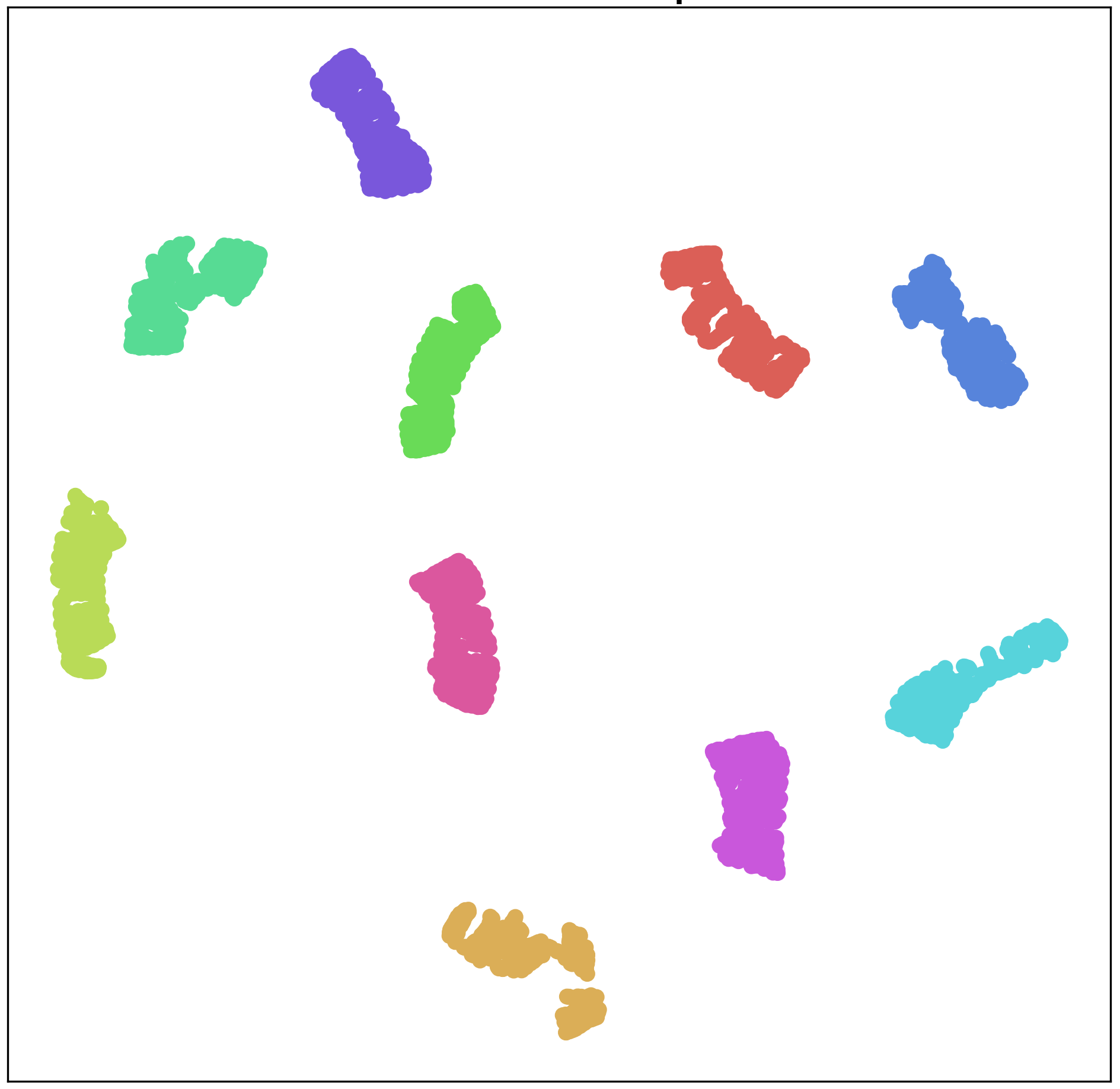}
         \caption{Src Baseline: (Kin.$\rightarrow$ RS.) \vspace{\baselineskip}}
         \label{Grasp_src_pcm_kin_real}
     \end{subfigure}
     \hfill
     \centering
     \begin{subfigure}[b]{0.25\textwidth}
         \centering
         \includegraphics[width=0.85\textwidth]{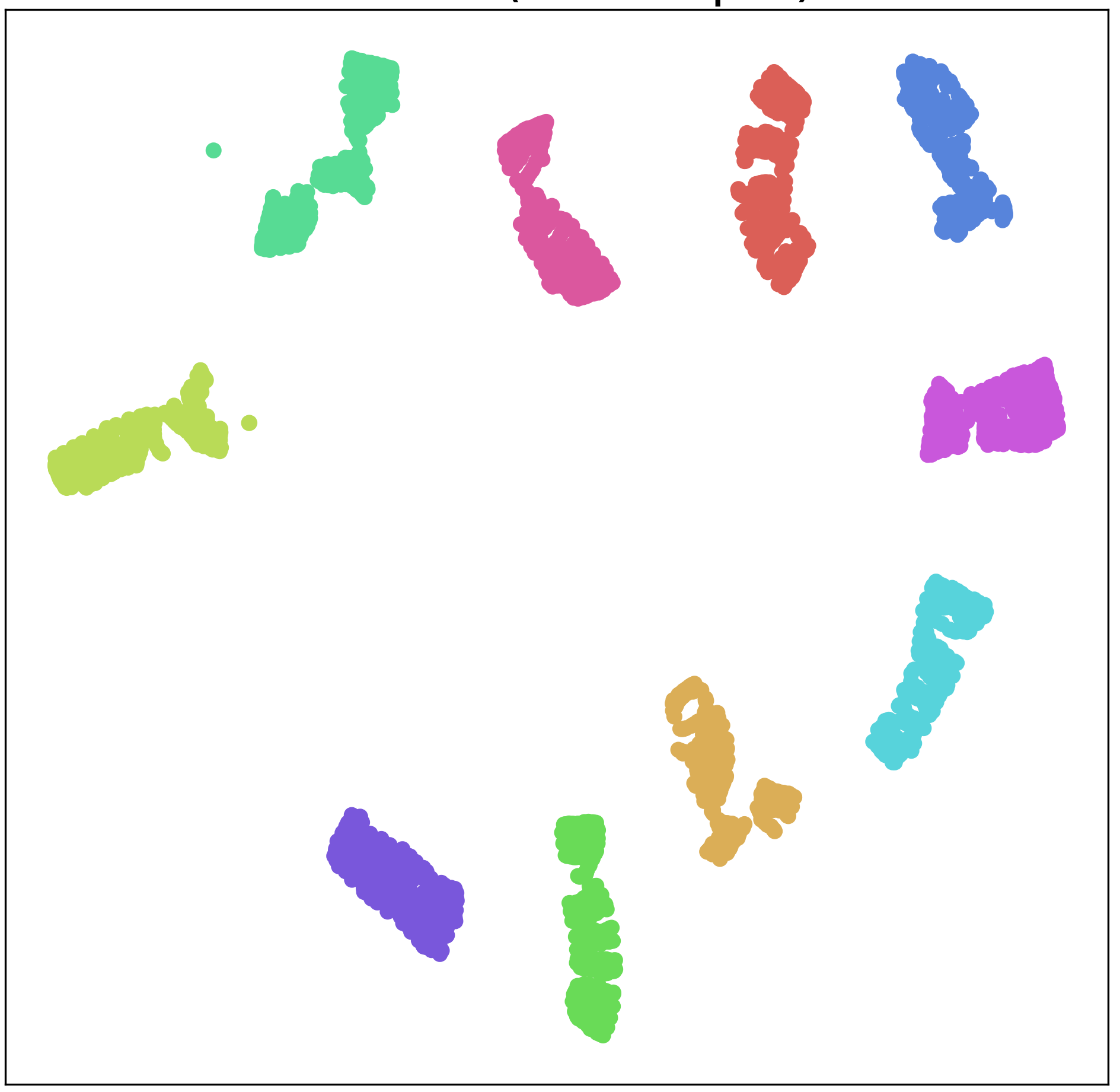}
         \caption{Src COT+SPST: (Kin.$\rightarrow$RS.)}
         \label{Grasp_src_COT_SPST_kin_real}
     \end{subfigure}
     \hfill
     \centering
     \begin{subfigure}[b]{0.24\textwidth}
         \centering
         \includegraphics[width=0.85\textwidth]{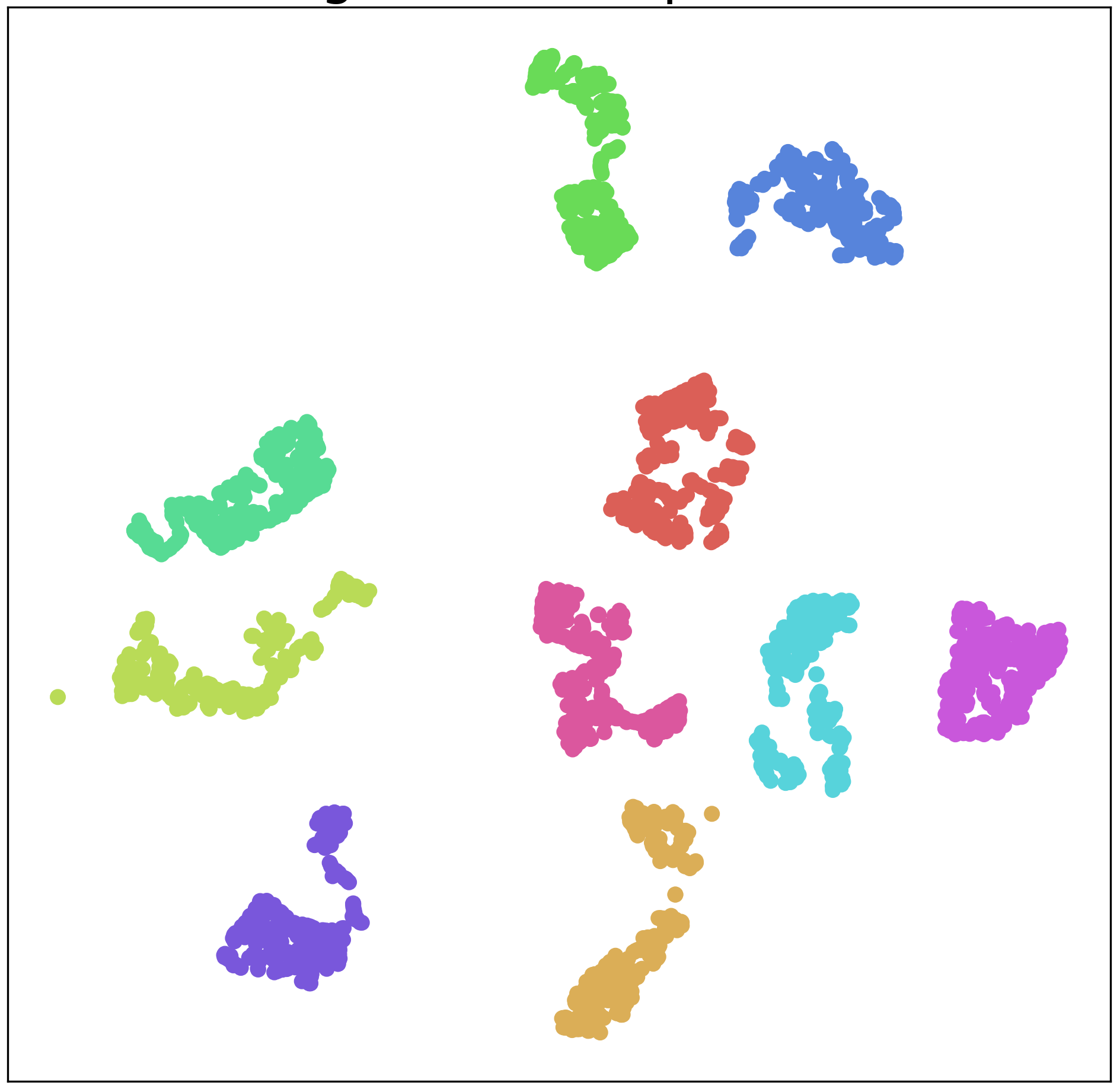}
         \caption{Trgt Baseline: (Kin.$\rightarrow$RS.) \vspace{\baselineskip}}
         \label{Grasp_trgt_pcm_kin_real}
     \end{subfigure}
     \hfill
     \centering
     \begin{subfigure}[b]{0.25\textwidth}
         \centering
         \includegraphics[width=0.85\textwidth]{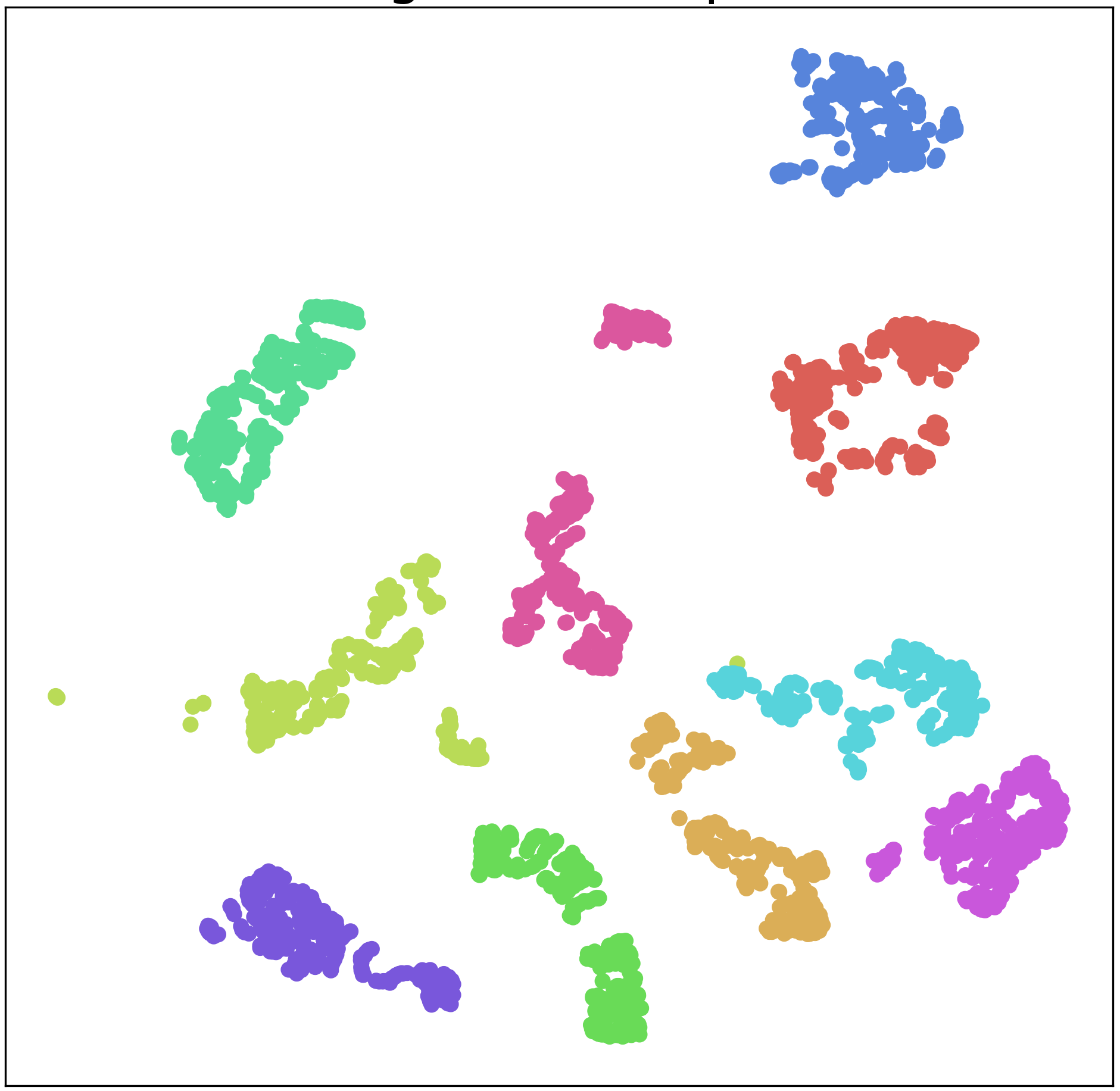}
         \caption{Trgt COT+SPST: (Kin.$\rightarrow$RS.)}
         \label{Grasp_trgt_COT_SPST_kin_real}
     \end{subfigure}
     \\
     \centering
     \begin{subfigure}[b]{0.24\textwidth}
         \centering
         \includegraphics[width=0.85\textwidth]{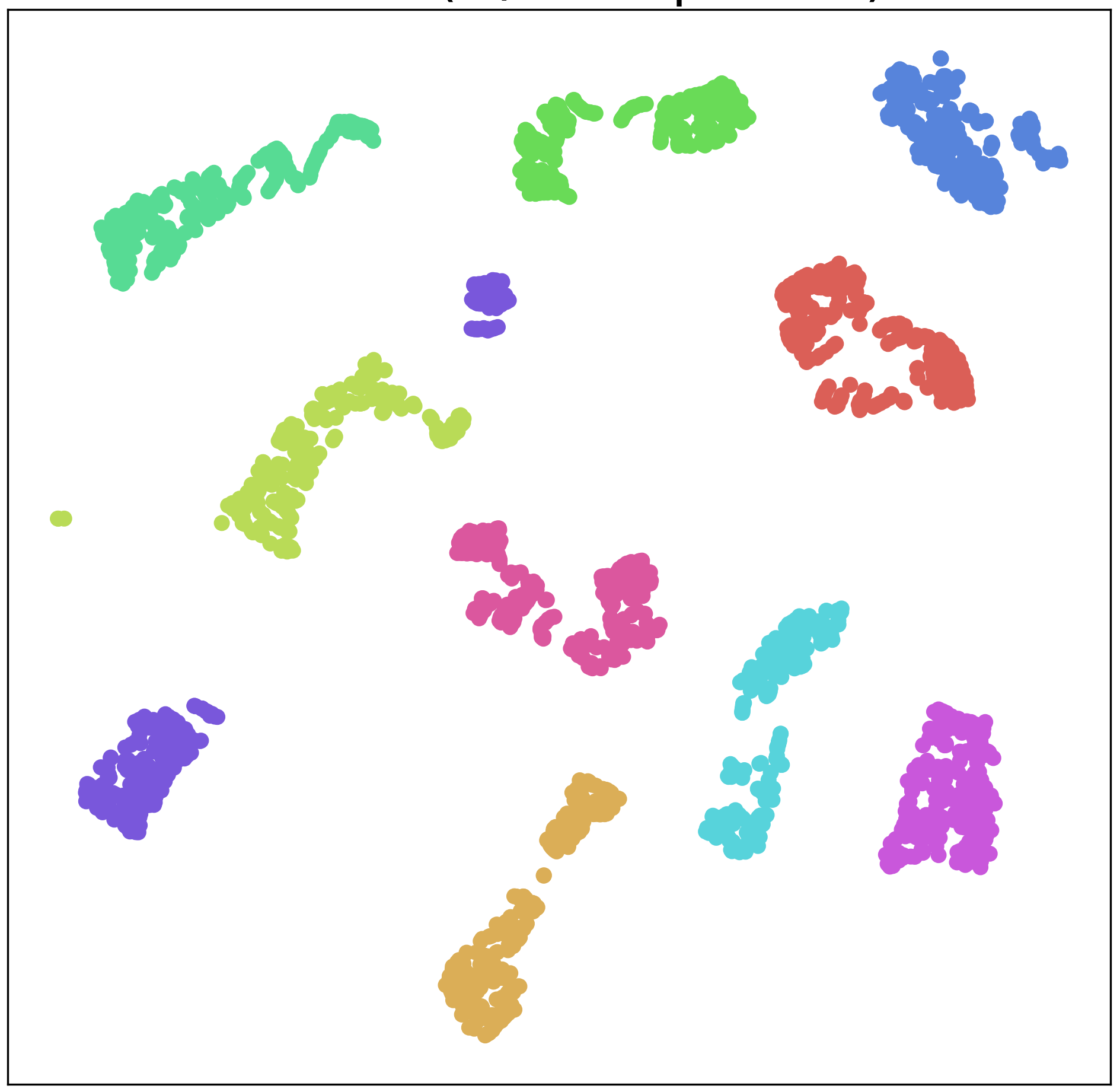}
         \caption{Src Baseline: (RS.$\rightarrow$Kin) \vspace{\baselineskip} \vspace{\baselineskip}}
         \label{Grasp_src_pcm_real_kin}
     \end{subfigure}
     \hfill
     \centering
     \begin{subfigure}[b]{0.25\textwidth}
         \centering
         \includegraphics[width=0.85\textwidth]{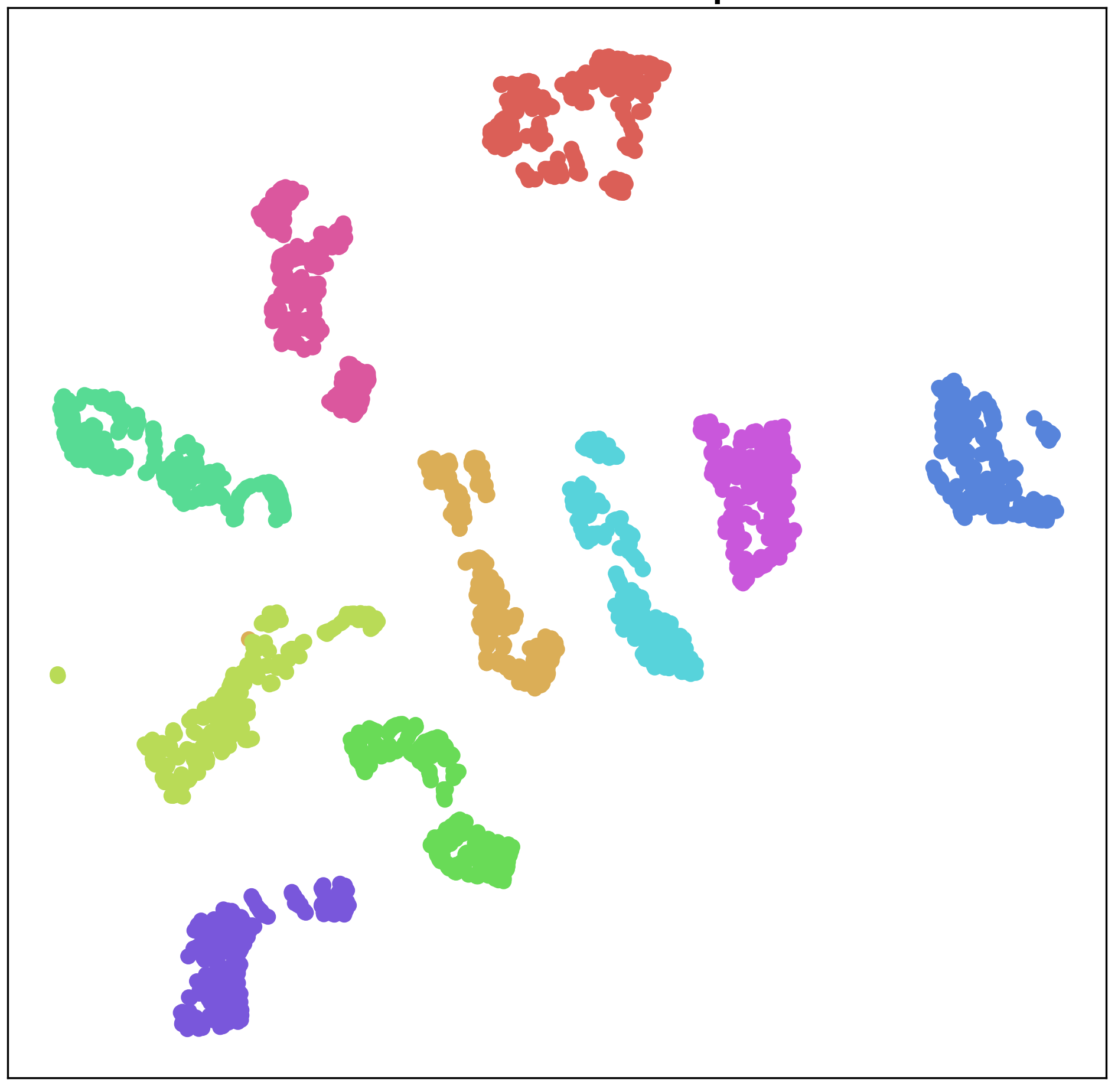}
         \caption{Src COT+SPST: (RS.$\rightarrow$Kin) \vspace{\baselineskip}}
         \label{Grasp_src_COT_SPST_real_kin}
     \end{subfigure}
    \hfill
    \centering
     \begin{subfigure}[b]{0.24\textwidth}
         \centering
         \includegraphics[width=0.85\textwidth]{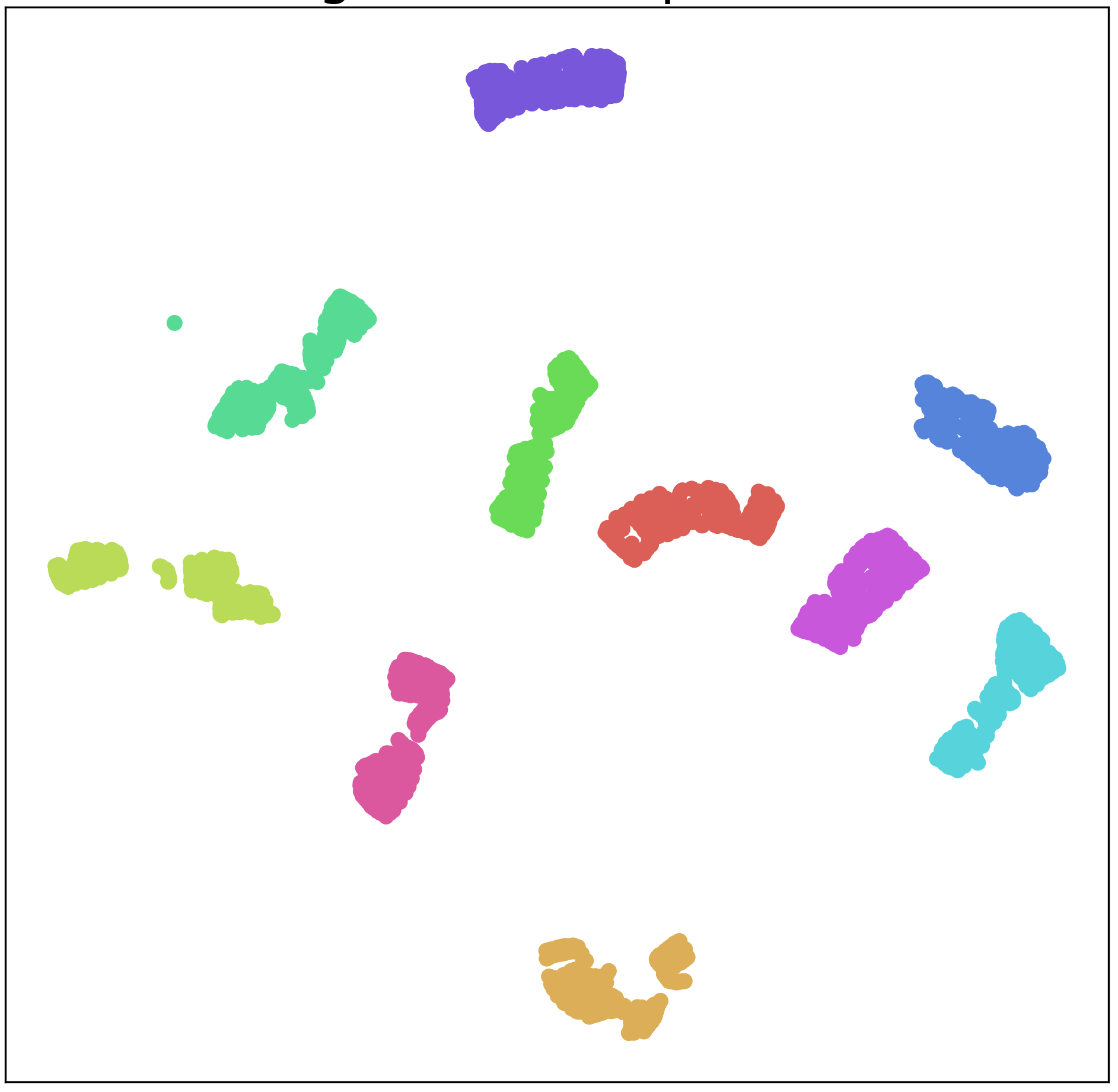}
         \caption{Trgt Baseline: (RS.$\rightarrow$Kin) \vspace{\baselineskip} \vspace{\baselineskip}}
         \label{Grasp_trgt_pcm_real_kin}
     \end{subfigure}
     \hfill
     \centering
     \begin{subfigure}[b]{0.25\textwidth}
         \centering
         \includegraphics[width=0.85\textwidth]{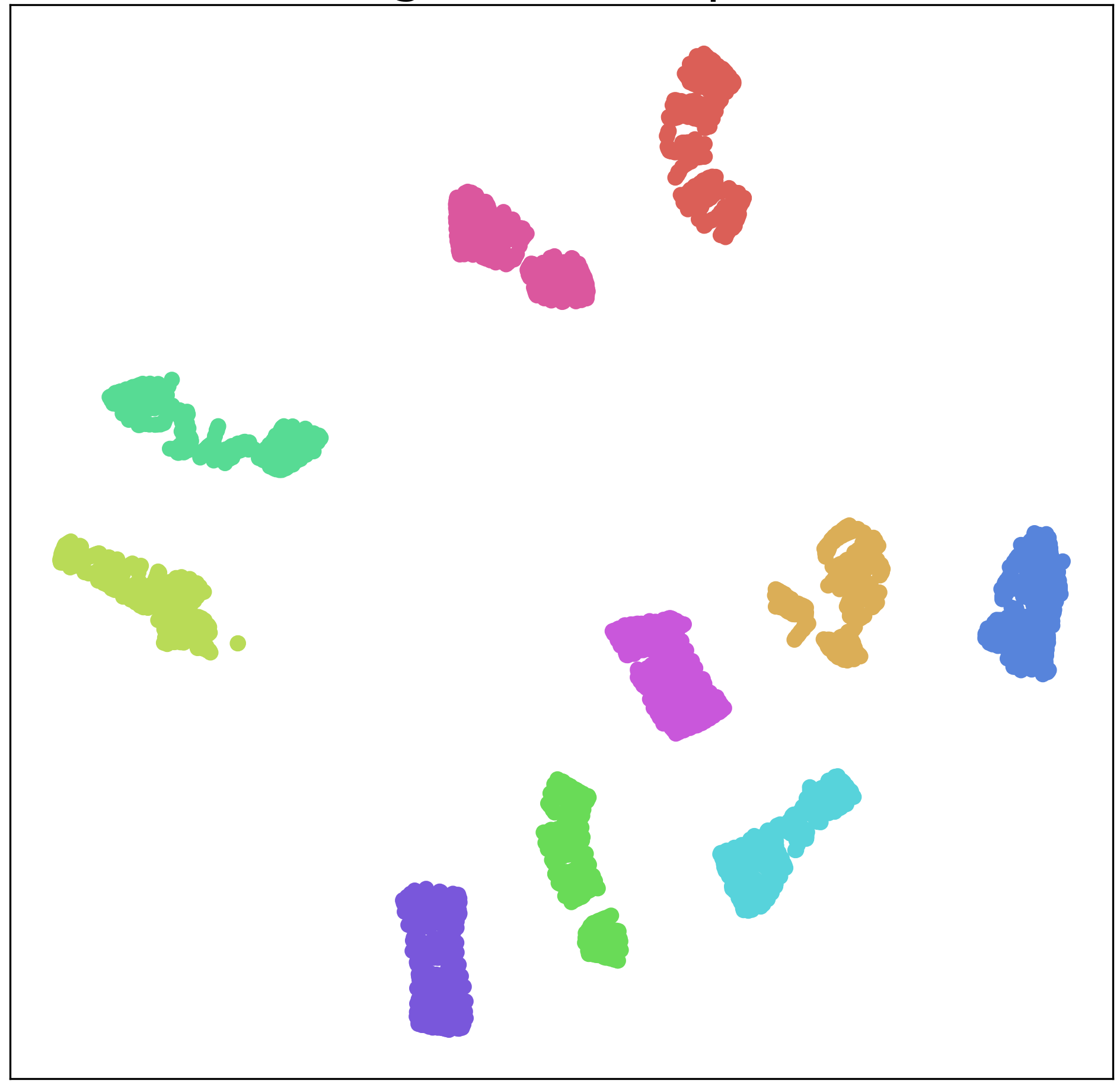}
         \caption{Trgt COT+SPST: (RS.$\rightarrow$Kin) \vspace{\baselineskip}}
         \label{Grasp_trgt_COT_SPST_real_kin}
     \end{subfigure}
\caption{$t$-SNE visualization of Source (first two columns) and Target (last two columns) test sets (10 classes) for baseline (only PCM w/o adaptation) and our COT with SPST on all experimental setups of GraspNetPC-10.}
\label{fig: Graspnet tSNE plots}
\end{figure*}
\cleardoublepage
\cleardoublepage
\cleardoublepage
\cleardoublepage
\section{t-SNE Visualization}
We visualize point cloud embeddings of our learned model for PointDA-10 and GraspNetPC-10 datasets using t-SNE. Figure \ref{fig: Pointnet tSNE plots} contains t-SNE plots
for all source-target experimental settings from PointDA-10. Figure \ref{fig: Graspnet tSNE plots} contains t-SNE plots for all source-target experimental settings from Graspnet-10. We use source and target test sets for plotting these t-SNE plots, except for the Syn. dataset from GraspNetPC-10 for which the test set is not available. We consider validation set for Syn. (synthetic) dataset.

Comparing the t-SNE plots between baseline (only PCM without adaptation) and our best performing method (COT with SPST strategy) in both source and target domains, we can observe inter-cluster distances getting increased with well-defined class separations in our proposed method. These plots also show class domain alignment.


\end{document}